%% file: DifferentialsKinMapping_final.tex
\documentclass[singlecolumn,10pt,twoside,final]{asme2ej_Draft}
\usepackage{amsfonts}
\usepackage{graphicx}
\usepackage{amsmath, amssymb, bm}
\usepackage[fixamsmath,disallowspaces]{mathtools}
\usepackage{xcolor}

\newtheorem{remark}{Remark}

\newtheorem{assumption}{Assumption}

\begin{document}

\parindent 0pt \parskip 2pt \setcounter{topnumber}{9} %
\setcounter{bottomnumber}{9} \renewcommand{\textfraction}{0.00001}

\renewcommand {\floatpagefraction}{0.999} \renewcommand{\textfraction}{0.01} %
\renewcommand{\topfraction}{0.999} \renewcommand{\bottomfraction}{0.99} %
\renewcommand{\floatpagefraction}{0.99} \setcounter{totalnumber}{9}

\title{ 
\vspace{-3ex}
An Overview of Formulae for the Higher-Order Kinematics of Lower-Pair Chains wit Applications in Robotics and Mechanism Theory
\vspace{-6mm}
} 
\author{
	Andreas M\"uller
	\affiliation{JKU Johannes Kepler University, Linz, Austria, a.mueller@jku.at}
\vspace{-7ex}
} 

\maketitle 

\begin{abstract}

\input{abstract}

\end{abstract} 

\section{Introduction}

A central part of the kinematics modeling is to express the configuration
(pose, posture) of an open kinematic chain in terms of assigned joint
variables. This functional relation is described by the \emph{kinematic
mapping (KM)}. The KM serves to describe the kinematics of serial robotic
arms and multibody systems (MBS) with tree topology, but also the loop
closure constraints of mechanisms and general MBS. The relation of the twist
of the kinematic chain and the joint velocities is determined by the
Jacobian of the KM, which serves as the forward kinematics Jacobian of a
serial robotic arm as well as the constraint Jacobian for a kinematic loop.

An important concept in modern kinematics is the \emph{product of
exponentials (POE)} \cite{Brockett1984,LynchPark2017,Murray,Selig}. Its
fundamental advantage is that the KM of a kinematic chain with lower pair
joints is completely parameterized in terms of (readily available) geometric
data rather than following restrictive modeling conventions such as DH
parameters. Moreover, using the POE makes it possible to derive closed form
algebraic expressions for partial derivatives of the KM and the \emph{%
geometric Jacobian}, but also for the time derivatives of the twist of a
kinematic chain, of arbitrary order. Since such formulations are scattered
in the literature, and presented in various different forms, e.g. \cite%
{RicoDuffy1996,Rico1999,MMTConstraints,MMTHighDer,LerbetBook,Lerbet1999,Karger1989,Karger1996,Selig,Murray,LynchPark2017,ParkBobrowPloen1995,Herve1978,Herve1982,Cervantes2009,Chevallier1984,Chevallier1991,Gallardo2008}
they are not yet established as a generally applicable modeling approach,
despite the recent interest in screw and Lie group modeling of mechanisms
and MBS \cite{MUBOScrews1,MUBOScrews2,ParkKimJangHong2018}.

The driving force behind the research on higher-order time derivatives and
partial derivatives of the geometric Jacobian has been the mobility and
singularity analysis of linkages and robots \cite%
{RicoRavani2003,RicoRavani2006,Rico2007,Karger1996}. A central result is
that these only require the derivatives of the instantaneous joint screws,
which are given by (nested) Lie-brackets, i.e. screw products, of the joint
screws. Further, a recent interest in explicit compact relations for
higher-order time derivatives of twists (accelerations, jerk, jounce/snap,
etc.) stems from the development of advanced methods for the optimal
trajectory planning and model-based control of robots and general MBS. In
particular, the control of robots equipped with series elastic actuators
(SEA) requires computation of the 4th time derivative of twists
(jounce/snap) \cite%
{deLuca1998,PalliMelchiorriDeLuca2008,BuondonnaDeLuca2015,Giusti2018}, and
the time optimal trajectory planning must respect the bounds on higher
derivatives of the inverse dynamics, which requires the derivatives of the
forward kinematics \cite{ConstantinescuCroft2000,ReiterTII2018}.
Accordingly, the path planning requires solving the higher-order kinematics
problem of a serial robotic arms, i.e. determination of higher time
derivatives of the velocity inverse kinematics solution.

The POE formulation gives rise to further closed form expressions that are
relevant for the mobility and singularity analysis of linkages, as well as
the control of robotic arms, which have not yet been reported in the
literature. One is the Taylor series expansion of the KM. It has been
observed recently that the local analysis based on higher-order time
derivatives may not be sufficient to deduce the mobility of certain
mechanisms \cite{ConnellyServatius1994,PabloMMT2018,JMR2018LocApprox}, and
that an exhaustive local mobility analysis must further investigate the
local geometry of the configuration space of a linkage, which can be pursued
with a Taylor series expansion of the KM. Another important relation is the
closed form of the time derivatives of the minors of the geometric Jacobian.
The latter facilitate the detection and analysis of kinematic singularities.

While various of the relations listed above were already reported in the
literature (although in different forms using different notations), several
of them have not yet been published.

Therefore this paper aims to provide a comprehensive overview of the closed
form expressions for higher-order kinematic relations of kinematic chains
with lower pair joints using the POE in a consolidated formulation, as far
as relevant for the kinematic analysis, motion planning, and control of
mechanisms and robots. The following \emph{mathematical} topics are addressed

\begin{itemize}
\item[$\circ $] multiple partial derivatives of the geometric Jacobian of
arbitrary order (section \ref{secPartialS})

\item[$\circ $] higher-order time derivatives of twists of a kinematic chain
(section \ref{secTimeDerS})

\item[$\bullet$] Taylor series expansion of the KM (section \ref{secSeriesKM}%
)

\item[$\bullet$] arbitrary time derivatives of the minors of the geometric
Jacobian (section \ref{secTimeDerMin})

\item[$\bullet$] Taylor series expansion of the minors of the geometric
Jacobian (section \ref{secSeriesMinor})

\item[$\bullet$] higher-order inverse kinematics of robotic arm (section \ref%
{secInvKinArm})

\item[$\bullet$] Taylor series expansion of the geometric Jacobian (section %
\ref{secTaylorJacobian})

\item[$\bullet$] Taylor series expansion of the solution of a kinematic loop
in terms of time derivatives of independent joint variables (section \ref%
{secLoopSolution})
\end{itemize}

where results that where already published are summarized (indicated by $%
\circ $) or novel relations are derived (indicated by $\bullet $). It is
discussed how these results can be applied to the following topics in \emph{%
robotics and mechanisms}

\begin{itemize}
\item[-] Higher-order forward kinematics of serial robotic manipulators and
MBS (section 5.4a)

\item[-] Higher-order inverse kinematics of non-redundant serial robotic
manipulators (section 8)

\item[-] Local mobility and singularity analysis of linkages (section 5.4d,
6.3, and 7.3)

\item[-] Computation of gradients of kinematic manipulability measures for
robot manipulators (section 5.4c)

\item[-] Determination of the generic structural mobility of linkages
(section 9.3)

\item[-] Approximate solutions of the loop constraints of linkages (section
10.2)
\end{itemize}

The paper is intended as a reference where the reader can look up the
relevant relations directly from the respective section while the
introduction section 2 should serve as a short introduction to the notation.

The paper is organized as follows. Section 2 introduces the POE formulation
for the KM. To this end, the screw coordinates for lower pair joints are
introduced. The geometric Jacobian is then introduced in section 3 where its
columns are identified as the instantaneous joint screws determined by frame
transformations of the joint screws. The explicit form of the partial
derivatives of the instantaneous joint screws are presented in section \ref%
{secPartialS}. The explicit form of repeated partial derivatives is
presented. Higher-order time derivatives of the twist of a rigid body of the
kinematic chain are presented in section \ref{secTimeDerS}. Closed form
algebraic expressions for derivatives up to second order are presented in
terms of partial derivatives of the Jacobian. For time derivatives of
arbitrary order, a recursive relation is presented. A recursive $O\left(
n\right) $ formulation up to 4th-order is also presented. It is discussed
how these relations can be applied to the higher-order forward kinematics
and inverse dynamics of a robotic arm, to compute the gradients of dexterity
measures, and to the local mobility analysis of linkages. Algebraic
relations for higher-order differentials of the KM are reported in section %
\ref{secSeriesKM}. A recursive relation for arbitrary order and explicit
expressions for differentials up to 4th order are given. The differentials
are used for a Taylor series expansion of the KM. It is discussed how such
series expansion can be used to approximate the local geometry of the
configuration space of a linkage. Section \ref{secJacDer} addresses the time
derivatives and the differentials of the minors of the Jacobian. The
presented closed form relations are applied to the higher-order
approximation of the motions of a linkage where the constraint Jacobian
exhibits a permanent drop of rank. In section \ref{secInvKinArm} the
higher-order inverse kinematics of a non-redundant serial chain (e.g.
robotic arm) is addressed. A recursive relation for the solution of the
inverse kinematics problem of arbitrary order is presented along with a
closed form relation for the solution up to order 4. It is briefly discussed
how these results can be used for the control of robots. Section \ref%
{secTaylorJacobian} investigates how the structural properties of a
kinematic chain, in particular its motion space, can be extracted from the
joint screw system. To this end, the geometric Jacobian is expanded into a
Taylor. Since this is given in terms of nested Lie brackets of joint screws,
it is concluded that the motion space is the Lie subgroup corresponding to
the Lie algebra generated by the joint screws. This result is related to the
well-known structural mobility formulae. The higher-order time derivatives
of the loop constraints are used to derive an algebraic expression for a
higher-order approximate solution of the geometric loop constraints. This is
presented in section \ref{secLoopSolution}. All formulations in sections 2-9
used the spatial representation of twists and screws. In various
applications, however, the body-fixed or hybrid representations are used.
The relation of the time derivatives of the latter are related to those of
the spatial representation in section \ref{secRepresentations}. This allows
application of all presented results also when other representations are
used, while exploiting the efficiency of the spatial formulations. The
necessary geometric background on the Lie group of rigid body motions and
the exponential mapping is summarized in appendix \ref{secGeomBackground},
which shallow the reader to follow without the need to (immediately) consult
secondary literature. For better readability the notation used in this paper
is summarized in appendix \ref{secNotation}.

The paper uses concepts and notations related to the Lie group $SE\left(
3\right) $ of rigid body motions. Details can be found in the textbooks \cite%
{LynchPark2017,Murray,Selig}.

Most of the reported relations have been implemented as package for the
computer algebra system Mathematica$^{\copyright }$. This package along with
several examples has been submitted as supplementary material in \cite%
{MendeleyDataset}.

\begin{remark}
In this paper the KM is understood as a mapping from joint space to $%
SE\left( 3\right) $, which is in accordance with its use in \cite%
{Donelan2007}. This must not be confused with the notion of the kinematic
mapping as introduced by Blaschke \cite{Blaschke1911,Blaschke1942}, or Study 
\cite{Study1891,Study1903}, see also \cite{BottemaRoth1990}. The latter is a
mapping from $SE\left( 3\right) $ to ${\mathbb{P}}^{7}$ that was used to
provide algebraic equations for the computational analysis of mechanism
kinematics \cite{HustySchroecker2009,PfurnerKong2017}.
\end{remark}

\section{Kinematic Mapping and the Product of Exponentials}

In the following a general simple kinematic chain \cite{McCarthy1990}
comprising $n$ 1-DOF lower pair joints and $n$ rigid links is considered,
where joint 1 connects to the ground at which a global reference frame $%
\mathcal{F}_{0}$ is defined. The motion of the kinematic chain is hence due
to its internal mobility. Denote with $\mathbf{q}\in {\mathbb{V}}^{n}$ the
vector of $n$ joint variables, where the joint space manifold is ${\mathbb{V}%
}^{n}={\mathbb{R}}^{n_{\mathrm{P}}}\times {\mathbb{T}}^{n_{\mathrm{R}}}$,
with $n_{\mathrm{R}}$ being the number of revolute joints and $n_{\mathrm{P}%
} $ the number of prismatic or helical joints.

At link $i$ a body-fixed reference frame $\bar{\mathcal{F}}_{i}$ is
attached, which coincides with $\mathcal{F}_{0}$ at reference configuration $%
\mathbf{q}=\mathbf{0}$ of the open kinematic chain. The spatial
configuration (posture) of link $i$ is uniquely represented by the $4\times
4 $ transformation matrix \cite{Angeles2007,Murray,Selig}%
\begin{equation}
\bar{\mathbf{C}}_{i}=\left( 
\begin{array}{cc}
\bar{\mathbf{R}}_{i} & \bar{\mathbf{r}}_{i} \\ 
\mathbf{0} & 1%
\end{array}%
\right) \in SE\left( 3\right)  \label{Cbar}
\end{equation}%
transforming homogenous coordinates of a point when expressed in the frame $%
\bar{\mathcal{F}}_{i}$, to those when expressed in the world frame $\mathcal{%
F}_{0}$, where $\bar{\mathbf{r}}_{i}\in {\mathbb{R}}^{3}$ is the position
vector of the origin of $\bar{\mathcal{F}}_{i}$ measured and resolved in $%
\mathcal{F}_{0}$, and $\bar{\mathbf{R}}_{i}\in SO\left( 3\right) $ is the
rotation matrix transforming coordinates of vectors resolved in $\bar{%
\mathcal{F}}_{i}$ to those when resolved in $\mathcal{F}_{0}$. 
\begin{figure}[h]
\centerline{\includegraphics[width=0.55%
\textwidth]{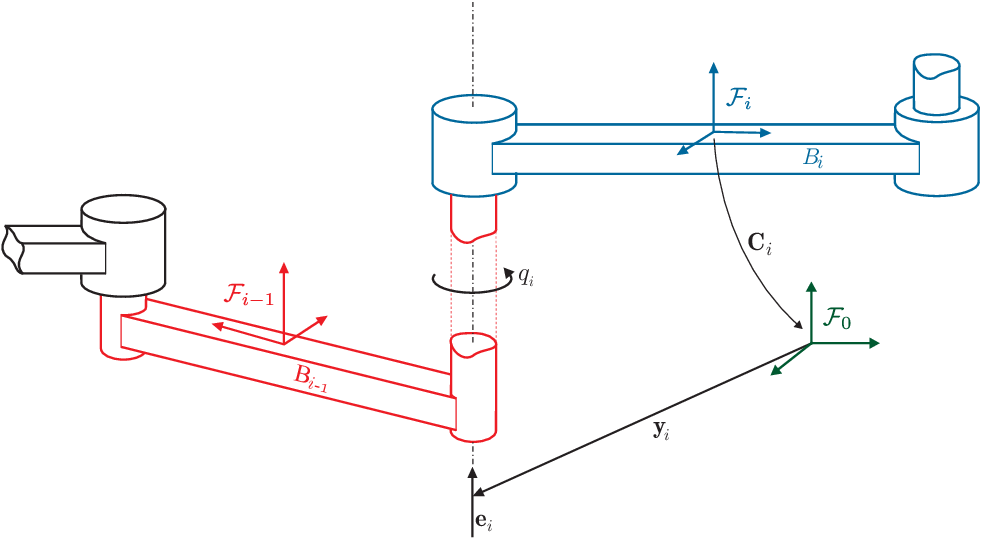}}
\caption{Definition of a screw associated to a 1-DOF lower pair joint.}
\label{figJointScrew}
\end{figure}

A linkage is a system of rigid bodies interconnected by lower pair joints.
The joint motions can thus be described as screw motions and hence be
expressed by the exponential of joint screw coordinates (sec. \ref{secExp}).
The configuration of link $i$ is determined by the joint variables as $\bar{%
\mathbf{C}}_{i}=f_{i}\left( \mathbf{q}\right) $, with the \emph{kinematic
mapping (KM) of link} $i$ 
\begin{equation}
f_{i}\left( \mathbf{q}\right) =\exp \left( \mathbf{Y}_{1}q_{1}\right) \exp
\left( \mathbf{Y}_{2}q_{2}\right) \cdot \ldots \cdot \exp \left( \mathbf{Y}%
_{i}q_{i}\right) ,i=1,\ldots ,n.  \label{fi}
\end{equation}%
The expression (\ref{fi}) is known as the product of exponentials (POE),
which can be attributed to Brockett \cite{Brockett1984} as well as to Herv%
\'{e} \cite{Herve1982}. The POE is crucial for deriving kinematic relations
in a compact form. The linkage kinematics is encoded in the screw coordinate
vectors $\mathbf{Y}_{1},\ldots ,\mathbf{Y}_{n}$, where%
\begin{equation}
\mathbf{Y}_{j}=\left( 
\begin{array}{c}
\mathbf{e}_{j} \\ 
\mathbf{y}_{j}\times \mathbf{e}_{j}+h_{j}\mathbf{e}_{j}%
\end{array}%
\right) \in {\mathbb{R}}^{6}  \label{Yi}
\end{equation}%
is the screw coordinate vector (ray coordinates) associated with joint $j$
in the reference configuration $\mathbf{q}=\mathbf{0}$, represented in $%
\mathcal{F}_{0}$. Therein, $\mathbf{e}_{j}\in {\mathbb{R}}^{3}$ is a unit
vector along the joint axis, $\mathbf{y}_{j}\in {\mathbb{R}}^{3}$ is a
position vector to a (any) point on that axis (fig. \ref{figJointScrew}).
Both vectors are resolved in $\mathcal{F}_{0}$. The pitch $h$ of the joint
determines the amount of translation along the joint axis per rotation.
Revolute and prismatic joints are special cases, with $h=0$ and \thinspace $%
h=\infty $, respectively.

The advantage of the POE formula is that it does not require modeling of the
relative inter-body transformations, which is a complicated step in the
classical kinematics formulations, of which the Denavit-Hartenberg
convention \cite{DenavitHartenberg1955,KhalilKleinfinger1986} is best known,
including two-frame conventions, e.g. \cite{ShethUicker1971}. A salient
feature of the POE formula is that it allows for a kinematic description
without the need to introduce separate body-fixed joint frames since all
joint screw coordinates $\mathbf{Y}_{i}$ are expressed in the global frame $%
\mathcal{F}_{0}$.

\begin{remark}
In the above formulation of the KM all body-fixed reference frames are
(implicitly) introduced such that, in the reference configuration $\mathbf{q}%
=\mathbf{0}$, they coincide with the global frame $\mathcal{F}_{0}$ (since $%
f_{i}\left( \mathbf{0}\right) =\mathbf{I}$). In many applications, a
dedicated reference frame $\mathcal{F}_{i}$ at body $i$ is used. Denote with 
$\mathbf{A}_{i}\in SE\left( 3\right) $ the configuration of the body-fixed
frame $\mathcal{F}_{i}$ relative to $\bar{\mathcal{F}}_{i}$, then the
spatial configuration of $\mathcal{F}_{i}$ is given as $\mathbf{C}_{i}=\bar{%
\mathbf{C}}_{i}\mathbf{A}_{i}$, and thus determined by the KM as $\mathbf{C}%
_{i}=f_{i}\left( \mathbf{q}\right) \mathbf{A}_{i}$. For simplicity, this
constant transformation will be omitted, and the basic formulation (\ref{fi}%
) will be used in this paper.
\end{remark}

\begin{remark}
Modeling mechanisms using 1-DOF joints has been the standard approach for
MBS modeling \cite{WittenburgBook} while advanced formulations explicitly
use higher-DOF joints \cite{Uicker2013}. It should be noted, however, that
all lower pair (and several higher pair) joints can be modeled as
combination of 1-DOF joints. The formulations in this paper assume 1-DOF\
joints but can be extended to multi-DOF joints. This is not shown the for
sake of compactness.
\end{remark}

\section{Spatial Twists and the Geometric Jacobian}

The twist (velocity screw) of a rigid body is the aggregate of its angular
velocity and the translational velocity of a reference point. It thus
depends on the selection of a reference point. Its coordinate representation
further depends on the frame in which the velocity vectors are resolved.

The \emph{spatial twist} of link $i$ is represented by the vector $\mathbf{V}%
_{i}^{\text{s}}=\left( 
\bm{\omega}%
_{i}^{\text{s}},\mathbf{v}_{i}^{\text{s}}\right) \in {\mathbb{R}}^{6}$,
where $%
\bm{\omega}%
_{i}^{\text{s}}$ is the angular velocity of the body relative to $\mathcal{F}%
_{0}$ resolved in $\mathcal{F}_{0}$, and $\mathbf{v}_{i}^{\text{s}}=\dot{%
\mathbf{r}}_{i}-%
\bm{\omega}%
_{i}^{\text{s}}\times \mathbf{r}_{i}$ is the translational velocity of the
(possibly imaginary) point in the body that is momentarily traveling through
the origin of the global frame measured and resolved in $\mathcal{F}_{0}$.

Denote with $\mathbf{u}_{j}\left( \mathbf{q}\right) $ a unit vector along
the current axis of joint $j$, and with $\mathbf{s}_{i}\left( \mathbf{q}%
\right) $ the position vector from the origin of $\mathcal{F}_{0}$ to any
point on the current axis, both resolved in $\mathcal{F}_{0}$. The spatial
twist of body $i$ is readily constructed as \cite{Angeles2007,MUBOScrews1}%
\begin{eqnarray}
\mathbf{V}_{i}^{\text{s}} &=&\dot{q}_{1}\left( 
\begin{array}{c}
\mathbf{u}_{1} \\ 
\mathbf{s}_{1}\times \mathbf{u}_{1}+h_{1}\mathbf{u}_{1}%
\end{array}%
\right) +\dot{q}_{2}\left( 
\begin{array}{c}
\mathbf{u}_{2} \\ 
\mathbf{s}_{2}\times \mathbf{u}_{2}+h_{2}\mathbf{u}_{2}%
\end{array}%
\right) +\ldots +\dot{q}_{i}\left( 
\begin{array}{c}
\mathbf{u}_{i} \\ 
\mathbf{s}_{i}\times \mathbf{u}_{i}+h_{i}\mathbf{u}_{i}%
\end{array}%
\right)  \notag \\
&=&\dot{q}_{1}\mathbf{S}_{1}+\dot{q}_{2}\mathbf{S}_{2}+\ldots +\dot{q}_{i}%
\mathbf{S}_{i}  \label{Vsi} \\
&=&\mathbf{J}_{i}^{\mathrm{s}}\left( \mathbf{q}\right) \dot{\mathbf{q}}
\label{VsiJ}
\end{eqnarray}%
where $h_{j}$ is the pitch of joint $j$, and 
\begin{equation}
\mathbf{S}_{j}:=\left( 
\begin{array}{c}
\mathbf{u}_{j} \\ 
\mathbf{s}_{j}\times \mathbf{u}_{j}+h_{j}\mathbf{u}_{j}%
\end{array}%
\right)  \label{Sigeom}
\end{equation}%
is the \emph{instantaneous screw coordinate} vector in \emph{spatial
representation} associated to joint $j$. The latter constitute the columns
of the \emph{geometric (spatial) Jacobian} of body $i$ 
\begin{equation}
\mathbf{J}_{i}^{\mathrm{s}}\left( \mathbf{q}\right) :=%
\Big%
(\mathbf{S}_{1}%
\hspace{-0.5ex}%
\left( \mathbf{q}\right) 
\Big%
|\cdots 
\Big%
|\mathbf{S}_{i}%
\hspace{-0.5ex}%
\left( \mathbf{q}\right) 
\Big%
|\mathbf{0}%
\Big%
|\cdots 
\Big%
|\mathbf{0}%
\Big%
)  \label{Ji}
\end{equation}

The instantaneous joint screw coordinates (\ref{Sigeom}) are obtained
analytically by transforming the screw coordinates $\mathbf{Y}_{j}$, in the
zero reference configuration, to the current configuration of body $j$
according to $\bar{\mathbf{C}}_{j}=f_{j}\left( \mathbf{q}\right) $ 
\begin{equation}
\mathbf{S}_{j}%
\hspace{-0.5ex}%
\left( \mathbf{q}\right) =\mathbf{Ad}_{f_{j}\left( \mathbf{q}\right) }%
\mathbf{Y}_{j},\ j\leq i  \label{Si}
\end{equation}%
with the Ad mapping in (\ref{Ad}). The latter is the $6\times 6$ matrix
transforming screw coordinate vectors. In the reference configuration $%
\mathbf{q}=\mathbf{0}$, the $\mathbf{S}_{j}$ are equal to $\mathbf{Y}_{j}$
in (\ref{Yi}), i.e. $\mathbf{s}_{j}\left( \mathbf{0}\right) =\mathbf{y}_{j}$
and $\mathbf{u}_{j}\left( \mathbf{0}\right) =\mathbf{e}_{j}$. Application of
the identity $\mathbf{Ad}_{\exp \left( \mathbf{Y}_{j}q_{j}\right) }\mathbf{Y}%
_{j}=\mathbf{Y}_{j}$ shows that $\mathbf{S}_{i}$ depends on the joint
variables $q_{1,}\ldots ,q_{i-1}$, and the expression (\ref{Si}) simplifies
to $\mathbf{S}_{j}%
\hspace{-0.5ex}%
\left( \mathbf{q}\right) =\mathbf{Ad}_{f_{j-1}\left( \mathbf{q}\right) }%
\mathbf{Y}_{j}$, for $j>1$. Consequently, the screw coordinate vector of the
first joint in spatial representation is constant: $\mathbf{S}_{1}\equiv 
\mathbf{Y}_{1}$.

From (\ref{Vsi}) follows immediately the recursive relation%
\begin{equation}
\mathbf{V}_{i}^{\text{s}}=\mathbf{V}_{i-1}^{\text{s}}+\mathbf{S}_{i}\dot{q}%
_{i},i=1,\ldots ,n  \label{Vsirec}
\end{equation}%
with $\mathbf{V}_{0}^{\text{s}}=\mathbf{0}$, which could be regarded as the
twist of the ground.

\begin{remark}
The definition of spatial twist may seem unusual. However, its advantage is
that the twists of all bodies, and the relative twists due to joint motions,
are represented in one common spatially fixed frame $\mathcal{F}_{0}$ (hence
the name) so that the twists of individual bodies can simply be added
without the need for a frame transformation of twists. The spatial
representation prevails in kinematics and mechanism theory, but it is also
increasingly used in multibody dynamics \cite{Featherstone2008}. In many
applications, including motion planning and control, as well as classical
dynamics modeling, a body-fixed representation is used \cite%
{Angeles2007,LynchPark2017,MUBOScrews2,Uicker2013}. This will be briefly
discussed in sec. \ref{secRepresentations}.
\end{remark}

\section{Partial Derivatives of Joint Screw Coordinates in Spatial
Representation%
\label{secPartialS}%
}

The non-zero partial derivatives of the instantaneous joint screw
coordinates are \cite{LynchPark2017,MMTHighDer,Murray,Selig}%
\begin{eqnarray}
\frac{\partial \mathbf{S}_{i}}{\partial q_{j}} &=&[\mathbf{S}_{j},\mathbf{S}%
_{i}]  \notag \\
&=&\mathbf{ad}_{\mathbf{S}_{j}}\mathbf{S}_{i},\ j<i.  \label{derSi}
\end{eqnarray}%
Derivatives w.r.t. $q_{j},j\geq i$ vanish as $\mathbf{S}_{i}$ only depends
on $q_{j},j\leq i$ and $[\mathbf{S}_{i},\mathbf{S}_{i}]=\mathbf{0}$. The
relation (\ref{derSi}) gives rise to a compact expression for repeated
partial derivatives. The repeated partial derivative w.r.t. the $\nu $
variables $q_{\alpha _{1}},\ldots ,q_{\alpha _{\nu }}$ attains the closed
form \cite{MMTHighDer}%
\begin{eqnarray}
\frac{\partial ^{\nu }\mathbf{S}_{i}}{\partial q_{\alpha _{1}}\partial
q_{\alpha _{2}}\cdots \partial q_{\alpha _{\nu }}} &=&\left[ \mathbf{S}%
_{\beta _{1}},\left[ \mathbf{S}_{\beta _{2}},\left[ \mathbf{S}_{\beta
_{3}},\ldots \left[ \mathbf{S}_{\beta _{\nu }},\mathbf{S}_{i}\right] \ldots %
\right] \right] \right]  \notag \\
&=&\mathbf{ad}_{\mathbf{S}_{\beta _{1}}}\mathbf{ad}_{\mathbf{S}_{\beta _{2}}}%
\mathbf{ad}_{\mathbf{S}_{\beta _{3}}}\cdots \mathbf{ad}_{\mathbf{S}_{\beta
_{\nu }}}\mathbf{S}_{i},\ \ \ \text{if }\alpha _{1},\ldots ,a_{\nu }<i
\label{dnS}
\end{eqnarray}%
where $\beta _{1}\leq \beta _{2}\leq \beta _{3}\leq \ldots \leq \beta _{\nu
}<i$ is the ordered set of indexes $\{\alpha _{1},\ldots ,\alpha _{\nu }\}$
(which accounts for the fact that partial derivatives commute but the
indexes are generally not ordered).

The relation (\ref{dnS}) can be written using a multi-index $\mathbf{a}%
=\left( a_{1},a_{2},\ldots ,a_{n}\right) \in \mathbb{N}^{n}$, where $%
a_{j}=0,1,2,\ldots $ is the number of partial derivations w.r.t. $q_{j}$.
Then (\ref{dnS}) can be expressed in the compact form%
\begin{equation}
\partial ^{\mathbf{a}}\mathbf{S}_{i}=\mathbf{ad}_{\mathbf{S}_{a_{1}}}^{a_{1}}%
\mathbf{ad}_{\mathbf{S}_{a_{2}}}^{a_{2}}\cdots \mathbf{ad}_{\mathbf{S}%
_{a_{i-1}}}^{a_{i-1}}\mathbf{S}_{i}=\prod\limits_{1\leq j\leq n}\mathbf{ad}_{%
\mathbf{S}_{j}}^{a_{j}}\mathbf{S}_{i}  \label{dnS2}
\end{equation}%
and $\partial ^{\mathbf{a}}\mathbf{S}_{i}=\mathbf{0}$ if $a_{j}\neq 0$ for
some $j\geq i$, where the ordered (right) matrix product (\ref{multiProd})
is used. With $=\partial ^{\mathbf{a}_{i-1}}\mathbf{S}_{i},\ $for $%
a_{i}=\ldots =a_{n}=0$, this can be written as 
\begin{equation}
\partial ^{\mathbf{a}}\mathbf{S}_{i}=\partial ^{\mathbf{a}_{i-1}}\mathbf{S}%
_{i},\ \text{for }a_{i}=\ldots =a_{n}=0.
\end{equation}

As an example, the partial derivatives of $\mathbf{S}_{4}$ are%
\begin{equation*}
\frac{\partial ^{3}}{\partial q_{2}^{3}}\frac{\partial }{\partial q_{1}}%
\frac{\partial ^{2}}{\partial q_{3}^{2}}\mathbf{S}_{4}=[\mathbf{S}_{1},[%
\mathbf{S}_{2},[\mathbf{S}_{2},[\mathbf{S}_{2},[\mathbf{S}_{3},[\mathbf{S}%
_{3},\mathbf{S}_{4}]]]]]]=\mathbf{ad}_{\mathbf{S}_{1}}\mathbf{ad}_{\mathbf{S}%
_{2}}^{3}\mathbf{ad}_{\mathbf{S}_{3}}^{2}\mathbf{S}_{4},\ \ \ \frac{\partial 
}{\partial q_{4}}\mathbf{S}_{4}=\mathbf{0}.
\end{equation*}

The expression (\ref{derSi}) for the partial derivative has been reported in 
\cite{KargerNovak1985,Karger1989} and later e.g. in \cite%
{Murray,Lerbet1999,Rico1999}. A geometric derivation of the partial
derivative of a screw can be found in \cite%
{Rico1999,Cervantes2004,Cervantes2009}, which provides insight into the
kinematic meaning of the Lie bracket in this context. Relations for partial
derivatives up to 3rd order were presented in \cite{Karger1996}. The
relation (\ref{dnS}) for the higher-order partial derivatives was first
presented in \cite{MMTHighDer}.

\section{Time Derivatives of Twists and Joint Screws%
\label{secTimeDerS}%
}

\subsection{Closed form expressions for lower degree derivatives}

The spatial twist of body $i$ is given by (\ref{Vsi}), and the time
derivatives of the twist $\mathbf{V}_{i}^{\text{s}}$ can be written
explicitly with the expressions (\ref{dnS}) for partial derivatives. For
instance the acceleration and jerk of body $i$ is respectively%
\begin{eqnarray}
\dot{\mathbf{V}}_{i}^{\text{s}} &=&\sum_{j\leq i}\mathbf{S}_{j}\ddot{q}%
_{j}+\sum_{k<j\leq i}\left[ \mathbf{S}_{k},\mathbf{S}_{j}\right] \dot{q}_{j}%
\dot{q}_{k}  \label{Vsidot} \\
\ddot{\mathbf{V}}_{i}^{\text{s}} &=&\sum_{j\leq i}\mathbf{S}_{j}\dddot{q}%
_{j}+2\sum_{k<j\leq i}\left[ \mathbf{S}_{k},\mathbf{S}_{j}\right] \dot{q}_{k}%
\ddot{q}_{j}+\sum_{k<j\leq i}\left[ \mathbf{S}_{k},\mathbf{S}_{j}\right] 
\ddot{q}_{k}\dot{q}_{j}+\sum_{l<k<j\leq i}\left[ [\mathbf{S}_{l},\mathbf{S}%
_{k}\right] ,\mathbf{S}_{j}]\dot{q}_{l}\dot{q}_{k}\dot{q}_{j}+\sum_{l,k<j%
\leq i}\left[ \mathbf{S}_{k},[\mathbf{S}_{l},\mathbf{S}_{j}\right] ]\dot{q}%
_{l}\dot{q}_{k}\dot{q}_{j}  \notag \\
&=&\sum_{j\leq i}\mathbf{S}_{j}\dddot{q}_{j}+2\sum_{l<k<j\leq i}\left[ 
\mathbf{S}_{l},\left[ \mathbf{S}_{k},\mathbf{S}_{j}\right] \right] \dot{q}%
_{l}\dot{q}_{k}\dot{q}_{j}+\sum_{k<j\leq i}%
\Big%
(\left[ \mathbf{S}_{k},\mathbf{S}_{j}\right] \left( \ddot{q}_{k}\dot{q}_{j}+2%
\dot{q}_{k}\ddot{q}_{j}\right) +\left[ \mathbf{S}_{k},[\mathbf{S}_{k},%
\mathbf{S}_{j}]\right] \dot{q}_{k}^{2}\ddot{q}_{j}%
\Big%
).  \label{Vsi2dot}
\end{eqnarray}%
The final form of (\ref{Vsi2dot}) is obtained using $\left[ \mathbf{S}_{j},%
\left[ \mathbf{S}_{l},\mathbf{S}_{k}\right] \right] =-\left[ \mathbf{S}_{k},%
\left[ \mathbf{S}_{j},\mathbf{S}_{l}\right] \right] -\left[ \mathbf{S}_{l},%
\left[ \mathbf{S}_{k},\mathbf{S}_{j}\right] \right] $, obtained with the
Jacobi identity (\ref{JacobiIdent}), and the skew symmetry of the Lie
bracket for summation range $l<k<j\leq i$. The manipulation of these
relations gets involved for higher orders. This can be avoided using
recursive expressions as derived in the next section.

\begin{remark}
From (\ref{Vsidot}) it is clear that the time derivatives of the spatial
twist are in fact screws \cite{Lipkin2005}, i.e. elements of $se\left(
3\right) $, since they are given in terms of Lie brackets. Therefore $\dot{%
\mathbf{V}}_{i}^{\text{s}}$ is called the 'acceleration motor' \cite%
{Brand1947} (see page 127) or the 'reduced acceleration' \cite{RicoDuffy1996}%
.
\end{remark}

The equation (\ref{Vsidot}) for the acceleration was presented in \cite%
{RicoDuffy1996}. A version of the relation (\ref{Vsi2dot}) for the jerk was
reported in \cite{Rico1999,GallardoRico2001,Gallardo2008}. An explicit
relation for the 4th time derivative (the jounce) was reported in \cite%
{LopezCustodio2017}. The relations (\ref{Vsidot}) and (\ref{Vsi2dot}) for
were also presented in \cite{Lerbet1999}.

The complexity of the relation for the $k$th-degree derivative $\mathrm{D}%
^{(k)}\mathbf{V}_{i}^{\text{s}}$ is of order $O\left( i^{k+1}\right) $.
Closed form expressions for derivatives of any degree can yet be derived
using the explicit relations (\ref{dnS}). However, the number of different
summations grows rapidly with the degree, and finding simple closed form
relations becomes very difficult.

\subsection{Derivatives of arbitrary degree using recursive relations for
time derivatives of instantaneous joint screws}

The use of complex expressions can be avoided by means of a recursive
formulation where the $k$th time derivative is expressed in terms of time
derivatives of degree up to $k-1$ \cite{MMTConstraints}. To this end,
introduce%
\begin{equation}
\mathsf{S}_{i}(\mathbf{q},\dot{\mathbf{q}}):=\sum_{j\leq i}\mathbf{S}%
_{j}\left( \mathbf{q}\right) \dot{q}_{j}.  \label{SSi}
\end{equation}%
The expression for the velocity (\ref{Vsi}) then becomes $\mathbf{V}_{i}=%
\mathsf{S}_{i}\left( \mathbf{q},\dot{\mathbf{q}}\right) $. The $k$th time
derivative thus amounts to evaluating $\mathrm{D}^{(k)}\mathbf{V}_{i}=%
\mathrm{D}^{(k)}\mathsf{S}_{i}(\mathbf{q},\dot{\mathbf{q}})$. With (\ref{SSi}%
), the $k$th time derivative of $\mathsf{S}_{i}$ is 
\begin{equation}
\mathrm{D}^{(k)}\mathsf{S}_{i}=\sum_{j\leq i}\sum_{l=0}^{k}\tbinom{k}{l}%
\mathrm{D}^{(l)}\mathbf{S}_{j}q_{j}^{(k-l+1)}  \label{DSik}
\end{equation}%
which involves time derivatives of the instantaneous joint screws. The first
derivative follows from (\ref{derSi}), along with (\ref{SSi}), as%
\begin{equation}
\dot{\mathbf{S}}_{i}=\sum_{j\leq i}[\mathbf{S}_{j},\mathbf{S}_{i}]\dot{q}%
_{j}=[\sum_{j\leq i}\mathbf{S}_{j},\mathbf{S}_{i}]\dot{q}_{j}=[\mathsf{S}%
_{i},\mathbf{S}_{i}]  \label{Sidot}
\end{equation}%
Higher derivatives are obtained, by noting that $\frac{\partial }{\partial
q_{i}}\left[ \mathbf{X},\mathbf{Y}\right] =[\frac{\partial }{\partial q_{i}}%
\mathbf{X},\mathbf{Y}]+[\mathbf{X},\frac{\partial }{\partial q_{i}}\mathbf{Y]%
}$, as%
\begin{equation}
\mathrm{D}^{(k)}\mathbf{S}_{i}=\sum_{l=0}^{k-1}\tbinom{k-1}{l}[\mathrm{D}%
^{(l)}\mathsf{S}_{i},\mathrm{D}^{(k-l-1)}\mathbf{S}_{i}].  \label{DSi}
\end{equation}%
The recursive relation (\ref{DSik}) for the $k$-th time derivative of $%
\mathsf{S}_{i}\left( \mathbf{q},\dot{\mathbf{q}}\right) $ involves time
derivatives of the screw coordinates $\mathbf{S}_{j},j\leq i$ of preceding
joints in the kinematic chain up to degree $k$. The latter in turn involve
derivatives of $\mathsf{S}_{i}$ and $\mathbf{S}_{i}$ up to degree $k-1$,
according to (\ref{DSi}). The relations (\ref{DSi}) and (\ref{DSik}) thus
allow for a recursive symbolic construction, respectively recursive
(numerical) evaluation, of the time derivatives of the instantaneous joint
screw coordinates $\mathbf{S}_{i}$, and hence of the time derivatives of $%
\mathsf{S}_{i}$, i.e. of the twist $\mathbf{V}_{i}$. The advantage of this
recursive form is that it is easy to implement and accounts for arbitrary
derivatives, in contrast to explicit relations, such as (\ref{Vsidot}) and (%
\ref{Vsi2dot}). The complexity indeed remains $O\left( i^{k+1}\right) $.

\subsection{Explicit recursive relations for lower degree derivatives}

In various applications low-degree derivatives are required only, and it may
be desirable to use explicit forms of the recursive relations. The time
derivatives (\ref{DSi}) of the instantaneous joint screws of order $k=1,2,3$
are for instance%
\begin{eqnarray}
\dot{\mathbf{S}}_{i} &=&[\mathbf{V}_{i}^{\text{s}},\mathbf{S}_{i}]=\mathbf{ad%
}_{\mathbf{V}_{i}^{\text{s}}}\mathbf{S}_{i} \\
\ddot{\mathbf{S}}_{i} &=&[\dot{\mathbf{V}}_{i}^{\text{s}},\mathbf{S}_{i}]+[%
\mathbf{V}_{i}^{\text{s}},[\mathbf{V}_{i}^{\text{s}},\mathbf{S}_{i}]]=%
\big%
(\mathbf{ad}_{\dot{\mathbf{V}}_{i}^{\text{s}}}+\mathbf{ad}_{\mathbf{V}_{i}^{%
\text{s}}}^{2}%
\big%
)\mathbf{S}_{i} \\
\dddot{\mathbf{S}}_{i} &=&[\ddot{\mathbf{V}}_{i}^{\text{s}},\mathbf{S}%
_{i}]+2[\dot{\mathbf{V}}_{i}^{\text{s}},[\mathbf{V}_{i}^{\text{s}},\mathbf{S}%
_{i}]]+[\mathbf{V}_{i}^{\text{s}},[\dot{\mathbf{V}}_{i}^{\text{s}},\mathbf{S}%
_{i}]]+[\mathbf{V}_{i}^{\text{s}},[\mathbf{V}_{i}^{\text{s}},[\mathbf{V}%
_{i}^{\text{s}},\mathbf{S}_{i}]]]  \notag \\
&=&%
\big%
(\mathbf{ad}_{\ddot{\mathbf{V}}_{i}^{\text{s}}}+2\mathbf{ad}_{\dot{\mathbf{V}%
}_{i}^{\text{s}}}\mathbf{ad}_{\mathbf{V}_{i}^{\text{s}}}+\mathbf{ad}_{%
\mathbf{V}_{i}^{\text{s}}}\mathbf{ad}_{\dot{\mathbf{V}}_{i}^{\text{s}}}+%
\mathbf{ad}_{\mathbf{V}_{i}^{\text{s}}}^{3}%
\big%
)\mathbf{S}_{i}.
\end{eqnarray}%
Inserting them into (\ref{DSik}) yields%
\begin{eqnarray}
\dot{\mathbf{V}}_{i}^{\text{s}} &=&\sum_{j\leq i}\left( \mathbf{S}_{j}\ddot{q%
}_{j}+[\mathbf{V}_{j}^{\text{s}},\mathbf{S}_{j}]\dot{q}_{j}\right)
=\sum_{j\leq i}%
\big%
(\ddot{q}_{j}\mathbf{I}+\dot{q}_{j}\mathbf{ad}_{\mathbf{V}_{j}^{\text{s}}}%
\big%
)\mathbf{S}_{j}  \label{Vsidot2} \\
\ddot{\mathbf{V}}_{i}^{\text{s}} &=&\sum_{j\leq i}\left( \mathbf{S}_{j}%
\dddot{q}%
\hspace{-0.5ex}%
_{j}+2[\mathbf{V}_{j}^{\text{s}},\mathbf{S}_{j}]\ddot{q}_{j}+\left( [\dot{%
\mathbf{V}}_{j}^{\text{s}},\mathbf{S}_{j}]+[\mathbf{V}_{j}^{\text{s}},[%
\mathbf{V}_{j}^{\text{s}},\mathbf{S}_{j}]]\right) \dot{q}_{j}\right)
\label{Vsi2dot2} \\
&=&\sum_{j\leq i}%
\big%
(\dddot{q}_{%
\hspace{-0.5ex}%
j}\mathbf{I}+2\ddot{q}_{j}\mathbf{ad}_{\mathbf{V}_{j}^{\text{s}}}+\dot{q}%
_{j}(\mathbf{ad}_{\dot{\mathbf{V}}_{j}^{\text{s}}}+\mathbf{ad}_{\mathbf{V}%
_{j}^{\text{s}}}^{2})%
\big%
)\mathbf{S}_{j}  \notag \\
\dddot{\mathbf{V}}_{i}^{\text{s}} &=&\sum_{j\leq i}%
\big%
(\mathbf{S}_{j}\ddot{\ddot{q}}_{j}+3[\mathbf{V}_{j}^{\text{s}},\mathbf{S}%
_{j}]\dddot{q}_{j}+3\left( [\dot{\mathbf{V}}_{j}^{\text{s}},\mathbf{S}_{j}]+[%
\mathbf{V}_{j}^{\text{s}},[\mathbf{V}_{j}^{\text{s}},\mathbf{S}_{j}]]\right) 
\ddot{q}_{j}  \notag \\
&&+\left( [\ddot{\mathbf{V}}_{j}^{\text{s}},\mathbf{S}_{j}]+2[\dot{\mathbf{V}%
}_{j}^{\text{s}},[\mathbf{V}_{j}^{\text{s}},\mathbf{S}_{j}]+[\mathbf{V}_{j}^{%
\text{s}},[\dot{\mathbf{V}}_{j}^{\text{s}},\mathbf{S}_{j}]]+[\mathbf{V}_{j}^{%
\text{s}},[\mathbf{V}_{j}^{\text{s}},[\mathbf{V}_{j}^{\text{s}},\mathbf{S}%
_{j}]]]\right) \dot{q}_{j}%
\Big%
)  \label{Vsi3dot2} \\
&=&\sum_{j\leq i}%
\big%
(\ddot{\ddot{q}}_{j}\mathbf{I}+3\dddot{q}_{j}\mathbf{ad}_{\mathbf{V}_{j}^{%
\text{s}}}+3\ddot{q}_{j}(\mathbf{ad}_{\dot{\mathbf{V}}_{j}^{\text{s}}}+%
\mathbf{ad}_{\mathbf{V}_{j}^{\text{s}}}^{2})+\dot{q}_{j}(\mathbf{ad}_{\ddot{%
\mathbf{V}}_{j}^{\text{s}}}+2\mathbf{ad}_{\dot{\mathbf{V}}_{j}^{\text{s}%
}}^{2}+\mathbf{ad}_{\mathbf{V}_{j}^{\text{s}}}\mathbf{ad}_{\dot{\mathbf{V}}%
_{j}^{\text{s}}}+\mathbf{ad}_{\mathbf{V}_{j}^{\text{s}}}^{3})%
\big%
)\mathbf{S}_{j}.  \notag
\end{eqnarray}%
These are recursive relations for the $k$th time derivative of $\mathbf{V}%
_{i}^{\text{s}}$ in terms of derivatives of twists of preceding bodies up to
order $k-1$ and the time derivatives of $\mathbf{q}\left( t\right) $, so
that the nested summations in (\ref{Vsidot}) and (\ref{Vsi2dot}) etc. are
avoided. Due to the use of spatial twists, the summation for body $i-1$ is
repeated in the summation for body $i$. Reusing the repeated term yields the
recursive relations%
\begin{eqnarray}
\mathbf{V}_{i}^{\text{s}} &=&\mathbf{V}_{i-1}^{\text{s}}+\mathbf{S}_{i}\dot{q%
}_{i}  \label{Vsrec} \\
\dot{\mathbf{V}}_{i}^{\text{s}} &=&\dot{\mathbf{V}}_{i-1}^{\text{s}}+\mathbf{%
S}_{i}\ddot{q}_{i}+[\mathbf{V}_{i}^{\text{s}},\mathbf{S}_{i}]\dot{q}_{i}=%
\dot{\mathbf{V}}_{i-1}^{\text{s}}+\left( \ddot{q}_{i}\mathbf{I}+\dot{q}_{i}%
\mathbf{ad}_{\mathbf{V}_{i}^{\text{s}}}\right) \mathbf{S}_{i}
\label{Vsdotrec} \\
\ddot{\mathbf{V}}_{i}^{\text{s}} &=&\ddot{\mathbf{V}}_{i-1}^{\text{s}}+%
\mathbf{S}_{i}\dddot{q}_{i}+2[\mathbf{V}_{i}^{\text{s}},\mathbf{S}_{i}]\ddot{%
q}_{i}+\left( [\dot{\mathbf{V}}_{i}^{\text{s}},\mathbf{S}_{i}]+[\mathbf{V}%
_{i}^{\text{s}},[\mathbf{V}_{i}^{\text{s}},\mathbf{S}_{i}]]\right) \dot{q}%
_{i}  \label{Vs2dotrec} \\
&=&\ddot{\mathbf{V}}_{i-1}^{\text{s}}+\left( \dddot{q}_{i}\mathbf{I}+2\ddot{q%
}_{i}\mathbf{ad}_{\mathbf{V}_{i}^{\text{s}}}+\dot{q}_{i}(\mathbf{ad}_{\dot{%
\mathbf{V}}_{i}^{\text{s}}}+\mathbf{ad}_{\mathbf{V}_{i}^{\text{s}%
}}^{2})\right) \mathbf{S}_{i}  \notag \\
\dddot{\mathbf{V}}_{i}^{\text{s}} &=&\dddot{\mathbf{V}}_{i-1}^{\text{s}}+%
\mathbf{S}_{i}\ddot{\ddot{q}}_{i}+3[\mathbf{V}_{i}^{\text{s}},\mathbf{S}_{i}]%
\dddot{q}_{%
\hspace{-0.5ex}%
i}+3\left( [\dot{\mathbf{V}}_{i}^{\text{s}},\mathbf{S}_{i}]+[\mathbf{V}_{i}^{%
\text{s}},[\mathbf{V}_{i}^{\text{s}},\mathbf{S}_{i}]]\right) \ddot{q}_{i}
\label{Vs3dotrec} \\
&&+\left( [\ddot{\mathbf{V}}_{i}^{\text{s}},\mathbf{S}_{i}]+2[\dot{\mathbf{V}%
}_{i}^{\text{s}},[\mathbf{V}_{i}^{\text{s}},\mathbf{S}_{i}]]+[\mathbf{V}%
_{i}^{\text{s}},[\dot{\mathbf{V}}_{i}^{\text{s}},\mathbf{S}_{i}]]+[\mathbf{V}%
_{i}^{\text{s}},[\mathbf{V}_{i}^{\text{s}},[\mathbf{V}_{i}^{\text{s}},%
\mathbf{S}_{i}]]]\right) \dot{q}_{i}  \notag \\
&=&\dddot{\mathbf{V}}_{i-1}^{\text{s}}+\left( \ddot{\ddot{q}}_{i}\mathbf{I}+3%
\dddot{q}_{%
\hspace{-0.5ex}%
i}\mathbf{ad}_{\mathbf{V}_{i}^{\text{s}}}+3\ddot{q}_{i}(\mathbf{ad}_{\dot{%
\mathbf{V}}_{i}^{\text{s}}}+\mathbf{ad}_{\mathbf{V}_{i}^{\text{s}}}^{2})+%
\dot{q}_{i}(\mathbf{ad}_{\ddot{\mathbf{V}}_{i}^{\text{s}}}+2\mathbf{ad}_{%
\dot{\mathbf{V}}_{i}^{\text{s}}}\mathbf{ad}_{\mathbf{V}_{i}^{\text{s}}}+%
\mathbf{ad}_{\mathbf{V}_{i}^{\text{s}}}\mathbf{ad}_{\dot{\mathbf{V}}_{i}^{%
\text{s}}}+\mathbf{ad}_{\mathbf{V}_{i}^{\text{s}}}^{3})\right) \mathbf{S}%
_{i}.  \notag
\end{eqnarray}%
The order of complexity of the recursive relations (\ref{Vsrec}-\ref%
{Vs3dotrec}) for the $k$th-order derivative $\mathrm{D}^{(k)}\mathbf{V}_{i}^{%
\text{s}}$ is $O\left( i\right) $. These recursive $O\left( i\right) $
relations can be efficiently implemented by avoiding repeated matrix
multiplication and reusing vector terms like $\mathbf{ad}_{\mathbf{V}_{i}^{%
\text{s}}}\mathbf{S}_{i}$ etc.

\subsection{Applications}

\subparagraph{a) Higher-order forward kinematics of a robotic arm%
\label{secFWKin1}%
}

A robotic arm is a serial kinematic chain where an EE is attached at the
terminal link $n$. The forward kinematic mapping $f_{n}$ in (\ref{fi})
determines the EE configuration. On velocity level, the forward kinematic
problem is to find the EE twist $\mathbf{V}_{n}^{\mathrm{s}}$ for given
state $\left( \mathbf{q},\dot{\mathbf{q}}\right) \in T{\mathbb{V}}^{n}$,
i.e. to evaluate (\ref{VsiJ}). The corresponding higher-order problem is to
compute the time derivatives of $\mathbf{V}_{n}^{\mathrm{s}}$ for given
derivatives of the joint variables $\mathbf{q}$. This is necessary for
smooth motion planning when, in addition to velocity and acceleration, also
limits on the jerk, jounce and possibly higher derivative, must be respected.

The closed form relations (\ref{Vsidot}) and (\ref{Vsi2dot}) can be used to
determine the EE acceleration and jerk. Alternatively, the time derivatives
of the EE twist of arbitrary degree can be determined with the recursive
relations (\ref{DSi}) and (\ref{DSik}), and in particular with (\ref{Vsrec}%
)-(\ref{Vs3dotrec}) up to 3rd degree. These recursive approach yield the
derivatives of the twists of all bodies.

\begin{remark}
The formulation in terms of spatial representation of twists seems
computationally advantageous since it does not involve frame
transformations, as apparent from (\ref{Vsrec})-(\ref{Vs3dotrec}) (twists of
the preceding bodies are simply added). This has been confirmed in \cite%
{OrinSchrader1984} where the recursive formulations for the velocity (i.e.
first-order) forward kinematics using spatial, body-fixed, and hybrid twists
were compared. Such an analysis comparing the higher-order formulation in
spatial and body-fixed representation (sec. \ref{secBodyFixedRep}) does not
yet exist.
\end{remark}

\subparagraph{b) Higher-order inverse dynamics of a robotic arm%
\label{secInvDyn}%
}

The inverse dynamics problem is to determine the generalized forces $\mathbf{%
Q}$ in the equations of motion (EOM)%
\begin{equation}
\mathbf{M}\left( \mathbf{q}\right) \ddot{\mathbf{q}}+\mathbf{g}\left( \dot{%
\mathbf{q}},\mathbf{q}\right) =\mathbf{Q}  \label{EOM}
\end{equation}%
of a robotic arm and to relate them to the actuator forces. This simply
requires evaluating the left hand side of (\ref{EOM}) for given $\mathbf{q}%
\left( t\right) $. Since this evaluation is time critical, efficient
recursive $O\left( n\right) $ inverse dynamics algorithms were developed 
\cite%
{AndersonCritchley2003,Angeles2007,Bae2001,Featherstone1983,Featherstone2008,Fijany1995,Hollerbach1980,ParkBobrowPloen1995,Rodriguez1987,Rodriguez1991,Rodriguez1992}%
. Any such $O\left( n\right) $ algorithm involves a recursive forward
kinematics run where the configurations, twists, and accelerations of all
bodies are determined from given $\mathbf{q}$, $\dot{\mathbf{q}}$, and $%
\ddot{\mathbf{q}}$ (which solves the forward kinematics problem of the
linkage). The number of operations thus depends on the number and type of
frame transformations involved. As apparent from (\ref{Vsrec}), no frame
transformations are required when using the spatial representation of
twists, and likewise wrenches (the twists $\mathbf{V}_{i}^{\text{s}}$ and $%
\mathbf{V}_{i-1}^{\text{s}}$ of body $i$ and $i-1$ are simply added and
complemented by the contribution $\mathbf{S}_{i}\dot{q}_{i}$ of the
connecting joint). Therefore, the computationally most efficient $O\left(
n\right) $ algorithms use the spatial representation of twists such as the
Articulated-Body and the Composite-Rigid-Body algorithm \cite%
{Featherstone2008}, where (\ref{Vsrec}) and (\ref{Vsdotrec}) provide
relations in the forward kinematic run.

In various applications the time derivatives of the actuation forces $%
\mathbf{Q}$ are necessary. Two such applications are 1) the optimal control
taking into account technical limitations of the drives, i.e. on $\mathbf{Q}$
and $\dot{\mathbf{Q}}$ \cite{ConstantinescuCroft2000,ReiterTII2018}, and 2)
the flatness-based control of manipulators with series elastic actuators
(SEA), which requires the second time derivative of $\mathbf{Q}$ \cite%
{deLuca1998,PalliMelchiorriDeLuca2008,BuondonnaDeLuca2015,Giusti2018}. The
latter translates to the second time derivative of the EOM (\ref{EOM}),
which involves $\dddot{\mathbf{q}}$ and $\ddot{\ddot{\mathbf{q}}}$ (see
section \ref{secInvKinApp}a). Recursive $O\left( n\right) $ algorithms for
evaluating the first time derivative of the EOM (\ref{EOM}) have been
proposed in \cite{Guarino2006,Guarino2009}, and for the second second time
derivative in \cite{BuondonnaDeLuca2015,BuondonnaDeLuca2016}, where the
recursive forward kinematics run additionally determines the jerk and jounce
of all bodies, respectively. These use the body-fixed representation of
twists (see \ref{secBodyFixedRep}), and are formulated in terms of DH
parameters rather than the (more user-friendly) joint screw coordinates. A
formulation using body-fixed and hybrid representation of twists was
presented in \cite{ICRA2017} using the Lie group formulation in terms of
joint screws. Aiming at maximum efficiency an $O\left( n\right) $ algorithm
can be developed using the spatial representation. The necessary relations
for the higher-order forward kinematics recursion are given by (\ref{Vsrec}%
)-(\ref{Vs3dotrec}).

\subparagraph{c) Gradients of local dexterity measures of a robotic arm}

A serial manipulator is an open kinematic chain with an end-effector (EE)
attached at its terminal body $n$. In a given configuration $\mathbf{q}\in {%
\mathbb{V}}^{n}$ of the manipulator, the EE twist generated by joint
velocities is determined with (\ref{VsiJ}) as $\mathbf{V}_{n}^{\text{s}}=%
\mathbf{J}_{n}^{\mathrm{s}}\left( \mathbf{q}\right) \dot{\mathbf{q}}$. This
Jacobian serves as forward kinematics Jacobian, and is denoted with $\mathbf{%
J}$ for simplicity. Kinematic dexterity measures are used to assess the
kinematic dexterity, i.e. the transmission of joint velocities to EE twists
(respectively the transmission from EE wrenches to joint torques/forces) at
a given configuration. The two established local dexterity (or
manipulability) measures for (possibly kinematically redundant) serial
manipulators are \cite{Merlet,Tsai,Yoshikawa,Lee,Ma} are

\begin{equation}
\mu =\sqrt{\det (\mathbf{JJ}^{T})},\ \sigma =1/\kappa (\mathbf{JJ}^{T}),
\label{musigma}
\end{equation}%
where $\kappa \left( \mathbf{A}\right) =\left\Vert \mathbf{A}\right\Vert
\left\Vert \mathbf{A}^{-1}\right\Vert $ is the condition number of a
rectangular matrix $\mathbf{A}$. Dexterity measures are used to select
optimal poses and to find optimal designs of robots \cite%
{Abdel-Malek,Angeles1992,GosselinAngeles1991,Elkady,KleinBlaho1987}. Also
optimal motion planning aims to maximize dexterity, i.e. to find $\mathbf{q}$
such that the above measures are maximized \cite%
{Chiacchio1990,Chiu1988,MED2003}. Gradient-based methods are often used,
which necessitates the gradient (and possibly Hessian) w.r.t. the joint
variables, i.e. the first and second partial derivatives.

The partial derivative of the measure $\mu $ is $\frac{\partial }{\partial
q_{i}}\mu _{3}=\frac{1}{2}\frac{1}{\mu }\frac{\partial }{\partial q_{i}}\det
(\mathbf{JJ}^{T})$. Defining the $n\times n$ matrix, with columns $\mathbf{A}%
_{i}$, 
\begin{equation}
\mathbf{A}:=\mathbf{JJ}^{T}=\left( \mathbf{A}_{1}\left\vert \underset{\,\;}{%
\overset{\,\;}{\cdots }}\right\vert \mathbf{A}_{n}\right)  \label{AJJT}
\end{equation}
the partial derivative of $\mu $ can be expressed as%
\begin{equation}
\frac{\partial }{\partial q_{i}}\mu =\frac{1}{2}\frac{1}{\mu }%
\sum_{k=1}^{n}\det \left( \mathbf{A}_{1}\left\vert \underset{\,\;}{\overset{%
\,\;}{\cdots }}\right\vert \frac{\partial }{\partial q_{i}}\mathbf{A}%
_{k}\left\vert \underset{\,\;}{\overset{\,\;}{\cdots }}\right\vert \mathbf{A}%
_{n}\right) .  \label{dDet}
\end{equation}%
The partial derivative of the $k$th column of $\mathbf{A}$ is determined
with $\partial _{q_{i}}\mathbf{A}=\partial _{q_{i}}\mathbf{JJ}^{T}+\left(
\partial _{q_{i}}\mathbf{JJ}^{T}\right) ^{T}$. The partial derivatives $%
\partial _{q_{i}}\mathbf{J}$ of the Jacobian, defined in (\ref{Ji}), are
given algebraically in closed form with (\ref{derSi}). The closed form
expression (\ref{dDet}) applies to non-redundant as well as redundant serial
manipulators.

For a non-redundant robotic arm, $\mathbf{J}$ is a square $n\times n$
matrix. If this has full rank, i.e. $\mu \neq 0$, then (\ref{dDet}) can be
simplified to%
\begin{equation}
\frac{\partial }{\partial q_{i}}\mu =\mu \,\mathrm{tr}(\mathbf{J}%
^{-1}\partial _{q_{i}}\mathbf{J})=\mu 
\hspace{-1ex}%
\sum_{k=i+1}^{n}\overline{\mathbf{J}}_{k}^{T}\mathbf{ad}_{\mathbf{S}_{i}}%
\mathbf{S}_{k}  \label{derMu}
\end{equation}%
with $\overline{\mathbf{J}}_{k}$ being the $k$th row of $\mathbf{J}^{-1}$.
The last term in (\ref{derMu}) follows with (\ref{derSi}). Although this
formula involves the inverse Jacobian, it is computationally advantageous if
the latter is already known, e.g. from solving the inverse kinematics.

The condition number of the $n\times n$ matrix (\ref{AJJT}) can be expressed
with the spectral norm, which is defined as $\left\Vert \mathbf{A}%
\right\Vert _{2}=\sqrt{\sum_{i,j=1}^{n}A_{ij}^{2}}=\sqrt{\sum_{i=1}^{n}%
\mathbf{A}_{i}^{T}\mathbf{A}_{i}}$ . Its partial derivatives are $\frac{%
\partial }{\partial q_{i}}\left\Vert \mathbf{A}\right\Vert _{2}=\frac{1}{%
\left\Vert \mathbf{A}\right\Vert }\sum_{j=1}^{n}\frac{\partial }{\partial
q_{i}}\mathbf{A}_{j}^{T}\mathbf{A}_{j}$, and thus 
\begin{equation}
\frac{\partial }{\partial q_{i}}\kappa \left( \mathbf{A}\right) =\frac{%
\left\Vert \mathbf{A}\right\Vert _{2}}{\left\Vert \mathbf{A}^{-1}\right\Vert
_{2}}\sum_{j=1}^{n}\frac{\partial }{\partial q_{i}}\mathbf{A}_{j}^{-T}%
\mathbf{A}_{j}^{-1}+\frac{\left\Vert \mathbf{A}^{-1}\right\Vert _{2}}{%
\left\Vert \mathbf{A}\right\Vert _{2}}\sum_{i=1}^{n}\frac{\partial }{%
\partial q_{i}}\mathbf{A}_{j}^{T}\mathbf{A}_{j}.  \label{DKapa}
\end{equation}%
The derivatives of the inverse condition number is thus%
\begin{eqnarray}
\frac{\partial }{\partial q_{i}}\frac{1}{\kappa \left( \mathbf{A}\right) }
&=&-\frac{\frac{\partial }{\partial q_{i}}\kappa \left( \mathbf{A}\right) }{%
\kappa ^{2}\left( \mathbf{A}\right) }  \notag \\
&=&-\frac{1}{\kappa ^{2}\left( \mathbf{A}\right) }%
\Big%
(\left\Vert \mathbf{A}^{-1}\right\Vert _{2}\sum_{j=1}^{n}\frac{\partial }{%
\partial q_{i}}\mathbf{A}_{j}^{T}\mathbf{A}_{j}+\left\Vert \mathbf{A}%
\right\Vert _{2}\sum_{j=1}^{n}\frac{\partial }{\partial q_{i}}\mathbf{A}%
_{j}^{-T}\mathbf{A}_{j}^{-1}%
\Big%
).  \label{DinvCond}
\end{eqnarray}%
The last term in (\ref{DinvCond}) can be evaluated using the identity $%
\partial _{q_{j}}\mathbf{A}^{-1}=-\mathbf{A}^{-1}\partial _{q_{i}}\mathbf{AA}%
^{-1}$. The final result follows by using $\partial _{q_{i}}\mathbf{A}%
=\partial _{q_{i}}\mathbf{JJ}^{T}+\mathbf{J}\partial _{q_{i}}\mathbf{J}^{T}$%
. For non-redundant serial robots, $\mathbf{A}$ is replaced by $\mathbf{J}$.

The Hessian of the measure $\mu $ in (\ref{musigma}) follows directly from (%
\ref{dDet}) as%
\begin{equation}
\frac{\partial ^{2}}{\partial q_{i}\partial q_{j}}\mu =-\frac{1}{4\mu ^{3}}%
\sum_{\substack{ k,l\leq n  \\ k\neq l}}\det \left( \mathbf{A}_{1}\left\vert 
\underset{\,\;}{\overset{\,\;}{\cdots }}\right\vert \frac{\partial }{%
\partial q_{i}}\mathbf{A}_{k}\left\vert \underset{\,\;}{\overset{\,\;}{%
\cdots }}\right\vert \frac{\partial }{\partial q_{j}}\mathbf{A}%
_{l}\left\vert \underset{\,\;}{\overset{\,\;}{\cdots }}\right\vert \mathbf{A}%
_{n}\right) +\sum_{k\leq n}\det \left( \mathbf{A}_{1}\left\vert \underset{%
\,\;}{\overset{\,\;}{\cdots }}\right\vert \frac{\partial ^{2}}{\partial
q_{i}\partial q_{i}}\mathbf{A}_{k}\left\vert \underset{\,\;}{\overset{\,\;}{%
\cdots }}\right\vert \mathbf{A}_{n}\right) .  \label{DDet}
\end{equation}%
For non-redundant robotic arms, at regular configurations ($\det \mathbf{J}%
\neq 0$) the Hessian is the partial derivative of (\ref{derMu}) 
\begin{eqnarray*}
\frac{\partial ^{2}}{\partial q_{i}\partial q_{j}}\mu &=&\mu \left( \mathrm{%
tr\,}(\mathbf{J}^{-1}\partial _{q_{i}}\partial _{q_{j}}\mathbf{J})+\mathrm{%
tr\,}(\mathbf{J}^{-1}\partial _{q_{i}}\mathbf{J})\mathrm{tr}(\mathbf{J}%
^{-1}\partial _{q_{j}}\mathbf{J})+\mathrm{tr}\left( (\mathbf{J}^{-1}\partial
_{q_{i}}\mathbf{J})(\mathbf{J}^{-1}\partial _{q_{j}}\mathbf{J})\right)
\right) \\
&=&\mu \left( \sum_{k=j+1}^{n}(\overline{\mathbf{J}}^{T}\mathbf{ad}_{\mathbf{%
S}_{i}}\mathbf{ad}_{\mathbf{S}_{j}}\mathbf{S}_{k})+\left( \sum_{k=i+1}^{n}%
\overline{\mathbf{J}}_{k}^{T}\mathbf{ad}_{\mathbf{S}_{i}}\mathbf{S}%
_{k}\right) 
\hspace{-1ex}%
\left( \sum_{k=j+1}^{n}\overline{\mathbf{J}}_{k}^{T}\mathbf{ad}_{\mathbf{S}%
_{j}}\mathbf{S}_{k}\right) +\mathrm{tr}\left( (\mathbf{J}^{-1}\partial
_{q_{i}}\mathbf{J})(\mathbf{J}^{-1}\partial _{q_{j}}\mathbf{J})\right)
\right) ,i\leq j.
\end{eqnarray*}%
The explicit relations for the Hessian of the inverse condition number $%
\sigma $ are not presented here.

\begin{remark}
Both measures are based on the left invariant metric on $SE\left( 3\right) $%
, which depends on the scaling of rotations and translations. This has been
a central topic for the design of isotropic manipulators performing spatial
EE motions \cite{Angeles1992,AngelesMMT2006,DoelPai1996,ZlatanoNenchev2005}.
Amended measures were proposed to tackle this problem. A physically sensible
left-invariant dexterity measure is obtained by incorporating the inertia
tensor $\bm{\Theta}$ of a manipulated object. This gives rise to the
dexterity measure $\mu _{\bm{\Theta}}:=\sqrt{\det (\mathbf{J}\bm{\Theta}^{-1}%
\mathbf{J}^{T})}$ \cite{ParkKim1998}. For non-redundant manipulators $%
\mathbf{J}\bm{\Theta}\mathbf{J}^{T}$ can be considered from a
differential-geometric point of view as the inverse of the pull-back metric
on $\mathbb{V}^{n}$ induced by the metric on $SE\left( 3\right) $ defined by 
$\bm{\Theta}$. These local measures have also been extended to non-holonomic
manipulators \cite{Bayle}.
\end{remark}

\subparagraph{d) Local analysis of smooth motions of linkages with kinematic
loops%
\label{secKinCone}%
}

Fixing the terminal body $n$ of the kinematic chain at the ground yields a
kinematic loop formed by the chain of $n$ 1-DOF lower pairs. This leads to
the geometric loop closure constraints $f_{n}\left( \mathbf{q}\right) =%
\mathbf{I}$, where the KM (\ref{fi}) serves as \emph{constraint mapping} for
the kinematic loop. The solution variety%
\begin{equation}
V=\{\mathbf{q}\in {\mathbb{V}}^{n}|f_{n}\left( \mathbf{q}\right) =\mathbf{I}%
\}  \label{V}
\end{equation}%
serves as the configuration space (c-space) of the single-loop linkage. The
local dimension of $V$ at $\mathbf{q}$ is the \emph{local DOF} of the
linkage \cite{Selig} denoted $\delta _{\mathrm{loc}}\left( \mathbf{q}\right)
=\dim _{\mathbf{q}}V$.

The analysis of the c-space (\ref{V}) is a central topic for the mobility
and reconfiguration analysis of mechanisms. Since an explicit solution of
the geometric constraints is impossible in general, a local analysis aims to
identify tangent vectors to the motion curve in $V$, i.e. possible
velocities $\dot{\mathbf{q}}$, through a general configuration $\mathbf{q}%
\in V$ (singular or regular point of $V$). To this end, loop constraints
expressed by the POE were first explored in \cite{Herve1978,Herve1982} and
lead to significant contributions to the higher-order mobility analysis of
general linkages \cite{Bustos2012,Lerbet1999,JMR2016,JMR2018,Rico1999}. The
central elements of any such method are the higher-order time derivatives of
the loop closure constraints. Such formulations of up to 3rd order were
reported in \cite{Lerbet1999}, up to 4th order in \cite{Rico1999}. The
relations reported in this paper provide constraints of arbitrary order.
These were already applied to mobility and singularity analysis in \cite%
{JMR2018}.

The velocity constraints are expressed using (\ref{Vsi}) and (\ref{SSi}) as%
\begin{equation}
\mathbf{0}=\mathbf{J}_{n}^{\mathrm{s}}\left( \mathbf{q}\right) \dot{\mathbf{q%
}}=\mathsf{S}_{n}\left( \mathbf{q},\dot{\mathbf{q}}\right) .
\label{VelConstr}
\end{equation}%
The differential (instantaneous) DOF of the linkage at $\mathbf{q}$ is $%
\delta _{\mathrm{diff}}\left( \mathbf{q}\right) =n-\mathrm{rank}~\mathbf{J}%
_{n}^{\mathrm{s}}\left( \mathbf{q}\right) $.

The $k$th time derivative of the velocity constraints (\ref{VelConstr}) is 
\begin{equation}
H^{\left( k\right) }%
\hspace{-0.5ex}%
(\mathbf{q},\dot{\mathbf{q}},\ldots ,\mathbf{q}^{\left( k\right) })=\mathbf{0%
}  \label{HighOrderConstr}
\end{equation}%
with%
\begin{equation}
H^{\left( 1\right) }%
\hspace{-0.6ex}%
\left( \mathbf{q},\dot{\mathbf{q}}\right) :=\mathsf{S}_{n}\left( \mathbf{q},%
\dot{\mathbf{q}}\right) ,\ H^{\left( 2\right) }%
\hspace{-0.6ex}%
\left( \mathbf{q},\dot{\mathbf{q}},\ddot{\mathbf{q}}\right) :=\frac{d}{dt}%
\mathsf{S}_{n}\left( \mathbf{q},\dot{\mathbf{q}}\right) ,\ \ldots \
,H^{\left( i\right) }%
\hspace{-0.5ex}%
(\mathbf{q},\dot{\mathbf{q}},\ldots ,\mathbf{q}^{\left( i\right) }):=\mathrm{%
D}^{(i-1)}\mathsf{S}_{n}\left( \mathbf{q},\dot{\mathbf{q}}\right) .
\label{H}
\end{equation}%
The mappings (\ref{H}) are evaluated using (\ref{DSik}) since the explicit
relations (\ref{Vsrec})-(\ref{Vs3dotrec}) may not be sufficient since the
necessary order of such an analysis is not know a priori.

A finite motion through $\mathbf{q}\in V$ satisfies the velocity constraints
and all its time derivatives. Thus, a vector $\dot{\mathbf{q}}$ is tangent
to a curve through $\mathbf{q}\in V$ if and only if $H^{\left( 1\right) }%
\hspace{-0.6ex}%
\left( \mathbf{q},\dot{\mathbf{q}}\right) =\mathbf{0}$ and if there is a $%
\ddot{\mathbf{q}}\in {\mathbb{R}}^{n}$ such that $H^{\left( 2\right) }%
\hspace{-0.6ex}%
\left( \mathbf{q},\dot{\mathbf{q}},\ddot{\mathbf{q}}\right) =\mathbf{0}$,
and so forth. This is formalized by the \emph{kinematic tangent cone} \cite%
{Lerbet1999,JMR2018}, denoted ${C_{\mathbf{q}}^{\text{K}}V}$, which is the
set of possible velocities at $\mathbf{q}\in V$. It is determined by the
sequence%
\begin{equation}
{C_{\mathbf{q}}^{\text{K}}V}=K_{\mathbf{q}}^{\kappa }\subset \ldots \subset
K_{\mathbf{q}}^{3}\subset K_{\mathbf{q}}^{2}\subset {K_{\mathbf{q}}^{1}}
\label{CqV}
\end{equation}%
where each%
\begin{equation}
K_{\mathbf{q}}^{i}:=\left\{ \mathbf{x}|\exists \mathbf{y},\mathbf{z},\ldots
\in {\mathbb{R}}^{n}:H^{\left( 1\right) }%
\hspace{-0.6ex}%
\left( \mathbf{q},\mathbf{x}\right) =\mathbf{0},H^{\left( 2\right) }%
\hspace{-0.6ex}%
\left( \mathbf{q},\mathbf{x},\mathbf{y}\right) =\mathbf{0},H^{\left(
3\right) }%
\hspace{-0.6ex}%
\left( \mathbf{q},\mathbf{x},\mathbf{y},\mathbf{z}\right) =\mathbf{0},\cdots
H^{\left( i\right) }%
\hspace{-0.6ex}%
\left( \mathbf{q},\mathbf{x},\mathbf{y},\mathbf{z,\ldots }\right) =\mathbf{0}%
\right\} .
\end{equation}%
is a cone (rather than a vector space) satisfying the inclusion $K_{\mathbf{q%
}}^{i-1}\subset {K_{\mathbf{q}}^{i}}$. There is a finite order $\kappa $ so
that (\ref{CqV}) terminates. The kinematic tangent cone characterizes the
tangent aspects of smooth finite motions, i.e. smooth finite curves in $V$,
through $\mathbf{q}$. This applies to regular as well as to singular
configurations, so that $V$ does not have to be a smooth manifold, and thus
allows for investigation of singularities and linkage mobility.

The limitation of this local analysis is that it only reveals tangents to
smooth motions but cannot capture non-smooth finite curves through a c-space
singularity, where no tangent is defined (see sec. \ref{secApproxV}).

\subparagraph{Example 1: 4C-Linkage with a shaky motion mode%
\label{sec4CKinCone}%
}

\begin{figure}[b]
\centerline{
\includegraphics[width=0.38\textwidth]{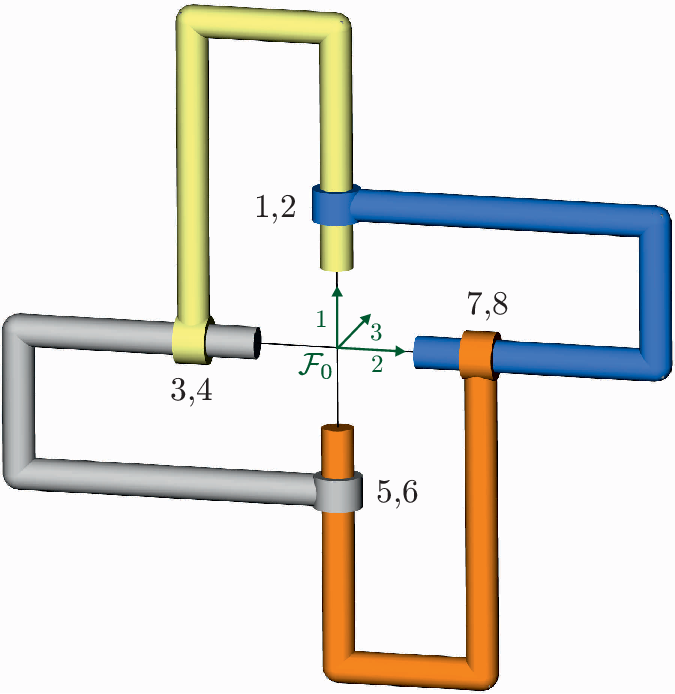}
}
\caption{A single-loop linkage comprising four cylindrical joints. The
latter are modeled as a combination of a revolute and a prismatic joint.
E.g. joint 1 and 2 represent one cylindrical joint.}
\label{fig4C}
\end{figure}

\begin{figure}[h]
\centerline{
\includegraphics[width=0.87\textwidth]{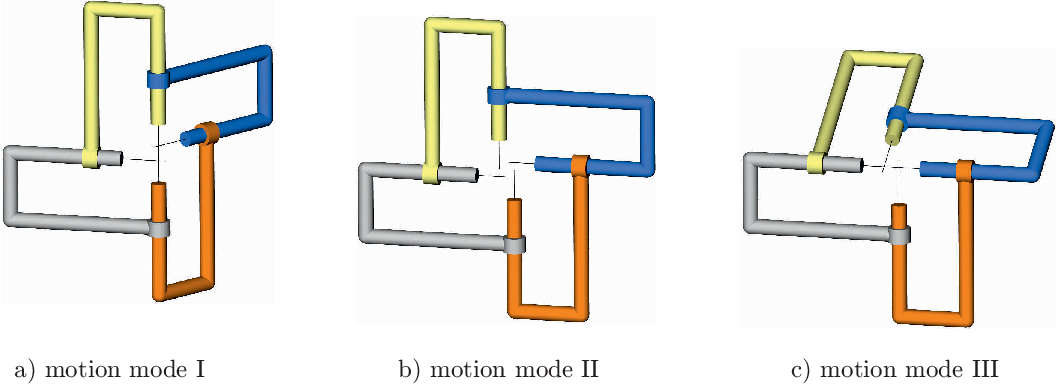}
}
\caption{Representative configurations in the three motion modes of the 4C
linkage in fig. \protect\ref{fig4C}, where a) $\mathbf{q}=\left(
0,0,a,b,0,0,-a,-b\right) $, b) $\mathbf{q}=\left( 0,a,0,b,0,-a,0,-b\right) $%
, and c) $\mathbf{q}=\left( a,b,0,0,-a,-b,0,0\right) $, with $a,b\in {%
\mathbb{R}}$. }
\label{fig4CModes}
\end{figure}

As a simple example consider the single-loop 4C linkage in fig. \ref{fig4C},
which was reported in \cite{LopezCustodio2017}. 
The cylindrical joints are modeled as combination of a revolute and
prismatic joint. The screw coordinates in the reference configuration $%
\mathbf{q}_{0}=\mathbf{0}$, represented in the shown frame $\mathcal{F}_{0}$%
, are%
\begin{equation}
\mathbf{Y}_{1}=\mathbf{Y}_{5}=\left( 1,0,0,0,0,0\right) ^{T},\mathbf{Y}_{2}=%
\mathbf{Y}_{6}=\left( 0,0,0,1,0,0\right) ^{T},\mathbf{Y}_{3}=\mathbf{Y}%
_{7}=\left( 0,1,0,0,0,0\right) ^{T},\mathbf{Y}_{4}=\mathbf{Y}_{8}=\left(
0,0,0,0,1,0\right) ^{T}.  \label{Y4C}
\end{equation}%
The mappings (\ref{H}) are evaluated with (\ref{DSik}). For instance, up to
order 3 these are

\begin{equation*}
H^{\left( 1\right) }%
\hspace{-0.6ex}%
\left( \mathbf{q}_{0},\mathbf{x}\right) =\left( 
\begin{array}{c}
x_{1}+x_{5} \\ 
x_{3}+x_{7} \\ 
0 \\ 
x_{2}+x_{6} \\ 
x_{4}+x_{8} \\ 
0%
\end{array}%
\right) ,\ \ H^{\left( 2\right) }%
\hspace{-0.6ex}%
\left( \mathbf{q}_{0},\mathbf{x},\mathbf{y}\right) =\left( 
\begin{array}{c}
y_{1}+y_{5} \\ 
y_{3}+y_{7} \\ 
x_{5}(-x_{3}+x_{7})+x_{1}(x_{3}+x_{7}) \\ 
y_{2}+y_{6} \\ 
y_{4}+y_{8} \\ 
-x_{4}x_{5}-x_{3}x_{6}+x_{6}x_{7}+x_{2}(x_{3}+x_{7})+x_{5}x_{8}+x_{1}(x_{4}+x_{8})%
\end{array}%
\right)
\end{equation*}%
\begin{equation*}
H^{\left( 3\right) }%
\hspace{-0.6ex}%
\left( \mathbf{q}_{0},\mathbf{x},\mathbf{y},\mathbf{z}\right) ={\small %
\left( 
\begin{array}{c}
-{x_{3}}{x_{5}}({x_{3}}-2{x_{7}})+{z_{1}}+{z_{5}} \\ 
-{x_{7}}({x_{1}}+{x_{5}})^{2}-{x_{1}}{x_{3}}({x_{1}}-2{x_{5}})+{z_{3}}+{z_{7}%
} \\ 
{x_{3}}{y_{1}}+2{x_{1}}{y_{3}}-{x_{5}}{y_{3}}-2{x_{3}}{y_{5}}+{x_{7}}({y_{1}}%
+{y_{5}})+2({x_{1}}+{x_{5}}){y_{7}} \\ 
2(-{x_{4}}{x_{5}}+{x_{8}}{x_{5}}+{x_{6}}{x_{7}}){x_{3}}-{x_{3}}^{2}{x_{6}}+2{%
x_{4}}{x_{5}}{x_{7}}+{z_{2}}+{z_{6}} \\ 
-2(-{x_{4}}{x_{5}}+{x_{8}}{x_{5}}-{x_{3}}{x_{6}}+{x_{6}}{x_{7}}+{x_{2}}({%
x_{3}}+{x_{7}})){x_{1}}-{x_{1}}^{2}({x_{4}}+{x_{8}})-{x_{5}}(-2{x_{2}}{x_{3}}%
+2({x_{2}}+{x_{6}}){x_{7}}+{x_{5}}{x_{8}})+{z_{4}}+{z_{8}} \\ 
-{x_{6}}{y_{3}}+2{x_{1}}{y_{4}}-{x_{5}}{y_{4}}+{x_{8}}({y_{1}}+{y_{5}})+{%
x_{4}}({y_{1}}-2{y_{5}})+{x_{7}}({y_{2}}+{y_{6}})+{x_{3}}({y_{2}}-2{y_{6}})+2%
{x_{6}}{y_{7}}+2{x_{2}}({y_{3}}+{y_{7}})+2({x_{1}}+{x_{5}}){y_{8}}%
\end{array}%
\right) }
\end{equation*}%
The first-order cone defined by $H^{\left( 1\right) }%
\hspace{-0.6ex}%
\left( \mathbf{q}_{0},\mathbf{x}\right) =0$ is%
\begin{equation}
K_{\mathbf{q}_{0}}^{1}=\{\mathbf{x}=(t,s,u,v,-t,-s,-u,-v)|t,s,u,v\in {%
\mathbb{R}}\}\subset {\mathbb{R}}^{8}.  \label{4CK1}
\end{equation}%
The second-order cone is implicitly defined by the system of second-order
polynomials $H^{\left( 1\right) }%
\hspace{-0.6ex}%
\left( \mathbf{q}_{0},\mathbf{x}\right) =H^{\left( 2\right) }%
\hspace{-0.6ex}%
\left( \mathbf{q}_{0},\mathbf{x},\mathbf{y}\right) =0$ as 
\begin{equation*}
K_{\mathbf{q}_{0}}^{2}=\{\mathbf{x}=\left(
x_{1},x_{2},x_{3},x_{4},x_{5},x_{6},x_{7},x_{8}\right) \in {\mathbb{R}}%
^{8}|x_{1}=-x_{5},x_{2}=-x_{6},x_{3}=-x_{7},x_{4}=-x_{8},x_{5}x_{7}=0,x_{6}x_{7}=-x_{5}x_{8},x_{5}^{2}x_{8}=0\}.
\end{equation*}%
The polynomial system can be factorized so that $K_{\mathbf{q}_{0}}^{2}$ is
the union of three vector spaces%
\begin{equation}
K_{\mathbf{q}_{0}}^{2}=K_{\mathbf{q}_{0}}^{2\left( I\right) }\cup K_{\mathbf{%
q}_{0}}^{2\left( II\right) }\cup K_{\mathbf{q}_{0}}^{2\left( III\right) }%
\begin{array}[t]{ll}
\ \ \ \text{with\ } & K_{\mathbf{q}_{0}}^{2\left( I\right) }=\{\mathbf{x}%
=(0,0,u,v,0,0,-u,-v)|u,v\in {\mathbb{R}}\} \\ 
& K_{\mathbf{q}_{0}}^{2\left( II\right) }=\{\mathbf{x}%
=(0,s,0,v,0,-s,0,-v)|s,v\in {\mathbb{R}}\} \\ 
& K_{\mathbf{q}_{0}}^{2\left( III\right) }=\{\mathbf{x}%
=(t,s,0,0-t,-s,0,0)|t,s\in {\mathbb{R}}\}.%
\end{array}
\label{4CK2}
\end{equation}%
The third- and higher-order cones are $K_{\mathbf{q}_{0}}^{i}=K_{\mathbf{q}%
_{0}}^{2},i\geq 2$. The kinematic tangent cone is thus $C_{\mathbf{q}_{0}}^{%
\mathrm{K}}V=K_{\mathbf{q}_{0}}^{2}$. Each of the three vector spaces in (%
\ref{4CK2}) is the tangent space to a manifold (the motion modes) passing
trough $\mathbf{q}_{0}$. The latter are the motion modes of the linkage
(shown in Fig. \ref{fig4CModes}). The point $\mathbf{q}_{0}$ is a c-space
singularity since $K_{\mathbf{q}_{0}}^{2}$ is not a vector space (the
reverse is not necessarily true). It is a bifurcation point so that any
smooth curve through $\mathbf{q}_{0}$ corresponds to one of the motion modes
where the linkage has DOF $\delta =2$.

\subparagraph{Example 2: Double Evans-Linkage%
\label{secEvansKinCone}%
}

There is a recent interest in linkages possessing singularities through
which no smooth motions are possible (in contrast to almost all
singularities considered in the literature that are characterized by the
intersection of motion modes). That is, the kinematic tangent cone (and any
analysis based on higher-order time derivatives) fails to reveal the
tangents to these non-smooth motions). So far, this phenomenon is known for
multi-loop 1-DOF linkages whose c-space possesses cusp singularities), for
which the planar linkage proposed in \cite{ConnellyServatius1994} is a
well-known example, but other (spatial) mechanisms were reported recently 
\cite{PabloMMT2018}.

\begin{figure}[h]
\centerline{
a)\includegraphics[width=0.47\textwidth]{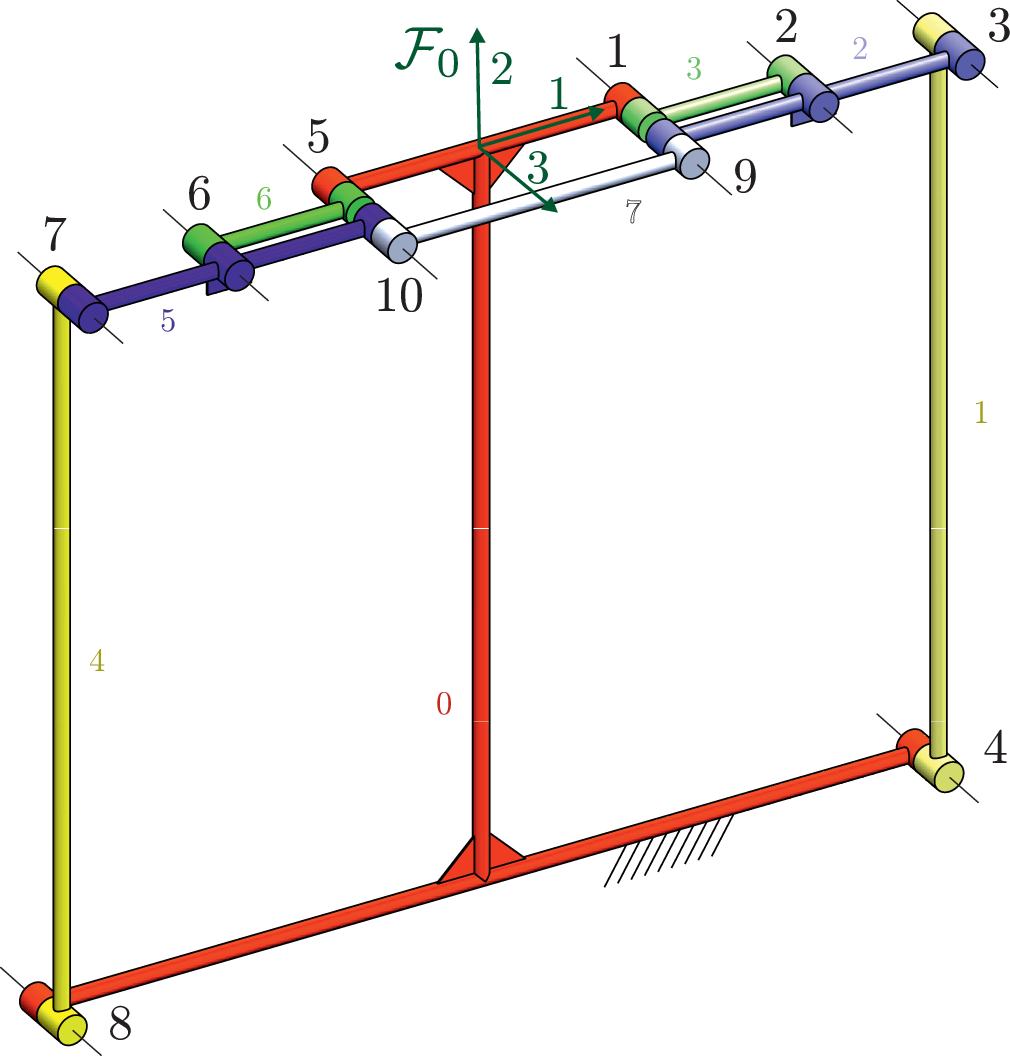}~~~~~
b)\includegraphics[width=0.47\textwidth]{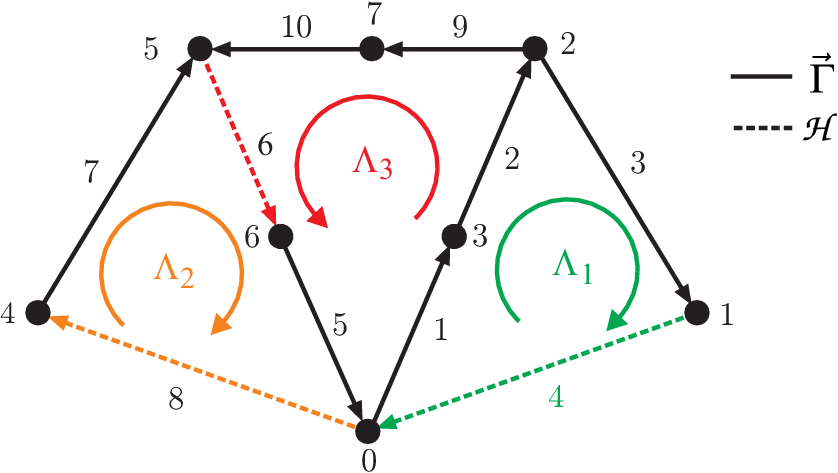}
}
\caption{a) Double-Evans linkage in a singular configuration, where the
c-space possesses a cusp (link numbers are omitted for clarity). b)
Topological graph and the fundamental cycles used for the analysis.}
\label{figDoubleEvans}
\end{figure}

The planar linkage constructed by combination of two Evans-linkages in fig. %
\ref{figDoubleEvans}a) is another 1 DOF example, which was presented in \cite%
{PabloMMT2018}. This linkage possesses three independent kinematic loops,
the fundamental cycles $\Lambda _{1},\Lambda _{2}$, and $\Lambda _{3}$,
indicated in the topological graph in fig \ref{figDoubleEvans}b). For each
of these loops the closure constraints are formulated \cite%
{Robotica2017,JMR2018}. Details are omitted here, but the detailed
calculation can be found in the accompanying Mathematica files \cite%
{MendeleyDataset}. The $i$th-order cone is then determined as%
\begin{equation}
K_{\mathbf{q}}^{i}:=\left\{ \mathbf{x}|\exists \mathbf{y},\mathbf{z},\ldots
\in {\mathbb{R}}^{n}:H_{l}^{\left( 1\right) }%
\hspace{-0.6ex}%
\left( \mathbf{q},\mathbf{x}\right) =\mathbf{0},\cdots H_{l}^{\left(
i\right) }%
\hspace{-0.6ex}%
\left( \mathbf{q},\mathbf{x},\mathbf{y},\mathbf{z,\ldots }\right) =\mathbf{0}%
,l=1,2,3\right\}
\end{equation}%
where the index $l$ indicates the kinematic loop. The screw coordinates in
the singular configuration $\mathbf{q}_{0}=\mathbf{0}$ are presented in \cite%
{PabloMMT2018}. The above analysis yields the cones

\begin{eqnarray}
K_{\mathbf{q}_{0}}^{1}
&=&%
\{x_{1}=s,x_{2}=-2s,x_{3}=s,x_{4}=0,x_{6}=-2t,x_{7}=t,x_{8}=0,x_{9}=2s-t,x_{10}=-s+2t;\ s,t\in 
{\mathbb{R}}\}  \label{EvansK1} \\
K_{\mathbf{q}_{0}}^{i}
&=&%
\{x_{1}=t,x_{2}=-2t,x_{3}=t,x_{4}=0,x_{6}=-2t,x_{7}=t,x_{8}=0,x_{9}=t,x_{10}=t;\ t\in 
{\mathbb{R}}\},i=2,3,4
\end{eqnarray}%
and $K_{\mathbf{q}_{0}}^{5}=\{\mathbf{0}\}$. Thus $C_{\mathbf{q}_{0}}^{%
\mathrm{K}}=\{\mathbf{0}\}$, which indicates that no smooth curve exists
through the singularity $\mathbf{q}_{0}$. It is known, however, that the
linkage is mobile with final DOF $\delta =1$ \cite{PabloMMT2018}. The above
analysis does not reveal this mobility since the linkage must stop when
traversing this singularity, i.e. there is no smooth motion through that
configuration. In order to deduce the correct finite DOF necessitates
analysis of the solution variety $V$, as discussed in the next section. 

\section{Series Expansion of the Kinematic Mapping%
\label{secSeriesKM}%
}

\subsection{Taylor series expansion of the kinematic mapping}

The kinematic mapping (\ref{fi}) is analytic and thus admits a Taylor series
expansion. W.l.o.g. the KM $f_{n}$ of the terminal body is considered, and
the index $n$ is omitted. The Taylor series of $f$ at $\mathbf{q}\in \mathbb{%
V}^{n}$ is 
\begin{equation}
f\left( \mathbf{q}+\mathbf{x}\right) =f\left( \mathbf{q}\right) +\sum_{k\geq
1}\frac{1}{k!}\left( \sum_{1\leq i\leq n}x_{i}\frac{\partial }{\partial q_{i}%
}\right) ^{k}%
\hspace{-0.9ex}%
f=f\left( \mathbf{q}\right) +\sum_{k\geq 1}\frac{1}{k!}\mathrm{d}^{k}f_{%
\mathbf{q}}\left( \mathbf{x}\right)  \label{fTaylor}
\end{equation}%
with the $k$th differential of $f$ 
\begin{eqnarray}
\mathrm{d}^{k}f_{\mathbf{q}}\left( \mathbf{x}\right) &=&\left( \sum_{1\leq
i\leq n}x_{i}\frac{\partial }{\partial q_{i}}\right) ^{k}%
\hspace{-0.9ex}%
f  \notag \\
&=&\sum_{\left\vert \mathbf{a}\right\vert =k}\frac{k!}{a_{1}!a_{2}!\cdots
a_{n}!}\frac{\partial ^{k}f}{\partial q_{1}^{a_{1}}\partial
q_{2}^{a_{2}}\ldots \partial q_{n}^{a_{n}}}x_{1}^{a_{1}}x_{2}^{a_{2}}\cdots
x_{n}^{a_{n}}=\sum_{\left\vert \mathbf{a}\right\vert =k}\frac{k!}{\mathbf{a}!%
}\mathbf{x}^{\mathbf{a}}\partial ^{\mathbf{a}}f  \notag \\
&=&\sum_{\alpha _{1},\alpha _{2},\ldots ,\alpha _{k}}x_{\alpha
_{1}}x_{\alpha _{2}}\cdots x_{\alpha _{n}}\frac{\partial ^{k}f}{\partial
q_{\alpha _{1}}\partial q_{\alpha _{2}}\ldots \partial q_{\alpha _{k}}}
\label{dkf2}
\end{eqnarray}%
evaluated at $\mathbf{q}$ (which is not indicated for sake of simplicity).

\begin{remark}
The differential can be expressed as $\mathrm{d}^{k}f_{\mathbf{q}}\left( 
\mathbf{x}\right) =(\mathrm{D}_{\mathbf{q}^{T}}^{k}f)(\mathbf{x}^{\otimes
k}\otimes \mathbf{I}_{4})$, where $\otimes $ denotes the Kronecker product
and $\mathbf{x}^{\otimes k}$ is $k$-fold tensor product $\mathbf{x}^{\otimes
k}=\mathbf{x}\otimes \ldots \otimes \mathbf{x}$ \cite%
{Steeb1991,Veter1973,deJong2018}. Moreover, higher-order derivatives can be
expressed compactly as Kronecker products. Since this merely serves for
'bookkeeping', but does not yield more insight, it will not be used in this
paper.
\end{remark}

\subsection{Higher-order differentials of the kinematic mapping}

The point of departure is the right-trivialized differential of the KM $f:{%
\mathbb{V}}^{n}\rightarrow SE\left( 3\right) $ at a configuration $\mathbf{q}%
\in {\mathbb{V}}^{n}$, which yields 
\begin{equation}
\mathrm{d}f_{\mathbf{q}}\left( \mathbf{x}\right) f\left( \mathbf{q}\right)
^{-1}=\sum_{i\leq n}\widehat{\mathbf{S}}_{i}\left( \mathbf{q}\right) x_{i}=%
\widehat{\mathsf{S}}_{n}\left( \mathbf{q},\mathbf{x}\right)  \label{df}
\end{equation}%
with $\mathsf{S}_{n}$ in (\ref{Si}), where $f\left( \mathbf{q}\right) ^{-1}$
is the inverse of the matrix $f\left( \mathbf{q}\right) \in SE\left(
3\right) $. Notice that higher-order differentials $\mathrm{d}^{k}f_{\mathbf{%
q}}$ of the KM are not the $k$-th differentials of $\mathsf{S}_{n}\,$when $%
f\left( \mathbf{q}\right) =\mathbf{I}$, since the equality $\mathrm{d}f_{%
\mathbf{q}}\left( \mathbf{x}\right) =\mathrm{d}f_{\mathbf{q}}\left( \mathbf{x%
}\right) f\left( \mathbf{q}\right) ^{-1}$ only holds for $k=1$. Introduce
the following mappings $h_{\mathbf{q}}^{\left( i\right) }:{\mathbb{R}}%
^{n}\rightarrow se\left( 3\right) $ 
\begin{align}
h_{\mathbf{q}}^{\left( 1\right) }%
\hspace{-0.5ex}%
\left( \mathbf{x}\right) & :=\sum_{i\leq n}\widehat{\mathbf{S}}_{i}\left( 
\mathbf{q}\right) x_{i},\ \ h_{\mathbf{q}}^{\left( 2\right) }%
\hspace{-0.5ex}%
\left( \mathbf{x}\right) :=\sum_{j,i\leq n}x_{i}x_{j}\frac{\partial \widehat{%
\mathbf{S}}_{i}}{\partial q_{j}},\ \ \ h_{\mathbf{q}}^{\left( 3\right) }%
\hspace{-0.5ex}%
\left( \mathbf{x}\right) :=\sum_{l,j,i\leq n}x_{i}x_{j}x_{l}\frac{\partial
^{2}\widehat{\mathbf{S}}_{i}}{\partial q_{l}\partial q_{j}}  \notag \\
h_{\mathbf{q}}^{\left( k\right) }%
\hspace{-0.5ex}%
\left( \mathbf{x}\right) & :=\sum_{\alpha _{1},\ldots ,\alpha _{k-1},i\leq
n}x_{i}x_{\alpha _{1}}\cdots x_{\alpha _{k-1}}\frac{\partial ^{k-1}\widehat{%
\mathbf{S}}_{i}}{\partial q_{\alpha _{1}}\cdots \partial q_{\alpha _{k-1}}}%
=\sum_{i\leq n}x_{i}\mathrm{d}^{k-1}\widehat{\mathbf{S}}_{i,\mathbf{q}%
}\left( \mathbf{x}\right) ,k>1  \label{h}
\end{align}%
where $\widehat{\mathbf{S}}_{i}$ is the $4\times 4$ matrix in (\ref{Xhat}).
The involved $k$-th differential of $\mathbf{S}_{i}$ at $\mathbf{q}$ is
defined as%
\begin{equation}
\mathrm{d}^{k}\mathbf{S}_{i,\mathbf{q}}\left( \mathbf{x}\right)
=\sum_{\alpha _{1},\ldots ,\alpha _{k}<i}x_{\alpha _{1}}\cdots x_{\alpha
_{k}}\frac{\partial ^{k}\mathbf{S}_{i}}{\partial q_{\alpha _{1}}\cdots
\partial q_{\alpha _{k}}}=\sum_{\left\vert \mathbf{a}_{i}\right\vert =k}%
\frac{k!}{\mathbf{a}_{i-1}!}\mathbf{x}^{\mathbf{a}_{i-1}}\partial ^{\mathbf{a%
}_{i-1}}\mathbf{S}_{i}  \label{dkS}
\end{equation}%
The last term in (\ref{dkS}) follows taking into account the index range in (%
\ref{dnS}) and (\ref{dnS2}). Application of Leibnitz' rule to (\ref{df})
yields%
\begin{equation}
h_{\mathbf{q}}^{\left( k\right) }%
\hspace{-0.5ex}%
\left( \mathbf{x}\right) =\sum_{i=0}^{k-1}\binom{k-1}{i}\mathrm{d}^{i+1}f_{%
\mathbf{q}}\left( \mathbf{x}\right) \mathrm{d}^{k-i-1}f_{\mathbf{q}%
}^{-1}\left( \mathbf{x}\right) =\mathrm{d}^{k}f_{\mathbf{q}}\left( \mathbf{x}%
\right) f\left( \mathbf{q}\right) ^{-1}+\sum_{i=1}^{k-1}\binom{k-1}{i-1}%
\mathrm{d}^{i+1}f_{\mathbf{q}}\left( \mathbf{x}\right) \mathrm{d}^{k-i}f_{%
\mathbf{q}}^{-1}\left( \mathbf{x}\right)  \label{h2}
\end{equation}%
where, with slight abuse of notation, $\mathrm{d}^{k}f_{\mathbf{q}}^{-1}$
denotes the the differential of the inverse of the matrix $f\left( \mathbf{q}%
\right) \in SE\left( 3\right) $ as function of $\mathbf{q}$. The
differential of $f$ follows from (\ref{h2}) as%
\begin{equation}
\mathrm{d}^{k}f_{\mathbf{q}}\left( \mathbf{x}\right) =h_{\mathbf{q}}^{\left(
k\right) }%
\hspace{-0.5ex}%
\left( \mathbf{x}\right) f\left( \mathbf{q}\right) -\sum_{i=1}^{k-1}\binom{%
k-1}{i-1}\mathrm{d}^{i}f_{\mathbf{q}}\left( \mathbf{x}\right) \mathrm{d}%
^{k-i}f_{\mathbf{q}}^{-1}%
\hspace{-0.5ex}%
\left( \mathbf{x}\right) f\left( \mathbf{q}\right) .  \label{dkf}
\end{equation}%
This is a recursive relation for $\mathrm{d}^{k}f_{\mathbf{q}}$ in terms of $%
\mathrm{d}^{i}f_{\mathbf{q}},i<k$, but also involves the differentials of $%
f\left( \mathbf{q}\right) ^{-1}$ up to order $k-1$.

The differential of order $k$ of $\mathbf{I}=f\left( \mathbf{q}\right)
f\left( \mathbf{q}\right) ^{-1}$ leads to%
\begin{equation}
\mathbf{0}=\sum_{i=0}^{k}\binom{k}{i}\mathrm{d}^{i}f_{\mathbf{q}}\left( 
\mathbf{x}\right) \mathrm{d}^{k-i}f_{\mathbf{q}}^{-1}%
\hspace{-0.5ex}%
\left( \mathbf{x}\right) =f\left( \mathbf{q}\right) \mathrm{d}^{k}f_{\mathbf{%
q}}^{-1}%
\hspace{-0.5ex}%
\left( \mathbf{x}\right) +\sum_{i=1}^{k}\binom{k}{i}\mathrm{d}^{i}f_{\mathbf{%
q}}\left( \mathbf{x}\right) \mathrm{d}^{k-i}f_{\mathbf{q}}^{-1}%
\hspace{-0.5ex}%
\left( \mathbf{x}\right)
\end{equation}%
and thus%
\begin{equation}
\mathrm{d}^{k}f_{\mathbf{q}}^{-1}\left( \mathbf{x}\right) =-f\left( \mathbf{q%
}\right) ^{-1}\sum_{i=1}^{k}\binom{k}{i}\mathrm{d}^{i}f_{\mathbf{q}}\left( 
\mathbf{x}\right) \mathrm{d}^{k-i}f_{\mathbf{q}}^{-1}%
\hspace{-0.5ex}%
\left( \mathbf{x}\right) .  \label{dkfinv}
\end{equation}%
The relation (\ref{dkfinv}) involves $\mathrm{d}^{i}f_{\mathbf{q}}^{-1},i<k$
and $\mathrm{d}^{i}f_{\mathbf{q}},i\leq k$. Recursive evaluation of the
latter, using (\ref{dkf}), requires $\mathrm{d}^{i}f_{\mathbf{q}},i<k$ and
thus $\mathrm{d}^{i}f_{\mathbf{q}}^{-1}%
\hspace{-0.5ex}%
\left( \mathbf{x}\right) ,i<k$.

It remains to determine the mappings $h_{\mathbf{q}}^{\left( k\right) }$,
which, according to (\ref{h}), requires the differentials of $\mathbf{S}_{i}$%
. The differential follows with (\ref{derSi}) as 
\begin{equation}
\mathrm{d}\mathbf{S}_{i,\mathbf{q}}\left( \mathbf{x}\right) =\sum_{j<i}[%
\mathbf{S}_{j}\left( \mathbf{q}\right) ,\mathbf{S}_{i}\left( \mathbf{q}%
\right) ]x_{j}.  \label{dS}
\end{equation}%
Applying Leibnitz' rule to (\ref{dS}) yields the relation for general order $%
k$ in terms of differentials of order less than $k$ 
\begin{equation}
\mathrm{d}^{k}\mathbf{S}_{i,\mathbf{q}}\left( \mathbf{x}\right)
=\sum_{j<i}\sum_{l=0}^{k-1}\binom{k-1}{l}[\mathrm{d}^{l}\mathbf{S}_{j,%
\mathbf{q}}\left( \mathbf{x}\right) ,\mathrm{d}^{k-l-1}\mathbf{S}_{i,\mathbf{%
q}}\left( \mathbf{x}\right) ]x_{j},k\geq 1.  \label{dkSii}
\end{equation}%
Inserting this into (\ref{h}) finally yields%
\begin{equation}
h_{\mathbf{q}}^{\left( k\right) }%
\hspace{-0.5ex}%
\left( \mathbf{x}\right) =\sum_{j<i<n}x_{i}x_{j}\sum_{l=0}^{k-2}\binom{k-2}{l%
}[\mathrm{d}^{l}\widehat{\mathbf{S}}_{j,\mathbf{q}}\left( \mathbf{x}\right) ,%
\mathrm{d}^{k-l-2}\widehat{\mathbf{S}}_{i,\mathbf{q}}\left( \mathbf{x}%
\right) ],k>1.  \label{hk2}
\end{equation}

In summary, the differential $\mathrm{d}^{k}f_{\mathbf{q}}$ of the kinematic
mapping (\ref{fi}) at $\mathbf{q}$ is determined recursively via the
relation (\ref{dkf}). This involves $\mathrm{d}^{k}f_{\mathbf{q}}^{-1}$,
which is determined recursively by (\ref{dkfinv}) in terms of $\mathrm{d}%
^{i}f_{\mathbf{q}},i<k$. It also involves $h_{\mathbf{q}}^{\left( k\right) }$%
, which deliver the actual contributions of the joint screw according to (%
\ref{hk2}) along with (\ref{dkSii}). For $k=0$ it is $\mathrm{d}^{k}\mathbf{S%
}_{j,\mathbf{q}}=\mathbf{S}_{j}\left( \mathbf{q}\right) $ and $\mathrm{d}%
^{k}f_{\mathbf{q}}^{-1}=f^{-1}\left( \mathbf{q}\right) $.

The $\mathrm{d}^{k}f_{\mathbf{q}}\left( \mathbf{x}\right) ,\mathrm{d}^{k}f_{%
\mathbf{q}}^{-1}%
\hspace{-0.5ex}%
\left( \mathbf{x}\right) ,h_{\mathbf{q}}^{\left( k\right) }%
\hspace{-0.5ex}%
\left( \mathbf{x}\right) ,\mathrm{d}^{k}\mathbf{S}_{i,\mathbf{q}}\left( 
\mathbf{x}\right) $ are homogenous polynomials in $\mathbf{x}$ of degree $k$.

\begin{remark}
The mappings (\ref{h}) can be evaluated using the explicit expression {}(\ref%
{dnS2}) for the partial derivatives as%
\begin{eqnarray}
h_{\mathbf{q}}^{\left( k+1\right) }%
\hspace{-0.5ex}%
\left( \mathbf{x}\right) &=&\sum_{\alpha _{1},\ldots ,\alpha _{k},i\leq
n}x_{i}x_{\alpha _{1}}\cdots x_{\alpha _{k}}\frac{\partial ^{k}\widehat{%
\mathbf{S}}_{i}}{\partial q_{\alpha _{1}}\cdots \partial q_{\alpha _{k}}}%
=\sum_{i\leq n}x_{i}\mathrm{d}^{k}\widehat{\mathbf{S}}_{i,\mathbf{q}}\left( 
\mathbf{x}\right) =\sum_{i\leq n}\sum_{\left\vert \mathbf{a}%
_{i-1}\right\vert =k}\frac{k!}{\mathbf{a}_{i-1}!}x_{i}\mathbf{x}^{\mathbf{a}%
_{i-1}}\partial ^{\mathbf{a}_{i-1}}\widehat{\mathbf{S}}_{i}  \notag \\
&=&\sum_{i\leq n}\sum_{\left\vert \mathbf{a}_{i-1}\right\vert =k}\frac{k!}{%
\mathbf{a}_{i-1}!}x_{i}\mathbf{x}^{\mathbf{a}_{i-1}}%
\Big%
(\prod\limits_{j=1}^{i-1}\mathbf{ad}_{\mathbf{S}_{j}}^{a_{j}}(\mathbf{S}_{i})%
\Big%
)^{\widehat{}}.  \label{hk}
\end{eqnarray}%
Evaluation of (\ref{hk}) in closed form requires repeated evaluation of all
involved Lie brackets, respectively the adjoint operations. The recursive
formulation (\ref{hk2}), on the other hand, is computationally more
efficient and easier to implement.
\end{remark}

\begin{remark}
If a low-order truncation of the Taylor series (\ref{fTaylor}) is
sufficient, (\ref{dkf}) along with (\ref{dkfinv}) can be rolled out
explicitly and the following relations (up to order 4, for instance) be used
directly%
\begin{eqnarray}
\mathrm{d}^{1}f_{\mathbf{q}}\left( \mathbf{x}\right) &=&h_{\mathbf{q}%
}^{\left( 1\right) }%
\hspace{-0.5ex}%
\left( \mathbf{x}\right) f\left( \mathbf{q}\right) \\
\mathrm{d}^{2}f_{\mathbf{q}}\left( \mathbf{x}\right) &=&\left( h_{\mathbf{q}%
}^{\left( 2\right) }%
\hspace{-0.5ex}%
\left( \mathbf{x}\right) +h_{\mathbf{q}}^{\left( 1\right) }%
\hspace{-0.5ex}%
\left( \mathbf{x}\right) h_{\mathbf{q}}^{\left( 1\right) }%
\hspace{-0.5ex}%
\left( \mathbf{x}\right) \right) f\left( \mathbf{q}\right) \\
\mathrm{d}^{3}f_{\mathbf{q}}\left( \mathbf{x}\right) &=&\left( h_{\mathbf{q}%
}^{\left( 3\right) }%
\hspace{-0.5ex}%
\left( \mathbf{x}\right) +h_{\mathbf{q}}^{\left( 1\right) }%
\hspace{-0.5ex}%
\left( \mathbf{x}\right) h_{\mathbf{q}}^{\left( 2\right) }%
\hspace{-0.5ex}%
\left( \mathbf{x}\right) +2h_{\mathbf{q}}^{\left( 2\right) }%
\hspace{-0.5ex}%
\left( \mathbf{x}\right) h_{\mathbf{q}}^{\left( 1\right) }%
\hspace{-0.5ex}%
\left( \mathbf{x}\right) +h_{\mathbf{q}}^{\left( 1\right) }%
\hspace{-0.5ex}%
\left( \mathbf{x}\right) h_{\mathbf{q}}^{\left( 1\right) }%
\hspace{-0.5ex}%
\left( \mathbf{x}\right) h_{\mathbf{q}}^{\left( 1\right) }%
\hspace{-0.5ex}%
\left( \mathbf{x}\right) \right) f\left( \mathbf{q}\right)  \notag \\
\mathrm{d}^{4}f_{\mathbf{q}}\left( \mathbf{x}\right) &=&\left( h_{\mathbf{q}%
}^{\left( 4\right) }%
\hspace{-0.5ex}%
\left( \mathbf{x}\right) +h_{\mathbf{q}}^{\left( 1\right) }%
\hspace{-0.5ex}%
\left( \mathbf{x}\right) h_{\mathbf{q}}^{\left( 3\right) }%
\hspace{-0.5ex}%
\left( \mathbf{x}\right) +3h_{\mathbf{q}}^{\left( 3\right) }%
\hspace{-0.5ex}%
\left( \mathbf{x}\right) h_{\mathbf{q}}^{\left( 1\right) }%
\hspace{-0.5ex}%
\left( \mathbf{x}\right) +3h_{\mathbf{q}}^{\left( 2\right) }%
\hspace{-0.5ex}%
\left( \mathbf{x}\right) h_{\mathbf{q}}^{\left( 2\right) }%
\hspace{-0.5ex}%
\left( \mathbf{x}\right) \right. \\
&&\ \ \left. +3h_{\mathbf{q}}^{\left( 2\right) }%
\hspace{-0.5ex}%
\left( \mathbf{x}\right) h_{\mathbf{q}}^{\left( 1\right) }%
\hspace{-0.5ex}%
\left( \mathbf{x}\right) h_{\mathbf{q}}^{\left( 1\right) }%
\hspace{-0.5ex}%
\left( \mathbf{x}\right) +2h_{\mathbf{q}}^{\left( 1\right) }%
\hspace{-0.5ex}%
\left( \mathbf{x}\right) h_{\mathbf{q}}^{\left( 2\right) }%
\hspace{-0.5ex}%
\left( \mathbf{x}\right) h_{\mathbf{q}}^{\left( 1\right) }%
\hspace{-0.5ex}%
\left( \mathbf{x}\right) +h_{\mathbf{q}}^{\left( 1\right) }%
\hspace{-0.5ex}%
\left( \mathbf{x}\right) h_{\mathbf{q}}^{\left( 1\right) }%
\hspace{-0.5ex}%
\left( \mathbf{x}\right) h_{\mathbf{q}}^{\left( 1\right) }%
\hspace{-0.5ex}%
\left( \mathbf{x}\right) h_{\mathbf{q}}^{\left( 1\right) }%
\hspace{-0.5ex}%
\left( \mathbf{x}\right) \right) f\left( \mathbf{q}\right) .
\end{eqnarray}
\end{remark}

\begin{remark}
The second-order term $h_{\mathbf{q}}^{\left( 2\right) }%
\hspace{-0.5ex}%
\left( \mathbf{x}\right) $ reads explicitly%
\begin{equation}
h_{\mathbf{q}}^{\left( 2\right) }%
\hspace{-0.5ex}%
\left( \mathbf{x}\right) =\sum_{j<i\leq n}x_{i}x_{j}\frac{\partial \widehat{%
\mathbf{S}}_{i}}{\partial q_{j}}=\sum_{j<i\leq n}x_{i}x_{j}\mathrm{ad}_{%
\widehat{\mathbf{S}}_{j}}(\widehat{\mathbf{S}}_{i}).
\end{equation}%
The $h_{\mathbf{q}}^{\left( k\right) }$ map to $se\left( 3\right) $, so $h_{%
\mathbf{q}}^{\left( 2\right) }%
\hspace{-0.5ex}%
\left( \mathbf{x}\right) \in se\left( 3\right) $ can be represented as
6-vector. Then%
\begin{eqnarray}
\sum_{j<i\leq n}x_{i}x_{j}\mathbf{ad}_{\mathbf{S}_{j}}\mathbf{S}_{i}
&=&\left( \mathbf{x}^{T}\otimes \mathbf{I}_{6}\right) \mathrm{diag}~\left( 
\mathbf{ad}_{\mathbf{S}_{1}},\ldots ,\mathbf{ad}_{\mathbf{S}_{n}}\right)
\left( 
\begin{array}{cccccc}
\mathbf{0} & \mathbf{S}_{2} & \mathbf{S}_{3} & \cdots & \mathbf{S}_{n-1} & 
\mathbf{S}_{n} \\ 
& \mathbf{0} & \mathbf{S}_{3} & \cdots & \mathbf{S}_{n-1} & \mathbf{S}_{n}
\\ 
&  & \mathbf{0} & \ddots & \vdots & \vdots \\ 
& \mathbf{0} &  &  & \mathbf{0} & \mathbf{S}_{n} \\ 
&  &  &  &  & \mathbf{0}%
\end{array}%
\right) \mathbf{x}  \notag \\
&=&\frac{1}{2}\left( \mathbf{x}^{T}\otimes \mathbf{I}_{6}\right) \left( 
\begin{array}{ccccccc}
\mathbf{0} & \mathbf{ad}_{\mathbf{S}_{1}}\mathbf{S}_{2} & \mathbf{ad}_{%
\mathbf{S}1}\mathbf{S}_{3} & \mathbf{ad}_{\mathbf{S}_{1}}\mathbf{S}_{4} & 
\cdots & \mathbf{ad}_{\mathbf{S}_{1}}\mathbf{S}_{n-1} & \mathbf{ad}_{\mathbf{%
S}_{1}}\mathbf{S}_{n} \\ 
\mathbf{ad}_{\mathbf{S}_{1}}\mathbf{S}_{2} & \mathbf{0} & \mathbf{ad}_{%
\mathbf{S}_{2}}\mathbf{S}_{3} & \mathbf{ad}_{\mathbf{S}_{2}}\mathbf{S}_{4} & 
\cdots & \mathbf{ad}_{\mathbf{S}_{2}}\mathbf{S}_{n-1} & \mathbf{ad}_{\mathbf{%
S}_{2}}\mathbf{S}_{n} \\ 
\mathbf{ad}_{\mathbf{S}_{1}}\mathbf{S}_{3} & \mathbf{ad}_{\mathbf{S}_{2}}%
\mathbf{S}_{3} & \mathbf{0} & \mathbf{S}_{4} & \cdots & \mathbf{S}_{n-1} & 
\mathbf{S}_{n} \\ 
\mathbf{ad}_{\mathbf{S}_{1}}\mathbf{S}_{4} & \mathbf{ad}_{\mathbf{S}_{2}}%
\mathbf{S}_{4} & \mathbf{ad}_{\mathbf{S}_{2}}\mathbf{S}_{n} & \mathbf{0} & 
& \mathbf{S}_{n-1} & \mathbf{S}_{n} \\ 
\vdots & \vdots & \vdots &  & \ddots & \vdots & \vdots \\ 
\mathbf{ad}_{\mathbf{S}_{1}}\mathbf{S}_{n-1} & \mathbf{ad}_{\mathbf{S}_{2}}%
\mathbf{S}_{n-1} & -\mathbf{S}_{3} & -\mathbf{S}_{4} & \cdots & \mathbf{0} & 
\mathbf{S}_{n} \\ 
\mathbf{ad}_{\mathbf{S}_{1}}\mathbf{S}_{n} & \mathbf{ad}_{\mathbf{S}_{2}}%
\mathbf{S}_{n} & -\mathbf{S}_{3} & -\mathbf{S}_{4} & \cdots & -\mathbf{S}%
_{n-1} & \mathbf{0}%
\end{array}%
\right) \mathbf{x}.
\end{eqnarray}%
This quadratic form has been used in \cite{WuIDETC2018} to identify immobile
shaky linkages.
\end{remark}

\begin{remark}
For the analysis of a particular configuration, the joint variables can
always be defined so that the configuration of interest is the reference
configuration with $\mathbf{q}=\mathbf{0}$. In this case the instantaneous
joints screw coordinates are $\mathbf{S}_{j}\left( \mathbf{0}\right) =%
\mathbf{Y}_{j}$. Then (\ref{fTaylor}) is the MacLaurin series of $f$.
\end{remark}

\subsection{Application: Local Approximation of the Configuration Space%
\label{secApproxV}%
}

The KM defines the c-space $V$ of a closed loop linkage in (\ref{V}). Its
(local) dimension at $\mathbf{q}\in V$ is the (local) finite DOF of the
linkage. The local geometry of $V$ reveals the finite mobility and c-space
singularities of the linkages. A global analysis of $V$ is not possible in
general. A local approximation of $V$ is given by replacing $f$ in (\ref{V})
with a finite truncation of its series expansion (\ref{fTaylor}). Since $%
f\left( \mathbf{q}\right) =\mathbf{I}$ for $\mathbf{q}\in V$, the $k$%
th-order local approximation of the c-space at $\mathbf{q}\in V$ is%
\begin{equation}
V_{\mathbf{q}}^{k}:=\{\mathbf{x}\in {\mathbb{R}}^{n}|\mathrm{d}f_{\mathbf{q}%
}\left( \mathbf{x}\right) +\frac{1}{2}\mathrm{d}^{2}f_{\mathbf{q}}\left( 
\mathbf{x}\right) +\ldots +\frac{1}{k!}\mathrm{d}^{k}f_{\mathbf{q}}\left( 
\mathbf{x}\right) =\mathbf{0}\}.  \label{Vk}
\end{equation}%
$V_{\mathbf{q}}^{\kappa }$ is an algebraic variety of degree $k$. The
dimension of $V_{\mathbf{q}}^{k}$ is the $k$th-order local DOF at $\mathbf{q}%
\in V$. There exist a neighborhood $U\left( \mathbf{q}\right) $ and an order 
$\kappa $ such that $V_{\mathbf{q}}^{\kappa }\cap U\left( \mathbf{q}\right)
=V\cap U\left( \mathbf{q}\right) $. That is, a finite approximation of order 
$\kappa $ is sufficient and thus $\delta _{\mathrm{loc}}\left( \mathbf{q}%
\right) =\dim V_{\mathbf{q}}^{\kappa }$.

The local mobility determination is an open problem in mechanism theory. An
approach that is applicable to general mechanisms is the local analysis of
the c-space $V$. No general computational framework was proposed so far, and
the above relations may serve as a basis for such a higher-order local
approximation, as proposed in \cite{JMR2018LocApprox}. One publication were
special cases are addressed is \cite{Chen2011}. A Taylor series
approximation is used in \cite{Chen2011} in order to characterize possible
motions at singularities and to check for the mobility of overconstrained
single loop linkages, in particular the Bennett conditions for 4R linkages.
The Taylor expansion was derived specifically for 4R linkages.

\begin{remark}
$V_{\mathbf{q}}^{k}$ is an algebraic variety of order $k$ approximating the
analytic variety $V$. The approximation order necessary to determine the
local DOF is not known a priori and depends on the linkage as well as the
configuration. At regular configurations of non-overconstrained linkages,
for which the rank of the constraint Jacobian $\mathbf{J}$ in (\ref%
{VelConstr}) is locally constant, a first-order approximation is sufficient.
In singular configurations, where the rank of $\mathbf{J}$ is not locally
constant, higher-order approximations are necessary. Moreover, the mobility
analysis demands a higher-order approximation since it is not known
beforehand whether the configuration is regular or singular, and whether the
linkage is overconstrained or not. The same applies to underconstrained
(shaky) linkages \cite{JMR2016,WohlhartShakiness,Wohlhart2010}, which
possess a higher differential DOF than local DOF even in regular
configurations. While the above relations deliver the $k$th-order
polynomials defining a local approximation it remains to analyze the
corresponding algebraic variety.
\end{remark}

\subparagraph{Example 1 (cont.): 4C-Linkage with a shaky motion mode}

The differentials of $f$ are determined with the recursive relation (\ref%
{dkf}). The first and second differential are%
\begin{equation*}
\mathrm{d}f_{\mathbf{q}_{0}}\left( \mathbf{x}\right) ={\small \left( 
\begin{array}{cccc}
0 & 0 & x_{3}+x_{7} & x_{2}+x_{6} \\ 
0 & 0 & -x_{1}-x_{5} & x_{4}+x_{8} \\ 
-x_{3}-x_{7} & x_{1}+x_{5} & 0 & 0 \\ 
0 & 0 & 0 & 0%
\end{array}%
\right) }
\end{equation*}%
\begin{equation*}
\mathrm{d}f_{\mathbf{q}_{0}}\left( \mathbf{x}\right) +\frac{1}{2}\mathrm{d}%
^{2}f_{\mathbf{q}_{0}}\left( \mathbf{x}\right) ={\small \left( 
\begin{array}{cccc}
-\frac{1}{2}(x_{3}+x_{7})^{2} & x_{3}x_{5} & x_{3}+x_{7} & x_{2}+x_{6} \\ 
x_{5}x_{7}+x_{1}(x_{3}+x_{7}) & -\frac{1}{2}(x_{1}+x_{5})^{2} & -x_{1}-x_{5}
& 0 \\ 
-x_{3}-x_{7} & x_{1}+x_{5} & -\frac{1}{2}%
((x_{1}+x_{5})^{2}+(x_{3}+x_{7})^{2}) & 
-x_{3}x_{6}+x_{5}x_{8}+x_{1}(x_{4}+x_{8}) \\ 
0 & 0 & 0 & 0%
\end{array}%
\right) .}
\end{equation*}%
The first-order approximation (always) is $V_{\mathbf{q}_{0}}^{1}=\ker 
\mathrm{d}f_{\mathbf{q}_{0}}=\ker \mathbf{J}\left( \mathbf{q}_{0}\right) ={%
K_{\mathbf{q}_{0}}^{1}}$ in (\ref{4CK1}). The second-order approximation is
defined by $\mathrm{d}f_{\mathbf{q}_{0}}\left( \mathbf{x}\right) +\frac{1}{2}%
\mathrm{d}^{2}f_{\mathbf{q}_{0}}\left( \mathbf{x}\right) =\mathbf{0}$, which
can be simplified to%
\begin{equation*}
V_{\mathbf{q}_{0}}^{2}=\{\mathbf{x}\in {\mathbb{R}}%
^{8}|x_{3}+x_{7}=0,x_{1}+x_{5}=0,x_{3}x_{5}=0,x_{5}x_{7}+x_{1}(x_{3}+x_{7})=0,-x_{3}x_{6}+x_{5}x_{8}+x_{1}(x_{4}+x_{8})=0\}.
\end{equation*}%
This can be solved, and it turns out that $V_{\mathbf{q}_{0}}^{2}=C_{\mathbf{%
q}_{0}}^{\mathrm{K}}V=K_{\mathbf{q}_{0}}^{2\left( I\right) }\cup K_{\mathbf{q%
}_{0}}^{2\left( II\right) }\cup K_{\mathbf{q}_{0}}^{2\left( III\right) }$,
with $C_{\mathbf{q}_{0}}^{\mathrm{K}}V$ in (\ref{4CK2}). Also for all higher
approximations it is $V_{\mathbf{q}_{0}}^{k}=C_{\mathbf{q}_{0}}^{\mathrm{K}%
}V,k\geq 2$, which is the union of vector spaces. Consequently, for all
possible finite motions there is a linear relation of the joint variable. It
also shows that all motion branches correspond to smooth finite motions
through the singularity, i.e. $\mathbf{q}$ is a bifurcation point. Details
can found in the provided Mathematica notebook \cite{MendeleyDataset}.

\subparagraph{Example 2 (cont.): Double Evans-Linkage}

It was concluded in sec. \ref{secEvansKinCone} that this linkage does not
admit smooth motions through the singular configuration, although it is
mobile. The presented approach to the local analysis can be applied to
multi-loop linkages by considering the topologically independent kinematic
loops. Denote with $f_{l}$ the mapping (\ref{fi}) defining the geometric
closure condition, then the c-space is $V=\{\mathbf{q}\in {\mathbb{V}}%
^{n}|f_{l}\left( \mathbf{q}\right) =\mathbf{I},l=1,2,3\}$. For $\Lambda _{1}$
the closure mapping is $f_{1}\left( \mathbf{q}\right) =\exp (\mathbf{Y}%
_{1}q_{1})\exp (\mathbf{Y}_{2}q_{2})\exp (\mathbf{Y}_{3}q_{3})\exp (\mathbf{Y%
}_{4}q_{4})$, for $\Lambda _{2}$ it is $f_{2}\left( \mathbf{q}\right) =\exp (%
\mathbf{Y}_{8}q_{8})\exp (\mathbf{Y}_{7}q_{7})\exp (\mathbf{Y}_{6}q_{6})\exp
(\mathbf{Y}_{5}q_{5})$, and for $\Lambda _{3}$ it is $f_{3}\left( \mathbf{q}%
\right) =\exp (\mathbf{Y}_{1}q_{1})\exp (\mathbf{Y}_{2}q_{2})\exp (\mathbf{Y}%
_{9}q_{9})\exp (\mathbf{Y}_{10}q_{10})\exp (\mathbf{Y}_{6}q_{6})\exp (%
\mathbf{Y}_{5}q_{5})$. Their series expansions, defining the approximation $%
V_{\mathbf{q}_{0}}^{k}$ in (\ref{Vk}), are easily constructed with the above
recursive relations (details are omitted again, for sake of readability).
This yields a system of polynomials of degree $i$. The crucial step,
however, is checking the real dimension of the so defined algebraic
varieties $V_{\mathbf{q}_{0}}^{k}$, or to even explicitly solve these
systems in order to obtain a parameterization of the motion curve. This will
possibly require using dedicated algorithms from algebraic geometry \cite%
{CoxLittleOShea2007,WalterHusty2010}. For this example, the dimension test
was performed with the software Singular \cite{SingularBook}. The
first-order approximation is, by definition, always $V_{\mathbf{q}}^{1}=K_{%
\mathbf{q}}^{1}$, i.e. $\dim V_{\mathbf{q}_{0}}^{1}=2$, see (\ref{EvansK1}).
The polynomials defining $V_{\mathbf{q}_{0}}^{i}$ were determined with the
recursive relations (\ref{dkf}) and (\ref{dkfinv}) using the provided
Mathematica package \cite{MendeleyDataset}, while the dimension of the
algebraic variety $V_{\mathbf{q}_{0}}^{i}$ defined by this system of
polynomials was determined with the software Singular \cite{SingularBook}.
Details can be found in the accompanying Mathematica notebook \cite%
{MendeleyDataset}. The higher-order approximations all have $\dim V_{\mathbf{%
q}_{0}}^{k}=1$. It is hence concluded that the linkage has finite local
mobility $\delta \left( \mathbf{q}_{0}\right) =1$.

\section{Time Derivatives and Series Expansion of Minors of the Geometric
Jacobian%
\label{secJacDer}%
}

\subsection{Time derivatives of Minors of the Geometric Jacobian%
\label{secTimeDerMin}%
}

W.l.o.g. the spatial Jacobian $\mathbf{J}_{n}^{\mathrm{s}}$ of body $n$ is
considered, and for the sake of simplicity, this is denoted with $\mathbf{J}$%
.

In the following, $\mathbf{J}_{%
\bm{\alpha}%
\bm{\beta}%
}$ denotes the $k\times k$ submatrix of $\mathbf{J}$, consisting of the
elements of the rows $\alpha _{i}\in \{1,\ldots ,6\}$ and columns $\beta
_{j}\in \{1,\ldots ,n\}$ of $\mathbf{J}$ summarized in the index sets $%
\bm{\alpha}%
=\{\alpha _{1},\ldots ,\alpha _{k}\},\alpha _{i-1}<\alpha _{i}$ and $%
\bm{\beta}%
=\{\beta _{1},\ldots ,\beta _{k}\},\beta _{j-1}<\beta _{j}$. The determinant
of this matrix is denoted with $m_{%
\bm{\alpha}%
\bm{\beta}%
}(\mathbf{q}):=\det \mathbf{J}_{%
\bm{\alpha}%
\bm{\beta}%
}(\mathbf{q})$, and referred to as the $%
\bm{\alpha}%
\bm{\beta}%
$-minor of $\mathbf{J}$ of order $k$. Notice that this differs from the
conventional definition of minors where the index set would indicate the
rows and columns that are eliminated from $\mathbf{J}$. Apparently, if $%
\mathbf{J}$ is a square $k\times k$ matrix, then the (only one) $k$ minor is 
$m_{%
\bm{\alpha}%
\bm{\beta}%
}=\det \mathbf{J}$.

The columns of $\mathbf{J}$ are the instantaneous joint screws (\ref{Si}).
Denote with $\mathbf{S}_{%
\bm{\alpha}%
j}=\left( S_{a_{1}j},S_{a_{2}j},\ldots ,S_{a_{k}j}\right) ^{T}$ the
subvector of the screw coordinate vector $\mathbf{S}_{j}=\left(
S_{1j},S_{2j},\ldots ,S_{6j}\right) ^{T}$ with elements according to row
indexes $%
\bm{\alpha}%
$. The time derivatives of $m_{%
\bm{\alpha}%
\bm{\beta}%
}$ are found, using the property of determinants \cite{Steeb1991}. Up to 5th
order, for instance, they are%
\begin{eqnarray*}
\frac{d}{dt}m_{%
\bm{\alpha}%
\bm{\beta}%
} &=&\sum_{\mu \in 
\bm{\beta}%
}\left\vert \mathbf{S}_{%
\bm{\alpha}%
\beta _{1}}\cdots \dot{\mathbf{S}}_{%
\bm{\alpha}%
\mu }\cdots \mathbf{S}_{%
\bm{\alpha}%
\beta _{k}}\right\vert \\
\frac{d^{2}}{dt^{2}}m_{%
\bm{\alpha}%
\bm{\beta}%
} &=&\sum_{\mu \in 
\bm{\beta}%
}\left\vert \mathbf{S}_{%
\bm{\alpha}%
\beta _{1}}\cdots \ddot{\mathbf{S}}_{%
\bm{\alpha}%
\mu }\cdots \mathbf{S}_{%
\bm{\alpha}%
\beta _{k}}\right\vert +\sum_{\mu \neq \nu \in 
\bm{\beta}%
}\left\vert \mathbf{S}_{%
\bm{\alpha}%
\beta _{1}}\cdots \dot{\mathbf{S}}_{%
\bm{\alpha}%
\mu }\cdots \dot{\mathbf{S}}_{%
\bm{\alpha}%
\nu }\cdots \mathbf{S}_{%
\bm{\alpha}%
\beta _{k}}\right\vert \\
\frac{d^{3}}{dt^{3}}m_{%
\bm{\alpha}%
\bm{\beta}%
} &=&\sum_{\mu \in 
\bm{\beta}%
}\left\vert \mathbf{S}_{%
\bm{\alpha}%
\beta _{1}}\cdots \dddot{\mathbf{S}}_{%
\bm{\alpha}%
\mu }\cdots \mathbf{S}_{%
\bm{\alpha}%
\beta _{k}}\right\vert +3\sum_{\mu \neq \nu \in 
\bm{\beta}%
}\left\vert \mathbf{S}_{%
\bm{\alpha}%
\beta _{1}}\cdots \ddot{\mathbf{S}}_{%
\bm{\alpha}%
\mu }\cdots \dot{\mathbf{S}}_{%
\bm{\alpha}%
\nu }\cdots \mathbf{S}_{%
\bm{\alpha}%
\beta _{k}}\right\vert +\sum_{\mu \neq \nu \neq \lambda \in 
\bm{\beta}%
}\left\vert \mathbf{S}_{%
\bm{\alpha}%
\beta _{1}}\cdots \dot{\mathbf{S}}_{%
\bm{\alpha}%
\mu }\cdots \dot{\mathbf{S}}_{%
\bm{\alpha}%
\nu }\cdots \dot{\mathbf{S}}_{%
\bm{\alpha}%
\lambda }\cdots \mathbf{S}_{%
\bm{\alpha}%
\beta _{k}}\right\vert \\
\frac{d^{4}}{dt^{4}}m_{%
\bm{\alpha}%
\bm{\beta}%
} &=&\sum_{\mu \in \mathbf{\beta }}\left\vert \mathbf{S}_{%
\bm{\alpha}%
\beta _{1}}\cdots \mathbf{S}_{%
\bm{\alpha}%
\mu }^{\left( 4\right) }\cdots \mathbf{S}_{%
\bm{\alpha}%
\beta _{k}}\right\vert +4\sum_{\mu \in \mathbf{\beta }}\left\vert \mathbf{S}%
_{%
\bm{\alpha}%
\beta _{1}}\cdots \dddot{\mathbf{S}}_{%
\bm{\alpha}%
\mu }\cdots \dot{\mathbf{S}}_{%
\bm{\alpha}%
\nu }\cdots \mathbf{S}_{%
\bm{\alpha}%
\beta _{k}}\right\vert +3%
\hspace{-1ex}%
\sum_{\mu \neq \nu \in 
\bm{\beta}%
}\left\vert \mathbf{S}_{%
\bm{\alpha}%
\beta _{1}}\cdots \ddot{\mathbf{S}}_{%
\bm{\alpha}%
\mu }\cdots \ddot{\mathbf{S}}_{%
\bm{\alpha}%
\nu }\cdots \mathbf{S}_{%
\bm{\alpha}%
\beta _{k}}\right\vert \\
&&+6%
\hspace{-1ex}%
\sum_{\mu \neq \nu \neq \lambda \in 
\bm{\beta}%
}\left\vert \mathbf{S}_{%
\bm{\alpha}%
\beta _{1}}\cdots \ddot{\mathbf{S}}_{%
\bm{\alpha}%
\mu }\cdots \dot{\mathbf{S}}_{%
\bm{\alpha}%
\nu }\cdots \dot{\mathbf{S}}_{%
\bm{\alpha}%
\lambda }\cdots \mathbf{S}_{%
\bm{\alpha}%
\beta _{k}}\right\vert +%
\hspace{-1ex}%
\sum_{\mu \neq \nu \neq \lambda \neq \rho \in 
\bm{\beta}%
}\left\vert \mathbf{S}_{%
\bm{\alpha}%
\beta _{1}}\cdots \dot{\mathbf{S}}_{%
\bm{\alpha}%
\mu }\cdots \dot{\mathbf{S}}_{%
\bm{\alpha}%
\nu }\cdots \dot{\mathbf{S}}_{%
\bm{\alpha}%
\lambda }\cdots \dot{\mathbf{S}}_{%
\bm{\alpha}%
\rho }\cdots \mathbf{S}_{%
\bm{\alpha}%
\beta _{k}}\right\vert \\
\frac{d^{5}}{dt^{5}}m_{%
\bm{\alpha}%
\bm{\beta}%
} &=&\sum_{\beta _{i}\in 
\bm{\beta}%
}\left\vert \mathbf{S}_{%
\bm{\alpha}%
\beta _{1}}\cdots \mathbf{S}_{%
\bm{\alpha}%
\beta _{i}}^{\left( 5\right) }\cdots \mathbf{S}_{%
\bm{\alpha}%
\beta _{k}}\right\vert +5\sum_{\beta _{i}\neq \beta _{j}\in 
\bm{\beta}%
}\left\vert \mathbf{S}_{%
\bm{\alpha}%
\beta _{1}}\cdots \mathbf{S}_{%
\bm{\alpha}%
\beta _{i}}^{\left( 4\right) }\cdots \dot{\mathbf{S}}_{%
\bm{\alpha}%
\beta _{j}}\cdots \mathbf{S}_{%
\bm{\alpha}%
\beta _{k}}\right\vert +10\sum_{\beta _{i}\neq \beta _{i}\in 
\bm{\beta}%
}\left\vert \mathbf{S}_{%
\bm{\alpha}%
\beta _{1}}\cdots \dddot{\mathbf{S}}_{%
\bm{\alpha}%
\beta _{i}}\cdots \ddot{\mathbf{S}}_{%
\bm{\alpha}%
\beta _{j}}\cdots \mathbf{S}_{%
\bm{\alpha}%
\beta _{k}}\right\vert \\
&&+10%
\hspace{-1ex}%
\sum_{\beta _{i}\neq \beta _{i}\neq \beta _{l}\in 
\bm{\beta}%
}%
\hspace{-0.5ex}%
\left\vert \mathbf{S}_{%
\bm{\alpha}%
\beta _{1}}\cdots \dddot{\mathbf{S}}_{%
\bm{\alpha}%
\beta _{i}}\cdots \dot{\mathbf{S}}_{%
\bm{\alpha}%
\beta _{j}}\cdots \dot{\mathbf{S}}_{%
\bm{\alpha}%
\beta _{l}}\cdots \mathbf{S}_{%
\bm{\alpha}%
\beta _{k}}\right\vert +10%
\hspace{-2ex}%
\sum_{\beta _{i}\neq \beta _{i}\neq \beta _{l}\neq \beta _{r}\in 
\bm{\beta}%
}\left\vert \mathbf{S}_{%
\bm{\alpha}%
\beta _{1}}\cdots \ddot{\mathbf{S}}_{%
\bm{\alpha}%
\beta _{i}}\cdots \dot{\mathbf{S}}_{%
\bm{\alpha}%
\beta _{j}}\cdots \dot{\mathbf{S}}_{%
\bm{\alpha}%
\beta _{l}}\cdots \dot{\mathbf{S}}_{%
\bm{\alpha}%
\beta _{r}}\cdots \mathbf{S}_{%
\bm{\alpha}%
\beta _{k}}\right\vert \\
&&+15%
\hspace{-1ex}%
\sum_{\beta _{i}\neq \beta _{i}\neq \beta _{l}\in 
\bm{\beta}%
}%
\hspace{-0.5ex}%
\left\vert \mathbf{S}_{%
\bm{\alpha}%
\beta _{1}}\cdots \ddot{\mathbf{S}}_{%
\bm{\alpha}%
\beta _{i}}\cdots \ddot{\mathbf{S}}_{%
\bm{\alpha}%
\beta _{j}}\cdots \dot{\mathbf{S}}_{%
\bm{\alpha}%
\beta _{l}}\cdots \mathbf{S}_{%
\bm{\alpha}%
\beta _{k}}\right\vert \ \ \ \ \ (\ast ) \\
&&+%
\hspace{-1ex}%
\sum_{\beta _{i}\neq \beta _{i}\neq \beta _{l}\neq \beta _{r}\neq \beta
_{s}\in 
\bm{\beta}%
}\left\vert \mathbf{S}_{%
\bm{\alpha}%
\beta _{1}}\cdots \dot{\mathbf{S}}_{%
\bm{\alpha}%
\beta _{i}}\cdots \dot{\mathbf{S}}_{%
\bm{\alpha}%
\beta _{j}}\cdots \dot{\mathbf{S}}_{%
\bm{\alpha}%
\beta _{l}}\cdots \dot{\mathbf{S}}_{%
\bm{\alpha}%
\beta _{r}}\cdots \dot{\mathbf{S}}_{%
\bm{\alpha}%
\beta _{s}}\cdots \mathbf{S}_{%
\bm{\alpha}%
\beta _{k}}\right\vert \ \ \ \ \ (\ast \ast ).
\end{eqnarray*}%
The apparent structure gives rise to a general expression. To this end, the
degrees of time derivatives of $\mathbf{S}_{%
\bm{\alpha}%
\beta _{j}},1\leq j\leq k$ are now indicated by a multi-index $\mathbf{a}%
_{k}=(a_{1},\ldots ,a_{k})\in {\mathbb{N}}^{k}$ so that for instance the
minor $\left\vert \mathbf{S}_{%
\bm{\alpha}%
\beta _{1}}\dddot{\mathbf{S}}_{%
\bm{\alpha}%
\beta _{2}}\dot{\mathbf{S}}_{%
\bm{\alpha}%
\beta _{3}}\mathbf{S}_{%
\bm{\alpha}%
\beta _{4}}\dot{\mathbf{S}}_{%
\bm{\alpha}%
\beta _{5}}\right\vert $ of order $k=5$ is written as $\left\vert \mathbf{S}%
_{%
\bm{\alpha}%
\beta _{1}}^{\left( a_{1}\right) }\mathbf{S}_{%
\bm{\alpha}%
\beta _{2}}^{\left( a_{2}\right) }\mathbf{S}_{%
\bm{\alpha}%
\beta _{3}}^{\left( a_{3}\right) }\mathbf{S}_{%
\bm{\alpha}%
\beta _{4}}^{\left( a_{4}\right) }\mathbf{S}_{%
\bm{\alpha}%
\beta _{5}}^{\left( a_{5}\right) }\right\vert $ with $\mathbf{a}_{k}=\left(
0,3,1,0,1\right) $. In the above expressions for $\frac{d^{\nu }}{dt^{\nu }}%
m_{%
\bm{\alpha}%
\bm{\beta}%
}$, it is $\left\vert \mathbf{a}_{k}\right\vert =a_{1}+\ldots +a_{k}=\nu $.
The minors in the sums contain repeated time derivatives of the same degree,
indicated by the same integers $a_{j}=a_{i}$. For instance, the sum $(\ast )$
contains $2!$ times the same term since $\ddot{\mathbf{S}}$ is repeated, and 
$(\ast \ast )$ contains $5!$ times the same summand due to the repetition of 
$\dot{\mathbf{S}}$. Generally, the number of repeated terms is $n_{i}!$,
where $n_{i}:=|\{a_{j}|a_{j}=i\}|$ is the number of times the $i$-th
derivative occurs. Not accounting for these repetitions, the coefficient
before a sum in the above expressions is the number of terms with the same
multi-index $\mathbf{a}_{k}$. This is the number of different permutations,
with repetition, of the indexes $\beta _{i}\in 
\bm{\beta}%
$, which is $\nu !/(a_{1}!\cdot \ldots \cdot a_{k}!)$, where $a_{i}$
indicates the number of times the index $\beta _{i}$ is repeated. The $\nu $%
-th time derivative thus attains the compact form%
\begin{equation}
\frac{d^{\nu }m_{%
\bm{\alpha}%
\bm{\beta}%
}}{dt^{\nu }}=\sum_{\left\vert \mathbf{a}_{k}\right\vert =\nu }\left\vert 
\mathbf{S}_{%
\bm{\alpha}%
\beta _{1}}^{\left( a_{1}\right) }\ \mathbf{S}_{%
\bm{\alpha}%
\beta _{2}}^{(a_{2})}\ \cdots \ \mathbf{S}_{%
\bm{\alpha}%
\beta _{k}}^{\left( a_{k}\right) }\right\vert \frac{\nu !}{\mathbf{a}%
_{k}!n_{1}!\cdots n_{\nu }!}.  \label{dmdt}
\end{equation}%
The division by $n_{1}!\cdots n_{\nu }!$ accounts for the fact the
permutations of terms with the same $a_{i}$ are already covered by the sum.
The necessary time derivatives of the joint screws can be determined using $%
\mathbf{S}_{%
\bm{\alpha}%
j}^{(i)}\equiv \mathrm{D}^{\left( i\right) }\mathbf{S}_{%
\bm{\alpha}%
j}$ in (\ref{DSi}) or by explicit evaluation via the partial derivatives in (%
\ref{dnS}). Notice that the relation (\ref{dmdt}) is applicable to a general
matrix.

\begin{remark}
The expression (\ref{dmdt}) could be amended in order to avoid multiple
occurrences of identical terms, so that the factor would become $\nu !/%
\mathbf{a}_{k}!$. To this end, the summation would have to be restricted
over indexes $\beta _{i}<\beta _{j}<\ldots <\beta _{l}$ whenever $%
a_{i}=a_{j}=\ldots =a_{l}$.
\end{remark}

The above derivation of (\ref{dmdt}) was not yet published in the
literature. It was, however, used in \cite{JMR2016} without a proof for
singularity analysis of linkages.

\begin{remark}
The time derivatives can also be determined in terms of partial derivatives
of the minors. The explicit relation for the partial derivatives of the
minors of arbitrary degree was presented in \cite{JMR2018}. For the special
case of the determinant of the Jacobian for spatial manipulators, i.e. $n=|%
\bm{\alpha}%
|=|%
\bm{\beta}%
|=6$, partial derivatives up to degree 3 were presented in \cite{Karger1996}.
\end{remark}

\subsection{Series Expansion of Minors of the Geometric Jacobian%
\label{secSeriesMinor}%
}

The $%
\bm{\alpha}%
\bm{\beta}%
$-minor of order $k$ of the Jacobian can be expanded in a Taylor series at $%
\mathbf{q}\in {\mathbb{V}}^{n}$

\begin{equation*}
m_{%
\bm{\alpha}%
\bm{\beta}%
}%
\hspace{-0.5ex}%
\left( \mathbf{q}+\mathbf{x}\right) =m_{%
\bm{\alpha}%
\bm{\beta}%
}%
\hspace{-0.5ex}%
\left( \mathbf{q}\right) +\mathrm{d}m_{%
\bm{\alpha}%
\bm{\beta}%
,\mathbf{q}}%
\hspace{-0.5ex}%
\left( \mathbf{x}\right) 
\hspace{-0.5ex}%
+%
\hspace{-0.5ex}%
\frac{1}{2}\mathrm{d}^{2}m_{%
\bm{\alpha}%
\bm{\beta}%
,\mathbf{q}}%
\hspace{-0.5ex}%
\left( \mathbf{x}\right) 
\hspace{-0.5ex}%
+%
\hspace{-0.5ex}%
\ldots 
\hspace{-0.5ex}%
+%
\hspace{-0.5ex}%
\frac{1}{\nu !}\mathrm{d}^{\nu }m_{%
\bm{\alpha}%
\bm{\beta}%
,\mathbf{q}}%
\hspace{-0.5ex}%
\left( \mathbf{x}\right) .
\end{equation*}

The relation (\ref{dmdt}) can be immediately carried over to the
differentials, which yields%
\begin{equation}
\mathrm{d}^{i}m_{%
\bm{\alpha}%
\bm{\beta}%
,\mathbf{q}}=\sum_{\left\vert \mathbf{a}_{k}\right\vert =i}\left\vert 
\mathrm{d}^{a_{1}}\mathbf{S}_{%
\bm{\alpha}%
\beta _{1},\mathbf{q}}\left( \mathbf{x}\right) \ \mathrm{d}^{a_{2}}\mathbf{S}%
_{%
\bm{\alpha}%
\beta _{2},\mathbf{q}}\left( \mathbf{x}\right) \ \cdots \ \mathrm{d}^{a_{k}}%
\mathbf{S}_{%
\bm{\alpha}%
\beta _{k},\mathbf{q}}\left( \mathbf{x}\right) \right\vert \frac{i!}{\mathbf{%
a}_{k}!n_{1}!\cdots n_{i}!}.  \label{dm}
\end{equation}%
This can be evaluated using the relations (\ref{dkSii}) for the
differentials of instantaneous joint screw coordinates.

\subsection{Applications}

\subsubsection{Local Analysis of Smooth Motions with certain Rank}

C-space singularities of linkages with kinematic loops are characterized by
a rank deficient constraint Jacobian. A deeper understanding of the
singularities of a linkage is gained by investigating possible motions with
certain rank.

Denote with $L_{k}$ the subvariety of $V$ where rank of $\mathbf{J}$ is less
than $k$. The Jacobian $\mathbf{J}$ has rank less than $k$ iff all $k$%
-minors vanish: $m_{%
\bm{\alpha}%
\bm{\beta}%
}=0,\left\vert 
\bm{\alpha}%
\right\vert =\left\vert 
\bm{\beta}%
\right\vert =k$, where $\left\vert 
\bm{\alpha}%
\right\vert $ is the cardinality (number of elements) of the set $%
\bm{\alpha}%
$. This gives rise to the following definition%
\begin{equation}
L_{k}\,:{=}\left\{ \mathbf{q}\in {\mathbb{V}}^{n}|f\left( \mathbf{q}\right) =%
\mathbf{I},m_{%
\bm{\alpha}%
\bm{\beta}%
}(\mathbf{q})=0\right. \left. \forall 
\bm{\alpha}%
\subseteq \{1,\ldots ,6\},%
\bm{\beta}%
\subseteq \{1,\ldots ,n\},\left\vert 
\bm{\alpha}%
\right\vert 
\hspace{-0.5ex}%
=%
\hspace{-0.5ex}%
\left\vert 
\bm{\beta}%
\right\vert 
\hspace{-0.5ex}%
=%
\hspace{-0.3ex}%
k\right\}  \label{Rk2}
\end{equation}

A finite motion $\mathbf{q}\left( t\right) $ in $L_{k}$, i.e. where $\mathrm{%
rank}~\mathbf{J}<k$, satisfies all higher-order constraints (\ref%
{HighOrderConstr}) and all time derivatives of $m_{%
\bm{\alpha}%
\bm{\beta}%
}$ vanish. The set of tangents to curves through $\mathbf{q}\in V$ where $%
\mathrm{rank}~\mathbf{J}<k$ forms the \emph{kinematic tangent cone to }$%
L_{k} $, denoted with ${C_{\mathbf{q}}^{\text{K}}}L_{k}\subseteq {C_{\mathbf{%
q}}^{\text{K}}V}$. To simplify notation, define the functions%
\begin{equation}
M_{%
\bm{\alpha}%
\bm{\beta}%
}^{\left( 1\right) }%
\hspace{-0.6ex}%
\left( \mathbf{q},\dot{\mathbf{q}}\right) {:=}\frac{d}{dt}m_{%
\bm{\alpha}%
\bm{\beta}%
}(\mathbf{q}),\ M_{%
\bm{\alpha}%
\bm{\beta}%
}^{\left( 2\right) }%
\hspace{-0.6ex}%
\left( \mathbf{q},\dot{\mathbf{q}},\ddot{\mathbf{q}}\right) {:=}\frac{d^{2}}{%
dt^{2}}m_{%
\bm{\alpha}%
\bm{\beta}%
}(\mathbf{q}),\ldots ,\ M_{%
\bm{\alpha}%
\bm{\beta}%
}^{\left( i\right) }%
\hspace{-0.4ex}%
(\mathbf{q},\dot{\mathbf{q}},\ldots ,\mathbf{q}^{\left( i\right) }){:=}\frac{%
d^{i}}{dt^{i}}m_{%
\bm{\alpha}%
\bm{\beta}%
}(\mathbf{q}).  \label{Mab}
\end{equation}%
The kinematic tangent cone to $L_{k}$ is then determined by the sequence%
\begin{equation}
{C_{\mathbf{q}}^{\text{K}}}L_{k}=K_{\mathbf{q}}^{k,\kappa }\subset \ldots
\subset K_{\mathbf{q}}^{k,3}\subset K_{\mathbf{q}}^{k,2}\subset {K_{\mathbf{q%
}}^{k,1}}  \label{CqLk}
\end{equation}%
with%
\begin{equation}
\begin{array}{ll}
K_{\mathbf{q}}^{k,i}:=\{\mathbf{x}|\exists \mathbf{y},\mathbf{z},\ldots \in {%
\mathbb{R}}^{n}: & H^{\left( 1\right) }%
\hspace{-0.6ex}%
\left( \mathbf{q},\mathbf{x}\right) =H^{\left( 2\right) }%
\hspace{-0.6ex}%
\left( \mathbf{q},\mathbf{x},\mathbf{y}\right) =\ldots \mathbf{=}H^{\left(
i\right) }%
\hspace{-0.6ex}%
\left( \mathbf{q},\mathbf{x},\mathbf{y},\mathbf{z,\ldots }\right) =\mathbf{0}%
, \\ 
& M_{%
\bm{\alpha}%
\bm{\beta}%
}^{\left( 1\right) }%
\hspace{-0.6ex}%
\left( \mathbf{q},\mathbf{x}\right) =M_{%
\bm{\alpha}%
\bm{\beta}%
}^{\left( 2\right) }%
\hspace{-0.6ex}%
\left( \mathbf{q},\mathbf{x},\mathbf{y}\right) =\ldots =M_{%
\bm{\alpha}%
\bm{\beta}%
}^{\left( i\right) }%
\hspace{-0.6ex}%
\left( \mathbf{q},\mathbf{x},\mathbf{y},\mathbf{z,\ldots }\right) =0, \\ 
& \forall 
\bm{\alpha}%
\subseteq \{1,\ldots ,6\},%
\bm{\beta}%
\subseteq \{1,\ldots ,n\},\left\vert 
\bm{\alpha}%
\right\vert 
\hspace{-0.5ex}%
=%
\hspace{-0.5ex}%
\left\vert 
\bm{\beta}%
\right\vert 
\hspace{-0.5ex}%
=k\}.%
\end{array}
\label{Kki}
\end{equation}%
In general, each $K_{\mathbf{q}}^{k,i}$ is a cone. The sequence terminates
with a finite $\kappa $. The latter indicates the order of differential
motions that do not correspond to finite motions with rank less than $k$.

The higher-order analysis of smooth motions with rank deficient Jacobian has
only been reported briefly in \cite{JMR2018}.

\paragraph{Example 1 (cont.): 4C-Linkage with a shaky motion mode}

The joint screw coordinate vectors in the reference configuration are given
in (\ref{Y4C}). At $\mathbf{q}_{0}=\mathbf{0}$ it is $\mathrm{rank}~\mathbf{J%
}\left( \mathbf{q}_{0}\right) =4$, and thus $\mathbf{q}_{0}\in L_{k},k\geq 5$%
. Since the maximal rank of $\mathbf{J}$ is 6, only $L_{5}$ and $L_{6}$ need
to be analyzed.

All $M_{%
\bm{\alpha}%
\bm{\beta}%
}^{\left( 1\right) }%
\hspace{-0.6ex}%
\left( \mathbf{q}_{0},\mathbf{x}\right) ,\left\vert 
\bm{\alpha}%
\right\vert =\left\vert 
\bm{\beta}%
\right\vert =6$ vanish, so that $K_{\mathbf{q}_{0}}^{6,1}=K_{\mathbf{q}%
_{0}}^{1}$. The non-trivial second derivatives of the 6-minors are%
\begin{equation}
\{M_{%
\bm{\alpha}%
\bm{\beta}%
}^{\left( 2\right) }%
\hspace{-0.6ex}%
\left( \mathbf{q}_{0},\mathbf{x}\right) ,\left\vert 
\bm{\alpha}%
\right\vert =\left\vert 
\bm{\beta}%
\right\vert
=6\}=%
\{-2x_{3}^{2},2x_{3}^{2},-2x_{3}x_{5},2x_{3}x_{5},-2x_{5}^{2},2x_{5}^{2},2x_{4}x_{5}-2x_{3}x_{6},-2x_{4}x_{5}+2x_{3}x_{6}\}.
\end{equation}%
This yields%
\begin{equation}
K_{\mathbf{q}_{0}}^{6,2}=\{|\mathbf{x}=(0,t,0,s,0,-t,0,-s),s,t\in {\mathbb{R}%
}\}\in {\mathbb{R}}^{8}.  \label{K62}
\end{equation}%
Proceeding for higher derivatives shows that $C_{\mathbf{q}_{0}}^{\text{K}%
}L_{6}=K_{\mathbf{q}_{0}}^{6,i},i\geq 2$.

The non-trivial derivatives of the 5-minors are%
\begin{equation}
\{M_{%
\bm{\alpha}%
\bm{\beta}%
}^{\left( 1\right) }%
\hspace{-0.6ex}%
\left( \mathbf{q}_{0},\mathbf{x}\right) ,\left\vert 
\bm{\alpha}%
\right\vert =\left\vert 
\bm{\beta}%
\right\vert =5\}=\{-x_{3},x_{3},-x_{4},x_{4},-x_{5},x_{5},-x_{6},x_{6}\}.
\end{equation}%
Therewith follows that $K_{\mathbf{q}_{0}}^{5,1}=\{\mathbf{0}\}\in {\mathbb{R%
}}^{8}$. Since the sequence (\ref{CqLk}) is non-increasing, it follows that $%
C_{\mathbf{q}_{0}}^{\text{K}}L_{5}=\{\mathbf{0}\}$. This shows that there is
no smooth finite motion with $\mathrm{rank}~\mathbf{J}=4$, so that in any
point of the neighborhood of $\mathbf{q}_{0}$ the rank (thus the
differential mobility) increases. There are 2-dimensional smooth finite
motions through $\mathbf{q}_{0}$ with $\mathrm{rank}~\mathbf{J}\leq 5$ whose
tangents are given by $C_{\mathbf{q}_{0}}^{\text{K}}L_{6}$. Since $\mathbf{J}
$ has rank 4 only at $\mathbf{q}_{0}$, these are 2-dim motions with $\mathrm{%
rank}~\mathbf{J}=5$. Because of $K_{\mathbf{q}_{0}}^{2\left( II\right) }=C_{%
\mathbf{q}_{0}}^{\text{K}}L_{6}$, in (\ref{4CK2}), the corresponding motion
mode II with $\mathrm{rank}~\mathbf{J}=5$ (Fig. \ref{fig4CModes}b) the
linkage is shaky since in this motion mode the local finite DOF is $\delta _{%
\mathrm{loc}}\left( \mathbf{q}_{0}\right) =2$ while the differential
(instantaneous) DOF is $\delta _{\mathrm{diff}}\left( \mathbf{q}_{0}\right)
=n-\mathrm{rank}~\mathbf{J}\left( \mathbf{q}_{0}\right) =3$. This 2-dim
manifold of configurations with rank 5 is characterized by $%
q_{1}=q_{3}=q_{5}=q_{7}=0$. Thus the 4C linkage is underconstrained. It
should be noticed that this could not be identified by a first-order
analysis of the singularity $\mathbf{q}_{0}$ nor by the higher-order
analysis of the finite mobility (sec. \ref{sec4CKinCone}) as pursued in \cite%
{LopezCustodio2017}.

\subsubsection{Local Approximation of the Set of Configurations with certain
Rank}

In addition to bifurcations of motion branches, there may be non-smooth
motions with rank-deficient Jacobian. Such 'stationary singularities' are
e.g. manifested as cusps in the c-space $V$. Local analysis requires an
approximation of the local geometry of $L_{k}$ at $\mathbf{q}$. The
subvariety of configurations with $\mathrm{rank}~\mathbf{J}<k$ is 
\begin{equation}
L_{k,\mathbf{q}}^{i}=\{\mathbf{x}\in {\mathbb{R}}^{n}|\mathrm{d}f_{\mathbf{q}%
}\left( \mathbf{x}\right) +\frac{1}{2}\mathrm{d}^{2}f_{\mathbf{q}}\left( 
\mathbf{x}\right) +\ldots +\frac{1}{i!}\mathrm{d}^{i}f_{\mathbf{q}}\left( 
\mathbf{x}\right) =\mathbf{0},\mathrm{d}m_{%
\bm{\alpha}%
\bm{\beta}%
,\mathbf{q}}%
\hspace{-0.5ex}%
\left( \mathbf{x}\right) 
\hspace{-0.5ex}%
+%
\hspace{-0.5ex}%
\frac{1}{2}\mathrm{d}^{2}m_{%
\bm{\alpha}%
\bm{\beta}%
,\mathbf{q}}%
\hspace{-0.5ex}%
\left( \mathbf{x}\right) 
\hspace{-0.5ex}%
+%
\hspace{-0.5ex}%
\ldots 
\hspace{-0.5ex}%
+%
\hspace{-0.5ex}%
\frac{1}{i!}\mathrm{d}^{i}m_{%
\bm{\alpha}%
\bm{\beta}%
,\mathbf{q}}%
\hspace{-0.5ex}%
\left( \mathbf{x}\right) =0,|%
\bm{\alpha}%
|=|%
\bm{\beta}%
|=k\}  \label{Lknu}
\end{equation}%
with the differentials of $f$ and $m_{%
\bm{\alpha}%
\bm{\beta}%
}$ in (\ref{dkf2}) and (\ref{dm}), respectively. There is a neighborhood $%
U\left( \mathbf{q}\right) $ and order $\kappa $ so that $L_{k,\mathbf{q}%
}^{\kappa }\cap U\left( \mathbf{q}\right) =L_{k}\cap U\left( \mathbf{q}%
\right) $.

This gives rise to a stratification of $V$ according to the rank.

\paragraph{Example 1 (cont.): 4C-Linkage with a shaky motion mode}

The analysis yields $L_{k,\mathbf{q}_{0}}^{6}=K_{\mathbf{q}%
_{0}}^{6,i},i=1,2,3,\ldots $, thus the finite motions with rank 5 are
smooth. The reader is referred to the provided Mathematica notebook \cite%
{MendeleyDataset} for details.

\section{Higher-Order Inverse Kinematics%
\label{secInvKinArm}%
}

\subsection{Inverse kinematics of a robotic arm}

\subsubsection{Recursive higher-order inverse kinematics}

A robotic arm is a serial kinematic chain with an EE attached at the
terminal link $n$. The $k$th-order inverse kinematics problem consists in
finding the $k$th time derivatives of the joint variables $\mathbf{q}\left(
t\right) $ for a given EE motion, i.e. given EE pose, EE twist, and its time
derivatives up to order $k-1$.

With (\ref{VsiJ}), the spatial twist of the terminal link $n$ is $\mathbf{V}%
_{n}^{\text{s}}=\mathbf{J}_{n}^{\text{s}}\dot{\mathbf{q}}$. For the sake of
simplicity, the Jacobian is denoted with $\mathbf{J}$.

\begin{assumption}
In the following, the robotic arm is assumed to be non-redundant. That is,
the DOF (i.e. the number $n$ of joint variables) is equal to the dimension
of the image space of the KM, so that the Jacobian $\mathbf{J}$ is a full
rank $n\times n$ matrix.
\end{assumption}

With $\mathrm{D}^{(k)}\mathbf{V}_{n}^{\text{s}}=\mathrm{D}^{(k)}\mathsf{S}%
_{i}$, the expression (\ref{DSik}) for time derivatives of the twist of
terminal body $n$ can be written as%
\begin{equation}
\mathrm{D}^{(k)}\mathbf{V}_{n}^{\text{s}}=\mathbf{J\left( \mathbf{q}\right) q%
}^{(k+1)}+\sum_{i\leq n}\sum_{l=1}^{k}\tbinom{k}{l}\mathrm{D}^{(l)}\mathbf{S}%
_{i}\left( \mathbf{q}\right) q_{i}^{(k-l+1)}.  \label{DkVn}
\end{equation}%
With the full-rank Jacobian (except at singularities), (\ref{DkVn}) can be
solved as%
\begin{equation}
\mathbf{q}^{(k)}=\mathbf{J}^{-1}%
\hspace{-0.5ex}%
\left( \mathbf{q}\right) 
\Big%
(\mathrm{D}^{(k-1)}\mathbf{V}_{n}^{\text{s}}-\sum_{i\leq n}\sum_{l=1}^{k-1}%
\tbinom{k-1}{l}\mathrm{D}^{(l)}\mathbf{S}_{i}\left( \mathbf{q}\right)
q_{i}^{(k-l)}%
\Big%
).  \label{qk}
\end{equation}%
This is the $k$th-order inverse kinematics solution, which gives the $k$th
time derivative $\mathbf{q}^{(k)}$, when given the EE twist $\mathbf{V}_{n}^{%
\text{s}}$ and the joint variables as well as their respective time
derivatives of up to order $k-1$. The configuration $\mathbf{q}$ is known
from (numerically) solving the geometric inverse kinematics problem, i.e.
solving $\mathbf{C}_{n}=f_{n}\left( \mathbf{q}\right) $ for $\mathbf{q}$ for
given $\mathbf{C}_{n}$.

\subsubsection{Higher-order inverse kinematics algorithm for a robotic arm%
\label{secInvKin}%
}

The expression (\ref{qk}) involves time derivatives of the joint screw
coordinates, given by (\ref{DSi}), of the twists of all bodies of the
kinematic chain, as well as time derivatives of $\mathbf{q}$. Therefore, (%
\ref{qk}) must be evaluated consecutively for increasing order. Starting
with the given configuration $\mathbf{q}$ and $\mathbf{V}_{n}^{\text{s}}$,
the first-order inverse kinematics is solved to obtain $\dot{\mathbf{q}}$.
This is then propagated through the kinematic chain in order to obtain $%
\mathbf{V}_{i}^{\text{s}},i<n$. Next, with the known $\mathbf{q},\dot{%
\mathbf{q}},\mathbf{V}_{i}^{\text{s}},i=1,\ldots ,n$, the second-order
inverse kinematics is solved for $\ddot{\mathbf{q}}$, which is then
propagated via forward kinematics to obtain $\dot{\mathbf{V}}_{i}^{\text{s}%
},i<n$. These steps are repeated until the $k$th-order solution is found.
This is summarized in the following inverse kinematics algorithm:%
\vspace{2ex}%

\begin{tabular}{l}
\hline
\textbf{Higher-order inverse kinematics algorithm:}%
\vspace{1ex}
\\ 
\hspace{1ex}%
Input: $\mathbf{q},\mathbf{V}_{n}^{\text{s}},\dot{\mathbf{V}}_{n}^{\text{s}},%
\ddot{\mathbf{V}}_{n}^{\text{s}},\ldots ,\mathrm{D}^{(k-1)}\mathbf{V}_{n}^{%
\text{s}}$%
\vspace{1ex}
\\ 
\hspace{2ex}%
FOR $r=1,\ldots ,k$ \\ 
\hspace{2ex}%
\begin{tabular}{lll}
1) & Evaluate (\ref{qk}) to get $\mathbf{q}^{(r)}$ & ($r$th-order inverse
kinematics solution of manipulator)%
\vspace{1ex}
\\ 
2) & Evaluate (\ref{DSik}) to get $\mathrm{D}^{(r-1)}\mathbf{V}_{i}^{\text{s}%
},i<n$ & ($r$th-order forward kinematics solution of linkage)%
\vspace{1ex}%
\end{tabular}
\\ 
\hspace{2ex}%
END%
\vspace{1ex}
\\ 
\hspace{1ex}%
Output: $\dot{\mathbf{q}},\ldots ,\mathbf{q}^{\left( k\right) }$ (primary
results), $\mathbf{V}_{i}^{\text{s}},,\dot{\mathbf{V}}_{i}^{\text{s}},\ddot{%
\mathbf{V}}_{i}^{\text{s}},\ldots ,\mathrm{D}^{(k-1)}\mathbf{V}_{i}^{\text{s}%
},i=1,\ldots ,n$ (secondary results) \\ \hline
\end{tabular}%
\vspace{2ex}%

The individual recursive runs 1) and 2) have complexity $O\left( n\right) $.
The overall complexity of the algorithm is dictated by the inversion of the
Jacobian. This can be alleviated, for wrist-partitioned robotic arms (which
are predominately used in praxis), since then the inversion of the $6\times
6 $ Jacobian splits into the inversion of three $3\times 3$ submatrices \cite%
{Angeles2007}. Alternatively, the linear system $\mathbf{Jq}^{(k)}=\mathrm{%
rhs}.$ in (\ref{qk}) can be solved using a tailored numerical solution
scheme.

For evaluation of $\mathrm{D}^{(r-1)}\mathbf{V}_{i}^{\text{s}}$ in step 2),
the derivatives $\mathrm{D}^{(l)}\mathsf{S}_{i},l\leq r-1$ already obtained
in the preceding evaluations of (\ref{DSik}) are reused.

\subsubsection{Explicit expressions for low orders}

For low order (see section \ref{secInvKinApp}a), it can be helpful to
explicitly roll out the relation (\ref{qk}). Up to order 4, this yields 
\begin{eqnarray}
\dot{\mathbf{q}} &=&\mathbf{J}^{-1}\mathbf{V}_{n}^{\text{s}}  \label{qdots1}
\\
\ddot{\mathbf{q}} &=&\mathbf{J}^{-1}%
\Big%
(\dot{\mathbf{V}}_{n}^{\text{s}}-\sum_{i\leq n}\dot{\mathbf{S}}_{i}\dot{q}%
_{i}%
\Big%
)  \notag \\
&=&\mathbf{J}^{-1}%
\Big%
(\dot{\mathbf{V}}_{n}^{\text{s}}-\sum_{i\leq n}\dot{q}_{i}[\mathbf{V}_{i}^{%
\text{s}},\mathbf{S}_{i}]%
\Big%
)=\mathbf{J}^{-1}%
\Big%
(\dot{\mathbf{V}}_{n}^{\text{s}}-\sum_{i\leq n}\dot{q}_{i}\mathbf{ad}_{%
\mathbf{V}_{i}^{\text{s}}}\mathbf{S}_{i}%
\Big%
)  \label{qdots2} \\
\dddot{\mathbf{q}} &=&\mathbf{J}^{-1}%
\Big%
(\ddot{\mathbf{V}}_{n}^{\text{s}}-\sum_{i\leq n}(2\dot{\mathbf{S}}_{i}\ddot{q%
}_{i}+\ddot{\mathbf{S}}_{i}\dot{q}_{i})%
\Big%
)  \notag \\
&=&\mathbf{J}^{-1}%
\Big%
(\ddot{\mathbf{V}}_{n}^{\text{s}}-\sum_{i\leq n}\left( 2\ddot{q}_{i}[\mathbf{%
V}_{i}^{\text{s}},\mathbf{S}_{i}]+\dot{q}_{i}\left( [\dot{\mathbf{V}}_{i}^{%
\text{s}},\mathbf{S}_{i}]+[\mathbf{V}_{i}^{\text{s}},[\mathbf{V}_{i}^{\text{s%
}},\mathbf{S}_{i}]]\right) \right) 
\Big%
)=\mathbf{J}^{-1}%
\Big%
(\ddot{\mathbf{V}}_{n}^{\text{s}}-\sum_{i\leq n}%
\big%
(2\ddot{q}_{i}\mathbf{ad}_{\mathbf{V}_{i}^{\text{s}}}+\dot{q}_{i}(\mathbf{ad}%
_{\dot{\mathbf{V}}_{i}^{\text{s}}}+\mathbf{ad}_{\mathbf{V}_{i}^{\text{s}%
}}^{2})%
\big%
)\mathbf{S}_{i}%
\Big%
)  \label{qdots3} \\
\ddot{\ddot{\mathbf{q}}} &=&\mathbf{J}^{-1}%
\Big%
(\dddot{\mathbf{V}}_{n}^{\text{s}}-\sum_{i\leq n}(\dot{\mathbf{S}}_{i}\dddot{%
q}_{i}+3\ddot{\mathbf{S}}_{i}\ddot{q}_{i}+\dddot{\mathbf{S}}_{i}\dot{q}_{i})%
\Big%
)  \notag \\
&=&\mathbf{J}^{-1}%
\Big%
(\dddot{\mathbf{V}}_{n}^{\text{s}}-\sum_{i\leq n}\left( \dddot{q}_{i}[%
\mathbf{V}_{i}^{\text{s}},\mathbf{S}_{i}]+3\ddot{q}_{i}\left( [\dot{\mathbf{V%
}}_{i}^{\text{s}},\mathbf{S}_{i}]+[\mathbf{V}_{i}^{\text{s}},[\mathbf{V}%
_{i}^{\text{s}},\mathbf{S}_{i}]]\right) +\dot{q}_{i}\left( [\ddot{\mathbf{V}}%
_{i}^{\text{s}},\mathbf{S}_{i}]+2[\dot{\mathbf{V}}_{i}^{\text{s}},[\mathbf{V}%
_{i}^{\text{s}},\mathbf{S}_{i}]]+[\mathbf{V}_{i}^{\text{s}},[\dot{\mathbf{V}}%
_{i}^{\text{s}},\mathbf{S}_{i}]]+[\mathbf{V}_{i}^{\text{s}},[\mathbf{V}_{i}^{%
\text{s}},[\mathbf{V}_{i}^{\text{s}},\mathbf{S}_{i}]]]\right) \right) 
\Big%
)  \notag \\
&=&\mathbf{J}^{-1}%
\Big%
(\dddot{\mathbf{V}}_{n}^{\text{s}}-\sum_{i\leq n}\left( \dddot{q}_{i}\mathbf{%
ad}_{\mathbf{V}_{i}^{\text{s}}}+3\ddot{q}_{i}(\mathbf{ad}_{\dot{\mathbf{V}}%
_{i}^{\text{s}}}+\mathbf{ad}_{\mathbf{V}_{i}^{\text{s}}}^{2})+\dot{q}_{i}(%
\mathbf{ad}_{\ddot{\mathbf{V}}_{i}^{\text{s}}}+2\mathbf{ad}_{\dot{\mathbf{V}}%
_{i}^{\text{s}}}\mathbf{ad}_{\mathbf{V}_{i}^{\text{s}}}+\mathbf{ad}_{\mathbf{%
V}_{i}^{\text{s}}}\mathbf{ad}_{\dot{\mathbf{V}}_{i}^{\text{s}}}+\mathbf{ad}_{%
\mathbf{V}_{i}^{\text{s}}}^{3})\right) \mathbf{S}_{i}%
\Big%
).  \label{qdots4}
\end{eqnarray}%
These are evaluated together with (\ref{Vsdotrec})-(\ref{Vs3dotrec}). First (%
\ref{qdots1}) is evaluated, which delivers the joint velocities for given EE
twist. Then (\ref{Vsdotrec}) is used to propagate the velocity and to
compute the twists of all links. These are used in (\ref{qdots2}) to obtain
the joint accelerations, which are then used to compute the accelerations of
all links with (\ref{Vs2dotrec}). This is continued analogously for the
higher derivatives. The complexity \thinspace for evaluating the terms in
brackets in (\ref{qdots2}-\ref{qdots4}) is still $O\left( n\right) $.

\subsubsection{Applications%
\label{secInvKinApp}%
}

\paragraph{a) Flatness-based control of robots with elastic actuators}

It is known that robotic arms with series elastic actuators (SEA) are
differentially flat control systems, which means that the derivatives of the
joint variables (inputs) can be expressed as function of the derivatives of
the EE twist (outputs), see sec. \ref{secInvDyn}b). For robotic arms
consisting of rigid links and SEA, the vector relative degree of $\mathbf{q}$
is 4 \cite{deLuca1998,PalliMelchiorriDeLuca2008}. Thus, derivatives of $%
\mathbf{q}$ up to 4th-order are necessary for application of flatness-based
control algorithms (so that the second time derivative of the EOM are
required, see sec. \ref{secInvDyn}).

The trajectory planning is usually carried out in the robot workspace, so
that the derivatives $\dot{\mathbf{q}},\ddot{\mathbf{q}},\dddot{\mathbf{q}},%
\ddot{\ddot{\mathbf{q}}}$ are not given a priori but must be determined from
derivatives of the EE twist up to 3rd-order by solving the inverse
kinematics problem. This problem is not discussed in any of the relevant
publications, rather $\dot{\mathbf{q}},\ddot{\mathbf{q}},\dddot{\mathbf{q}},%
\ddot{\ddot{\mathbf{q}}}$ are assumed to be given.

\paragraph{b) Control of robots with structural flexibility}

Controlling inherently flexible robots (such as elastic lightweight arms)
must ensure that the trajectories are sufficiently smooth so to avoid
excitation of vibrations. To this end, the higher-order time derivatives of
the joint coordinates up to a certain order $k$ must be bounded. This
applies in particular to the optimal control, e.g. time optimal control, of
robotic arms. Within the trajectory planning, by numerical solution of the
optimal control problem, the constraints $\mathbf{q}_{\min }^{\left(
k\right) }\leq \mathbf{q}^{\left( k\right) }\left( t\right) \leq \mathbf{q}%
_{\max }^{\left( k\right) }$ must be included, with a selected order $k$. It
should be mentioned that for redundant manipulators the solution needs to be
amended in order to account for possible self motions \cite{ReiterTII2018}.

\subsection{Generalized Inverse Kinematics of a Kinematic Chain}

When the motions of all bodies in the kinematic chain are known, the inverse
kinematics problem is to determine the corresponding joint motions. This is
the counterpart of the forward kinematic problem of a linkage, which is to
determine the motion of all bodies for given joint motion.

The equation (\ref{Vsrec}) is an overdetermined system in $\dot{q}_{i}$, and
so are (\ref{Vsdotrec})-(\ref{Vs3dotrec}) in $\ddot{q}_{i},\dddot{q}_{i}$,
and $\ddot{\ddot{q}}_{i}$, respectively. They possesses unique solutions
that give rise to the following recursive relations

\begin{eqnarray}
\dot{q}_{i} &=&\mathbf{S}_{i}^{+}\left( \mathbf{V}_{i}^{\text{s}}-\mathbf{V}%
_{i-1}^{\text{s}}\right)  \label{InvKinRec} \\
\ddot{q}_{i} &=&\mathbf{S}_{i}^{+}\left( \dot{\mathbf{V}}_{i}^{\text{s}}-%
\dot{\mathbf{V}}_{i-1}^{\text{s}}-\dot{q}_{i}\mathbf{ad}_{\mathbf{V}_{i}^{%
\text{s}}}\mathbf{S}_{i}\right)  \label{InvKinRec2} \\
\dddot{q}_{i} &=&\mathbf{S}_{i}^{+}\left( \ddot{\mathbf{V}}_{i}^{\text{s}}-%
\ddot{\mathbf{V}}_{i-1}^{\text{s}}-\left( 2\ddot{q}_{i}\mathbf{ad}_{\mathbf{V%
}_{i}^{\text{s}}}+\dot{q}_{i}(\mathbf{ad}_{\dot{\mathbf{V}}_{i}^{\text{s}}}+%
\mathbf{ad}_{\mathbf{V}_{i}^{\text{s}}}^{2})\right) \mathbf{S}_{i}\right)
\label{InvKinRec3} \\
\ddot{\ddot{q}}_{i} &=&\mathbf{S}_{i}^{+}\left( \dddot{\mathbf{V}}_{i}^{%
\text{s}}-\dddot{\mathbf{V}}_{i-1}^{\text{s}}-\left( 3\dddot{q}_{i}\mathbf{ad%
}_{\mathbf{V}_{i}^{\text{s}}}+3\ddot{q}_{i}(\mathbf{ad}_{\dot{\mathbf{V}}%
_{i}^{\text{s}}}+\mathbf{ad}_{\mathbf{V}_{i}^{\text{s}}}^{2})+\dot{q}_{i}(%
\mathbf{ad}_{\ddot{\mathbf{V}}_{i}^{\text{s}}}+2\mathbf{ad}_{\dot{\mathbf{V}}%
_{i}^{\text{s}}}\mathbf{ad}_{\mathbf{V}_{i}^{\text{s}}}+\mathbf{ad}_{\mathbf{%
V}_{i}^{\text{s}}}\mathbf{ad}_{\dot{\mathbf{V}}_{i}^{\text{s}}}+\mathbf{ad}_{%
\mathbf{V}_{i}^{\text{s}}}^{3})\right) \mathbf{S}_{i}\right)
\label{InvKinRec4}
\end{eqnarray}%
with $\mathbf{S}_{i}^{+}=\mathbf{S}_{i}^{T}/\left( \mathbf{S}_{i}^{T}\mathbf{%
S}_{i}\right) =\mathbf{S}_{i}^{T}/\left\Vert \mathbf{S}_{i}\right\Vert ^{2}$%
, where $\left\Vert \mathbf{S}_{i}\right\Vert ^{2}=\mathbf{S}_{i}^{T}\mathbf{%
S}_{i}=\left\Vert 
\bm{\xi}%
_{i}\right\Vert ^{2}+\left\Vert 
\bm{\eta}%
_{i}\right\Vert ^{2}$.

The explicit form of the recursive expressions can be simplified. Noticing,
with (\ref{ad}) and $\mathbf{S}_{i}=\left( 
\bm{\xi}%
_{i},%
\bm{\eta}%
_{i}\right) ,\mathbf{V}_{i}^{\text{s}}=\left( 
\bm{\omega}%
_{i}^{\text{s}},\mathbf{v}_{i}^{\text{s}}\right) $, that%
\begin{eqnarray*}
-\mathbf{S}_{i}^{T}\mathbf{ad}_{\mathbf{V}_{i}^{\text{s}}}\mathbf{S}_{i} &=&%
\mathbf{S}_{i}^{T}\mathbf{ad}_{\mathbf{S}_{i}}\mathbf{V}_{i}^{\text{s}%
}=\left( 
\begin{array}{cc}
\bm{\xi}%
_{i}^{T}\widetilde{%
\bm{\xi}%
}_{i}+%
\bm{\eta}%
_{i}^{T}\widetilde{%
\bm{\eta}%
}_{i} & \ \ 
\bm{\eta}%
_{i}^{T}\widetilde{%
\bm{\xi}%
}_{i}%
\end{array}%
\right) \mathbf{V}_{i}^{\text{s}} \\
&=&\left( 
\begin{array}{cc}
\mathbf{0} & \ \ 
\bm{\eta}%
_{i}^{T}\widetilde{%
\bm{\xi}%
}_{i}%
\end{array}%
\right) \mathbf{V}_{i}^{\text{s}}=-(\widetilde{%
\bm{\xi}%
}_{i}%
\bm{\eta}%
_{i})^{T}\mathbf{v}_{i}^{\text{s}}
\end{eqnarray*}%
the relation (\ref{InvKinRec2}) for the acceleration, for instance, becomes%
\begin{equation}
\ddot{q}_{i}=\mathbf{S}_{i}^{+}\left( \dot{\mathbf{V}}_{i}^{\text{s}}-\dot{%
\mathbf{V}}_{i-1}^{\text{s}}\right) -\frac{\dot{q}_{i}}{\left\Vert \mathbf{S}%
_{i}\right\Vert ^{2}}\left( 
\bm{\xi}%
_{i}\times 
\bm{\eta}%
_{i}\right) ^{T}\mathbf{v}_{i}^{\text{s}}.  \label{InvKinRec22}
\end{equation}%
The last term in (\ref{InvKinRec22}) allows a geometric interpretation
recalling that $%
\bm{\xi}%
_{i}\times 
\bm{\eta}%
_{i}$ is the coordinate vector to the point on the instantaneous joint screw
axis which is closest to the origin of $\mathcal{F}_{0}$.

The above solutions (\ref{InvKinRec2}-\ref{InvKinRec4}) are exact as long as
the twists of all bodies and their derivatives are consistent with the
kinematics, i.e. that they satisfy the inter-body constraints due to the
joints. If this is not the case, then (\ref{InvKinRec2}-\ref{InvKinRec4})
represent the unique solution with minimum error. This is for instance the
case when processing measurement data of motion capture systems to generate
motions of a human body model.

\begin{remark}
In the above relations (\ref{InvKinRec})-(\ref{InvKinRec4}), the screw
coordinate vector $\mathbf{S}_{i}$ is regarded as a vector in ${\mathbb{R}}%
^{6}$ with Euclidean norm $\left\Vert \mathbf{S}_{i}\right\Vert ^{2}$, and $%
\mathbf{S}_{i}^{+}$ is the pseudoinverse of $\mathbf{S}_{i}$ according to
the metric of ${\mathbb{R}}^{6}$. While these relations follow with basic
linear algebra, it should be remarked that there is no frame invariant inner
product of screws. Geometrically, $\mathbf{S}_{i}^{T}\mathbf{S}_{i}$ must be
interpreted as the pairing of the screw coordinates representing the twist $%
\mathbf{S}_{i}\dot{q}_{i}$ with some screw coordinates representing a wrench
with intensity $f$ that is not reciprocal to the former, i.e. a wrench that
performs work on the twist. The latter is given in ray coordinates by $f%
\mathbf{S}_{i}$.
\end{remark}

\section{Taylor-Series Expansion of the Spatial Jacobian%
\label{secTaylorJacobian}%
}

\subsection{Taylor-series of instantaneous joint screws}

The joint screw coordinates as function of the joint variables $\mathbf{q}$
admit the Taylor series 
\begin{eqnarray}
\mathbf{S}_{i}%
\hspace{-0.5ex}%
\left( \mathbf{q+x}\right) &=&\mathbf{S}_{i}%
\hspace{-0.5ex}%
\left( \mathbf{q}\right) +\sum_{k\geq 1}\frac{1}{k!}\mathrm{d}^{k}\mathbf{S}%
_{i,\mathbf{q}}\left( \mathbf{x}\right)  \label{SiSeries1} \\
&=&\mathbf{S}_{i}%
\hspace{-0.5ex}%
\left( \mathbf{q}\right) +\sum_{j<i}x_{j}\frac{\partial \mathbf{S}_{i}}{%
\partial q_{j}}+\frac{1}{2!}\sum_{k\leq j<i}x_{k}x_{j}\frac{\partial ^{2}%
\mathbf{S}_{i}}{\partial q_{k}\partial q_{j}}+\frac{1}{3!}\sum_{l\leq k\leq
j<i}x_{l}x_{k}x_{j}\frac{\partial ^{3}\mathbf{S}_{i}}{\partial q_{l}\partial
q_{k}\partial q_{j}}+\ldots  \notag \\
&=&\mathbf{S}_{i}%
\hspace{-0.5ex}%
\left( \mathbf{q}\right) +\sum_{j<i}x_{j}\left[ \mathbf{S}_{j},\mathbf{S}%
_{i}]\right] +\frac{1}{2!}\sum_{k\leq j<i}x_{k}x_{j}\left[ \mathbf{S}_{k},%
\left[ \mathbf{S}_{j},\mathbf{S}_{i}\right] \right] +\frac{1}{3!}\sum_{l\leq
k\leq j<i}x_{l}x_{k}x_{j}\left[ \mathbf{S}_{l},[\mathbf{S}_{k},\left[ 
\mathbf{S}_{j},\mathbf{S}_{i}\right] \right] ]+\ldots  \label{SiSeries2}
\end{eqnarray}%
where the differential (\ref{dkS}) is now written explicitly as sum of Lie
brackets (\ref{dnS}). The series (\ref{SiSeries2}) can be evaluated
directly, or the recursive relation (\ref{dkSii}) can be used to evaluate (%
\ref{SiSeries1}).

\subsection{Taylor-series of the Jacobian and the involutive closure of the
image space of the kinematic mapping}

As in section a), consider the terminal link $n$ of a kinematic chain, and,
for simplicity, denote its KM with $f$ and its spatial Jacobian with $%
\mathbf{J}$. The series expansion of the spatial Jacobian of the serial
chain with $n$ joints is 
\begin{equation}
\mathbf{J}\left( \mathbf{q+x}\right) :=%
\Big%
(\mathbf{S}_{1}%
\hspace{-0.5ex}%
\left( \mathbf{q}\right) 
\Big%
|\mathbf{S}_{2}%
\hspace{-0.5ex}%
\left( \mathbf{q+x}\right) 
\Big%
|\mathbf{S}_{3}%
\hspace{-0.5ex}%
\left( \mathbf{q+x}\right) 
\Big%
|\cdots 
\Big%
|\mathbf{S}_{n}%
\hspace{-0.5ex}%
\left( \mathbf{q+x}\right) 
\Big%
)  \label{Jseries}
\end{equation}%
with $\mathbf{S}_{i}=\mathbf{S}_{i}%
\hspace{-0.5ex}%
\left( \mathbf{q}\right) $ and the expansion of the instantaneous joint
screws (\ref{SiSeries2}). In accordance with (\ref{Si}), the screw
coordinate of the first joint is constant. The $i$th column of the spatial
Jacobian depends on the increments of joint variables of the preceding
joints $j=1,\ldots ,i-1$.

The image space of $\mathbf{J}\left( \mathbf{q}\right) $ is the $se\left(
3\right) $-subspace of possible twists of the terminal link $n$ of the open
kinematic chain at $\mathbf{q}$. The space of possible twist at another
configuration is determined by inserting a particular $\mathbf{x}\in {%
\mathbb{R}}^{n}$ in (\ref{Jseries}). The form (\ref{SiSeries2}) allows to
relate these spaces and thus to estimate the image space of the KM.

From (\ref{SiSeries2}) follows that, for any $\mathbf{x}\in {\mathbb{R}}^{n}$%
, 
\begin{equation}
\mathrm{im}~\mathbf{J}\left( \mathbf{q}+\mathbf{x}\right) \subseteq \mathrm{%
span}~\left( \mathbf{S}_{i},\left[ \mathbf{S}_{j},\mathbf{S}_{i}]\right] ,%
\left[ \mathbf{S}_{k},\left[ \mathbf{S}_{j},\mathbf{S}_{i}\right] \right] ,%
\left[ \mathbf{S}_{l},[\mathbf{S}_{k},\left[ \mathbf{S}_{j},\mathbf{S}_{i}%
\right] \right] ,\left[ \mathbf{S}_{m},[\mathbf{S}_{l},[\mathbf{S}_{k},\left[
\mathbf{S}_{j},\mathbf{S}_{i}\right] \right] ],\ldots \right)
\end{equation}%
with $\mathbf{S}_{i}=\mathbf{S}_{i}%
\hspace{-0.5ex}%
\left( \mathbf{q}\right) $. A basis for any subalgebra of the semi-direct
product $se\left( 3\right) =so\left( 3\right) \ltimes {\mathbb{R}}^{3}$
(semidirect product of two 3-dim Lie algebras) is obtained after at most
three-fold application of the Lie bracket (this was discussed for $se\left(
3\right) $ in \cite{Hao1998}). Consequently, the image space of $\mathbf{J}$
at any $\mathbf{q}$ is a vector subspace of the Lie algebra $\overline{D}=%
\mathrm{span}~\left( \mathbf{S}_{i},\left[ \mathbf{S}_{j},\mathbf{S}_{i}]%
\right] ,\left[ \mathbf{S}_{k},\left[ \mathbf{S}_{j},\mathbf{S}_{i}\right] %
\right] ,\left[ \mathbf{S}_{l},[\mathbf{S}_{k},\left[ \mathbf{S}_{j},\mathbf{%
S}_{i}\right] \right] \right) $. The latter is the involutive closure of the
screw system $\{\mathbf{S}_{1},\ldots ,\mathbf{S}_{n}\}$. The expression of $%
f_{i}$ in (\ref{fi}) along with (\ref{Si}) shows that $\mathrm{Ad}_{f_{i}}%
\overline{D}=\overline{D},i=1,\ldots ,n$, which simply means that the motion
of any joint of the chain does not change the space of possible twists that
the terminal link can perform for all possible configurations. Thus, the
smallest $se\left( 3\right) $-subalgebra, i.e. the vector space, to which
the terminal twist belongs for any possible configuration of the chain, is%
\begin{equation}
\overline{D}=\mathrm{span}~\left( \mathbf{Y}_{i},\left[ \mathbf{Y}_{j},%
\mathbf{Y}_{i}]\right] ,\left[ \mathbf{Y}_{k},\left[ \mathbf{Y}_{j},\mathbf{Y%
}_{i}\right] \right] ,\left[ \mathbf{Y}_{l},[\mathbf{Y}_{k},\left[ \mathbf{Y}%
_{j},\mathbf{Y}_{i}\right] \right] ,\ i,j,k,l=1,\ldots ,n\right) .
\label{Dclos}
\end{equation}%
The corresponding Lie group, denoted $G:=\exp \overline{D}$, is the smallest
subgroup containing the image space of the KM.

Thus, an involutive closure of the image space of the KM can be determined
by 3-fold Lie brackets of the joint screws $\mathbf{Y}_{i}$ in the reference
configuration. The $\mathbf{Y}_{i}$ can be determined in any reference
configuration, even singular.

\subsection{Applications}

\paragraph{a) Structural mobility formulae for estimating the finite DOF of
linkages}

The local finite DOF $\delta _{\mathrm{loc}}\left( \mathbf{q}\right) $ of a
single-loop linkage in configuration $\mathbf{q}$ is the local dimension of
the c-space $V$ at $\mathbf{q}$, which is the solution variety (\ref{V}) of
the loop constraints $f\left( \mathbf{q}\right) =\mathbf{I}$. The maximal
possible number $m$ of independent constraints is the dimension of the image
space of $f$, which is the maximal rank of $\mathbf{J}\left( \mathbf{q}%
\right) $ for $\mathbf{q}\in {\mathbb{V}}^{n}$. Since the dimension of the
loop algebra is never lower than that of $\mathrm{im}\,f$, it follows that $%
\delta _{\mathrm{loc}}\left( \mathbf{q}\right) \geq n-\dim \mathrm{im}%
\,f\geq n-\dim \overline{D}$. It thus provides an upper bound on the number
of independent constraints.

The dimension of the loop algebra (respectively of the motion group) is the
parameter appearing in all \emph{structural mobility} formulae \cite%
{Gogu2005}, which estimate the (internal) finite DOF by relating the DOFs of
bodies and the number of constraints imposed by the joints. Best known is
the Chebychev--Kutzbach--Gr\"{u}bler (CKG) formula that computes the
structural mobility as $\delta _{\mathrm{str}}=g\left( n_{\mathrm{B}%
}-1\right) -\sum\nolimits_{i}^{n_{\mathrm{J}}}\left( g-f_{i}\right)
=\sum\nolimits_{i}^{n_{\mathrm{J}}}f_{i}-g\left( n_{\mathrm{J}}-n_{\mathrm{B}%
}+1\right) $, wherein the characteristic parameter $g$ indicates the assumed
DOF of the unconstrained bodies, i.e. before assembling the mechanism.
Common choices are $g=3$ for 'planar' and 'spherical', and $g=6$ for
'spatial' mechanisms. That is, members of the mechanism are a priori
supposed to be restricted to a certain motion subspace. The CKG formula can
be written as $\delta _{\mathrm{str}}=\sum\nolimits_{i}^{n_{\mathrm{J}%
}}f_{i}-g\gamma $, where $\gamma =n_{\mathrm{J}}-n_{\mathrm{B}}+1$ is the
number fundamental cycles (FCs) $\Lambda _{1},\ldots ,\Lambda _{\gamma }$ of
the mechanism's topological graph \cite{Topology,Robotica2017}. Separating
this for the individual FCs yields $\delta _{\mathrm{str}}=\sum%
\nolimits_{i}f_{i}-(g_{1}+g_{2}\ldots +g_{\gamma })$. The CKG formula
computes the correct DOF for mechanisms without kinematic loops. For a
closed loop mechanism ($\gamma >0$) $g_{l}$ is the number of constraints
imposed on the FC $\Lambda _{l}$ in order to close the corresponding
kinematic loop. For a given linkage the number of constraints that are
independent for a general configuration $\mathbf{q}\in V$ depends on the
particular geometry. The fact that the latter cannot be inferred directly
from the geometry for overconstrained linkages is still a topic of research.
However, since each FC corresponds to a kinematic chain, the dimension of
the loop algebra provides an upper estimate of the number of independent
constraints: $g_{l}=\dim \overline{D}_{l}$. Possible values are $%
g_{l}=1,2,3,4,6$ since there is no 5-dimensional $SE\left( 3\right) $%
-subgroup. This admits a systematic treatment without the need to guess
about the motion characteristic $g$, which is usually assumed known a priori
rather than determined from the linkage kinematics.

The motion subgroup associated to a kinematic loop has been the central
element in the classification proposed by Herv\'{e} \cite%
{Herve1978,Herve1982}. According to this classification, a linkage is called
'trivial' if the CKG formula with $g_{l}=\dim \overline{D}_{l}$ yields the
correct finite mobility. It is called 'exceptional' if the mobility can be
explained by the intersection of the motion subgroups associated to two
subchains (obtained by opening the loop), which are also be determined by (%
\ref{Dclos}). Otherwise the linkage is called 'paradoxical'.

Rico \& Ravani \cite{RicoGallargoRavani2003,RicoRavani2003} developed
mobility formulae based on the Lie algebra $\overline{D}$ generated by the
chain. These mobility criteria were refined by using the intersection of Lie
algebras generated by subchains in certain order \cite%
{RicoRavani2006,Rico2007}. These closure algebras are also determined by (%
\ref{Dclos}).

The different motion spaces generated by individual kinematic chains were
used addressed in \cite{Sugimoto2001} for generation of motion equations of
parallel manipulators.

\subparagraph{Example 1 (cont.): 4C-Linkage with a shaky motion mode}

The loop algebra of the 4C linkage in fig. \ref{fig4C} is determined with
the screw coordinates (\ref{Y4C}). The loop algebra is invariant and can
even be deduced from this singular configuration. According to (\ref{Dclos}%
), the nested Lie brackets yield $\overline{D}=se\left( 3\right) $. As shown
in sec. \ref{sec4CKinCone} the linkage has local DOF $\delta =2$. The 4C
linkage thus obeys the CKG formula with $g=\dim \overline{D}=6$ and $%
\sum\nolimits_{i}f_{i}=8$, i.e. $\delta _{\mathrm{str}}=8-6=2$. According to
the terminology proposed by Herv\'{e} \cite{Herve1978,Herve1982}, this
linkage is trivial.

\subparagraph{Example 2: 4-bar with 2R and 2C joints}

\begin{figure}[h]
\centerline{
a)\includegraphics[width=0.47\textwidth]{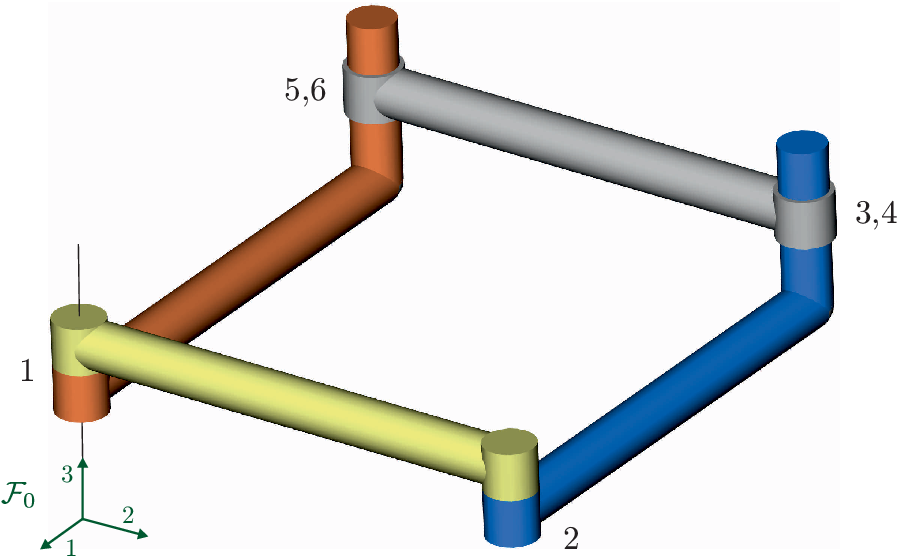}~~~~~~
b)\includegraphics[width=0.45\textwidth]{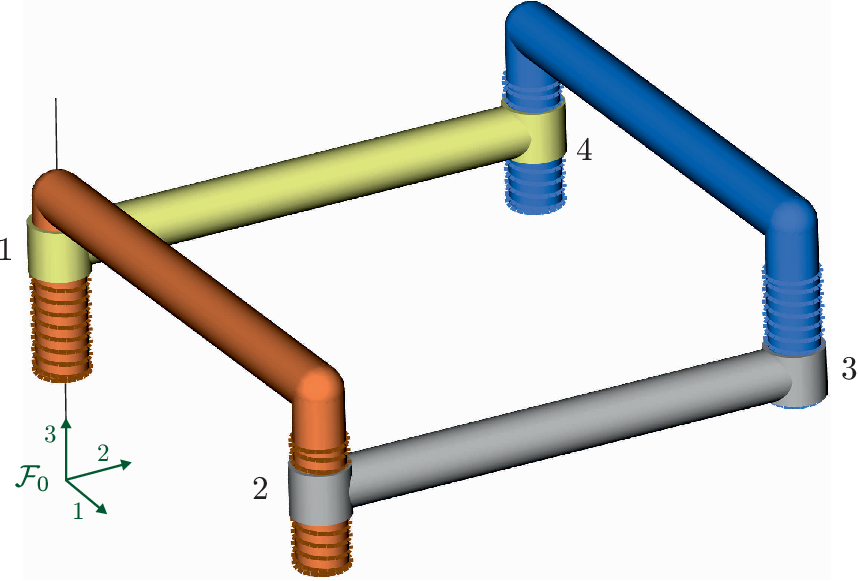}
}
\caption{a) 2-DOF linkage comprising two revolute and two cylindrical
joints. b) 1-DOF linkage comprising 4 helical joints.}
\label{fig2R2C}
\end{figure}

The linkage shown in fig. \ref{fig2R2C}a) comprises two revolute and two
cylindrical joints (modeled as a revolute followed by a prismatic joint).
The links have equal lengths $L$. The joint screw coordinates in the shown
reference configuration are%
\begin{eqnarray}
\mathbf{Y}_{1} &=&\left( 0,0,1,0,0,0\right) ^{T},\mathbf{Y}_{2}=\left(
0,0,1,0,-L,0\right) ^{T},\mathbf{Y}_{3}=\left( 0,0,1,L,-L,0\right) ^{T} 
\notag \\
\mathbf{Y}_{4} &=&\left( 0,0,0,0,0,1\right) ^{T},\mathbf{Y}_{5}=\left(
0,0,1,L,0,0\right) ^{T},\mathbf{Y}_{6}=\left( 0,0,0,0,0,1\right) ^{T}.
\end{eqnarray}%
The closure algebra is readily found as $\overline{D}=se\left( 2\right)
\times {\mathbb{R}}$, the algebra of planar motions and translations along
the plane normal. Thus $g=\dim \overline{D}=4$, and the CKG formula yields $%
\delta _{\mathrm{str}}=6-4=2$. This is a correct estimation, and the linkage
is trivial. The calculation can be found in the provided Mathematica
notebook \cite{MendeleyDataset}.

\subparagraph{Example 3: Delassus 4H linkage}

The linkage comprising four helical joints with respective pitch $%
h_{1},h_{2},h_{3},h_{4}$ is shown in fig. \ref{fig2R2C}b). The condition for
mobility is that $h_{1}+h_{3}=h_{2}+h_{4}$. Links 1 and 3 have length $a$,
and links 2 and 4 have length $b$. In the shown reference configuration the
joint screw coordinates are%
\begin{equation}
\mathbf{Y}_{1}=\left( 0,0,1,0,0,h_{1}\right) ^{T},\mathbf{Y}_{2}=\left(
0,0,1,0,-a,h_{2}\right) ^{T},\mathbf{Y}_{3}=\left( 0,0,1,b,-a,h_{3}\right)
^{T},\mathbf{Y}_{4}=\left( 0,0,1,b,0,h_{4}\right) ^{T}.  \notag
\end{equation}%
The closure algebra is found as $\overline{D}=so\left( 2\right) \times {%
\mathbb{R}}^{3}$, which is the algebra of Sch\"{o}nflies motions (Scara
motions) with $g=\dim \overline{D}=4$. The CKG formula would incorrectly
estimate the DOF $\delta _{\mathrm{str}}=4-4=0$. The linkage is paradoxical.

For the special case that $h=h_{1}=h_{2}=h_{3}=h_{4}$ the closure algebra
becomes%
\begin{equation*}
\overline{D}=\mathrm{span}~\left( \left( 0,0,0,0,1,0\right) ^{T},\left(
0,0,0,1,0,0\right) ^{T},\left( 0,0,1,0,0,h\right) ^{T}\right) =H_{h}\ltimes {%
\mathbb{R}}^{2}
\end{equation*}%
which is the algebra of planar motions and screw motions with pitch $h$
perpendicular to the plane of motion. Then $g=\dim \overline{D}=3$ and the
CKG formula gives the correct result $\delta _{\mathrm{str}}=4-3=1$, i.e.
the linkage with this geometry is trivial. Again, the calculation can be
found in the provided Mathematica notebook \cite{MendeleyDataset}. 

\paragraph{b) Elimination of redundant loop constraints}

The existence of redundant loop constraints is a critical issue in
computational kinematics and multibody dynamics \cite%
{JalonMUBO2013,Wojtyra2005} since they lead to velocity constraints with
singular coefficient matrix. Simulation tools usually do not distinguish
between different motion spaces and always assign the maximal number of
constraints. The standard approach is to employ numerically robust
algorithms for solving redundant linear systems (e.g. SVD), which leads to a
significant increase computational effort.

However, instead of such a purely numerical approach, the problem can be
addressed making use of the loop algebra.

Consider a single kinematic loop. The velocity constraints are given by (\ref%
{VelConstr}). The maximal rank of the Jacobian is generally less than 6
depending on the motion space of the kinematic chain, for which the loop
algebra provides an upper bound. The idea is to reduce the constraints to
this subalgebra. A basis for the loop algebra is given by (\ref{Dclos}).
Denote with $\mathbf{B}$ the matrix with $\mathbf{Y}_{i}$ and the Lie
brackets in (\ref{Dclos}) as its columns. It has rank $g=\dim \overline{D}%
\leq 6$ and admits a singular value decomposition $\mathbf{B}=\mathbf{U}^{T}%
\bm{\Sigma}\mathbf{V}$, with $\mathbf{U}^{T}\mathbf{U}=\mathbf{I}_{6},%
\mathbf{V}^{T}\mathbf{V}=\mathbf{I}_{g}$ and $\bm{\Sigma}=\mathrm{diag}%
\,\left( \sigma _{1},\ldots ,\sigma _{g},0,\ldots ,0\right) $, where $\sigma
_{1},\ldots ,\sigma _{g}$ are the non-zero singular values. The constraint
Jacobian has $\mathrm{rank}~\mathbf{J}\leq g$. It can be reduced to the $%
g\times n$ matrix $\overline{\mathbf{J}}:=\overline{\mathbf{U}}\mathbf{J}$
where $\overline{\mathbf{U}}$ is the $g\times 6$ matrix consisting of the
first $g$ rows of $\mathbf{U}$. The reduced constraints are then $\overline{%
\mathbf{J}}\dot{\mathbf{q}}=\mathbf{0}$. This method never removes
independent constraints. It always removes the correct number of redundant
constraints for so-called trivial and exceptional linkages. It is
numerically well-posed since the basis matrix (consisting of screw
coordinates) is well-conditioned. Its application to multi-loop linkages was
presented in \cite{RedConstraints2014}.

\section{Higher-Order Solution of Loop Closure Constraints%
\label{secLoopSolution}%
}

Due to the closure constraints, the joint coordinates of a kinematic loop
are dependent. A set of independent coordinates can be selected to
parameterize the configuration of the linkage. The motion of the linkage can
be determined from a prescribed motion of the independent joint coordinates
by solution of the geometric loop constraints. The latter cannot be solved
analytically, but a higher-order approximate solution of the closure
constraints in form of a power series can be used. This problem has been
addressed in \cite{Milenkovic2012} where a series expansion was derived in
terms of screw coordinates. It was also mentioned as an application of the
formulation presented in \cite{Karsai2002}. In the recent publication \cite%
{deJong2018} a systematic approach to the higher-order solution was
presented.

\subsection{Time derivatives of the solution of loop closure constraints}

Possible joint velocities $\dot{\mathbf{q}}$ of a closed loop must satisfy
the velocity loop constraints (\ref{VelConstr}). Admissible accelerations
are the solutions of the acceleration constraints, i.e. of the time
derivative of (\ref{VelConstr}). Finally, the $k$th time derivative of $%
\mathbf{q}$ must satisfy the constraints (\ref{HighOrderConstr}).

\begin{assumption}
\label{AssumptionRank}%
The Jacobian in the velocity loop constraints (\ref{VelConstr}) is a full
rank $m\times n$ matrix, so that the differential and local DOF of the
kinematic loop is $\delta _{\mathrm{diff}}\left( \mathbf{q}\right) =\delta _{%
\mathrm{loc}}\left( \mathbf{q}\right) =n-m$ (i.e. the linkage is not
shaky/underconstrained). It is assumed that its rank is locally constant
near the considered point $\mathbf{q}\in {\mathbb{R}}^{n}$.
\end{assumption}

The solution of the velocity constraints (\ref{VelConstr}) can be expressed
in terms of $\delta $ independent joint velocities. To this end, the joint
coordinate vector $\mathbf{q}$ is split into the vector $\mathbf{u}\in {%
\mathbb{V}}^{\delta }$ of $\delta $ selected independent coordinates and the
vector $\mathbf{d}\in {\mathbb{V}}^{m}$ comprises the $m$ dependent
coordinates, and are collected in the (rearranged) vector of joint
coordinates $\overline{\mathbf{q}}:=\left( \mathbf{d},\mathbf{u}\right) $.
The constraints are accordingly written as%
\begin{equation}
\mathbf{0}=\mathbf{J}_{\mathrm{d}}\dot{\mathbf{d}}+\mathbf{J}_{\mathrm{u}}%
\dot{\mathbf{u}}
\end{equation}%
with corresponding partitioning of the Jacobian%
\begin{eqnarray}
\mathbf{J}_{\mathrm{d}} &=&\left( \mathbf{S}_{\alpha _{1}},\ldots \mathbf{S}%
_{\alpha _{m}}\right) ,\alpha _{1}<\ldots <\alpha _{m}\in I_{\mathrm{d}} \\
\mathbf{J}_{\mathrm{u}} &=&\left( \mathbf{S}_{\alpha _{1}},\ldots \mathbf{S}%
_{\alpha _{\delta }}\right) ,\alpha _{1}<\ldots <\alpha _{\delta }\in I_{%
\mathrm{u}}  \notag
\end{eqnarray}%
where $I_{\mathrm{d}}$ and $I_{\mathrm{u}}$ is the index set of the $m$
dependent and the $\delta $ independent joint coordinates, respectively. A
solution of (\ref{VelConstr}) is given in terms of the orthogonal complement 
$\mathbf{F}$ of $\mathbf{J}$ as%
\begin{equation}
\dot{\overline{\mathbf{q}}}=\mathbf{F}\dot{\mathbf{u}},\text{ with }\mathbf{F%
}:=\left( 
\begin{array}{c}
\mathbf{D} \\ 
\mathbf{I}_{\delta }%
\end{array}%
\right) ,\ \mathbf{D:}=-\mathbf{J}_{\mathrm{d}}^{-1}\mathbf{J}_{\mathrm{u}}.
\label{qs}
\end{equation}%
Deriving solutions for higher time derivatives of $\mathbf{q}$ in terms of
those of $\mathbf{u}$ boils down to time derivatives of $\dot{\mathbf{d}}=%
\mathbf{D}\dot{\mathbf{u}}$:%
\begin{equation}
\mathbf{d}^{\left( k\right) }=\sum_{i=0}^{k-1}\tbinom{k-1}{l}\mathrm{D}%
^{\left( l\right) }\mathbf{D}\mathrm{D}^{\left( k-l\right) }\mathbf{u}.
\label{rs}
\end{equation}%
According to the definition (\ref{qs}), it is 
\begin{equation}
\mathrm{D}^{\left( k\right) }\mathbf{D}=-\sum_{l=0}^{k}\tbinom{k}{l}\mathrm{D%
}^{\left( l\right) }\mathbf{J}_{\mathrm{d}}^{-1}\mathrm{D}^{\left(
k-l\right) }\mathbf{J}_{\mathrm{u}}.  \label{DR}
\end{equation}%
Rearranging the derivatives of $\mathbf{J}_{\mathrm{d}}\mathbf{J}_{\mathrm{d}%
}^{-1}=\mathbf{I}$, the derivatives of $\mathbf{J}_{\mathrm{d}}^{-1}$ are
found as 
\begin{equation}
\mathrm{D}^{\left( k\right) }\mathbf{J}_{\mathrm{d}}^{-1}=-\mathbf{J}_{%
\mathrm{d}}^{-1}\sum_{l=1}^{k}\tbinom{k}{l}\mathrm{D}^{\left( l\right) }%
\mathbf{J}_{\mathrm{d}}\mathrm{D}^{\left( k-l\right) }\mathbf{J}_{\mathrm{d}%
}^{-1}.  \label{DJinv}
\end{equation}%
Time derivatives of $\mathbf{J}_{\mathrm{d}}$ and $\mathbf{J}_{\mathrm{u}}$
are known with those of the $\mathbf{S}_{i},i=1,\ldots ,n$. However, the
screw coordinates must be considered as functions of the $\delta $
independent coordinates $\mathbf{u}$. This dependence is indicated by the
notation $\overline{\mathbf{S}}_{i}\left( \mathbf{u}\right) :=\mathbf{S}%
_{i}\left( \mathbf{q}\left( \mathbf{u}\right) \right) $, $\overline{\mathsf{S%
}}_{i}\left( \mathbf{u}\right) :=\mathsf{S}_{i}\left( \mathbf{q}\left( 
\mathbf{u}\right) \right) $, and $\overline{\mathbf{F}}\left( \mathbf{u}%
\right) :=\mathbf{F}\left( \mathbf{q}\left( \mathbf{u}\right) \right) $.

Due to (\ref{qs}), partial derivatives of $\overline{\mathbf{S}}_{i}$ w.r.t.
the independent coordinates are%
\begin{equation}
\frac{\partial }{\partial u_{l}}\overline{\mathbf{S}}_{i}=\sum_{j\leq n}%
\frac{\partial }{\partial q_{j}}\mathbf{S}_{i}F_{jl}=\sum_{j\leq n}[%
\overline{\mathbf{S}}_{j},\overline{\mathbf{S}}_{i}]F_{jl}.
\end{equation}%
Thus the time derivatives of $\overline{\mathbf{S}}_{i}$ are 
\begin{equation}
\dot{\overline{\mathbf{S}}}_{i}=\sum_{l\in I_{\mathrm{u}}}\sum_{j\leq i}[%
\overline{\mathbf{S}}_{j},\overline{\mathbf{S}}_{i}]F_{jl}\dot{u}_{l}=[%
\overline{\mathsf{S}}_{i},\overline{\mathbf{S}}_{i}]
\end{equation}%
where now, in analogy to (\ref{SSi}), 
\begin{equation}
\overline{\mathsf{S}}_{i}:=\sum_{l\in I_{\mathrm{u}}}\sum_{j\leq i}\overline{%
\mathbf{S}}_{j}F_{jl}\dot{u}_{l}.
\end{equation}%
Higher-order time derivatives then follow with Leibnitz' rule, similarly to (%
\ref{DSi}) and (\ref{DSik}), as%
\begin{eqnarray}
\mathrm{D}^{(k)}\overline{\mathbf{S}}_{i} &=&\sum_{l=0}^{k-1}\tbinom{k-1}{l}[%
\mathrm{D}^{(l)}\overline{\mathsf{S}}_{i},\mathrm{D}^{(k-l-1)}\overline{%
\mathbf{S}}_{i}]  \label{DSibar} \\
\mathrm{D}^{(k)}\overline{\mathsf{S}}_{i} &=&\sum_{j\leq i}\sum_{l=0}^{k}%
\tbinom{k}{l}\mathrm{D}^{(l)}\overline{\mathbf{S}}_{j}q_{j}^{(k-l+1)}.
\label{DSibar2}
\end{eqnarray}%
Noting (\ref{qs}), the time derivatives of $\mathbf{q}$ in (\ref{DSibar2})
are determined by those of $\mathbf{u}$ as 
\begin{equation}
\overline{\mathbf{q}}^{(k)}=\sum_{l=0}^{k-1}\tbinom{k-1}{l}\mathrm{D}^{(l)}%
\mathbf{Fu}^{\left( k-l\right) }.  \label{qku}
\end{equation}%
Finally, with (\ref{qs}), the derivatives of the orthogonal complement are
available with (\ref{DR}) as 
\begin{equation}
\mathrm{D}^{(k)}\mathbf{F}=\left( 
\begin{array}{c}
\mathrm{D}^{(k)}\mathbf{D} \\ 
\mathbf{0}_{\delta }%
\end{array}%
\right) ,k\geq 1  \label{DF}
\end{equation}%
In summary, the relations (\ref{rs},\ref{DR},\ref{DJinv}) allow to determine
solutions of the higher-order loop constraints, where the time derivatives
of the joint screw coordinates are given by (\ref{DSibar},\ref{DSibar2}).
These relations are recursive.

\begin{remark}
The curve parameter $t$ does not have to be time. It can be any parameter
describing the curve $\mathbf{q}\left( t\right) $.
\end{remark}

\begin{remark}
The treatment presented in the following is still applicable even if the
assumption \ref{AssumptionRank}, that the constraint Jacobian has full rank,
is not satisfied. It must only be assumed that the rank is locally constant
in $V$ (i.e. the linkage is not in a singularity). Then $\mathbf{J}_{\mathrm{%
d}}$ is not full rank and its pseudoinverse must be used.
\end{remark}

\subsection{Approximate Solution of Closure Constraints}

The higher-order solutions can be used to compute $k$th-order approximate
solutions of the motion of a closed-loop linkage. Given the independent
joint coordinates $\mathbf{u}\left( t\right) $ and their time derivatives up
to order $k$ as function of a parameter $t$ (e.g. time), then a $k$th-order
approximate solution for the motion of the linkage is%
\begin{equation}
\mathbf{q}\left( t+\Delta t\right) =\mathbf{q}\left( t\right) +\Delta t\dot{%
\mathbf{q}}\left( t\right) +\frac{\Delta t^{2}}{2}\ddot{\mathbf{q}}\left(
t\right) +\ldots +\frac{\Delta t^{k}}{k!}\mathbf{q}^{(k)}\left( t\right) .
\label{Deltaq}
\end{equation}%
In (\ref{Deltaq}) the solution (\ref{qku}) in terms of the independent
coordinates is used.

\paragraph{Example: Planar 4-Bar linkage}

As a simple example consider the planar 4-bar linkage in the configuration
shown in

\begin{figure}[h]
\centerline{\includegraphics[width=0.47%
\textwidth]{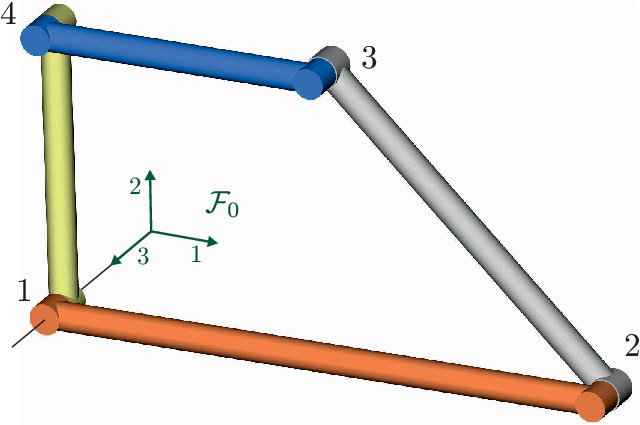}}
\caption{Planar 4-bar linkage in reference configuration $\mathbf{q}=\mathbf{%
0}$.}
\label{fig4bar_MotionAppr}
\end{figure}

Fig. \ref{fig4bar_MotionAppr}. The screw coordinates are then 
\begin{equation*}
\mathbf{Y}_{1}=\left( 0,0,1,0,0,0\right) ^{T},\mathbf{Y}_{2}=\left(
0,0,1,0,-2,0\right) ^{T},\mathbf{Y}_{3}=\left( 0,0,1,1,-1,0\right) ^{T},%
\mathbf{Y}_{4}=\left( 0,0,1,1,0,0\right) ^{T}.
\end{equation*}

\paragraph{a)}

First $q_{4}$ is used as the independent joint angle, i.e. it serves as
input for a kinematic motion analysis. The index set of dependent and
independent coordinates are $I_{\mathrm{d}}=\{1,2,3\}$ and $I_{\mathrm{u}%
}=\{4\}$, respectively. The recursive relations yield the time derivatives
of $\mathbf{q}\left( t\right) =\left( q_{1}\left( t\right) ,q_{2}\left(
t\right) ,q_{3}\left( t\right) ,q_{4}\left( t\right) \right) $ up to $4$th
order, for instance, in terms of those of $q_{4}\left( t\right) $ as%
\begin{equation*}
\dot{\mathbf{q}}=\left( 
\begin{array}{c}
-\frac{1}{2}\dot{q}_{4} \\ 
\frac{1}{2}\dot{q}_{4} \\ 
-\dot{q}_{4} \\ 
\dot{q}_{4}%
\end{array}%
\right) ,\ddot{\mathbf{q}}=\left( 
\begin{array}{c}
\frac{1}{4}\left( \dot{q}_{4}^{2}-2\ddot{q}_{4}\right) \\ 
\frac{1}{4}\left( 2\ddot{q}_{4}+\dot{q}_{4}^{2}\right) \\ 
-\ddot{q}_{4}-\frac{1}{2}\dot{q}_{4}^{2} \\ 
\ddot{q}_{4}%
\end{array}%
\right) ,\dddot{\mathbf{q}}=\left( 
\begin{array}{c}
\frac{1}{4}\left( -2\dddot{q}_{4}+3\dot{q}_{4}^{3}+3\dot{q}_{4}\ddot{q}%
_{4}\right) \\ 
\frac{1}{4}\left( 2\dddot{q}_{4}+3\dot{q}_{4}\ddot{q}_{4}\right) \\ 
\frac{1}{4}\left( -4\dddot{q}_{4}-3\dot{q}_{4}^{3}-6\dot{q}_{4}\ddot{q}%
_{4}\right) \\ 
\dddot{q}_{4}%
\end{array}%
\right) ,\ddot{\ddot{\mathbf{q}}}=\left( 
\begin{array}{c}
\frac{1}{4}\left( -2\ddot{\ddot{q}}_{4}+3\ddot{q}_{4}^{2}+8\dot{q}_{4}^{4}+4%
\dddot{q}_{4}\dot{q}_{4}+18\dot{q}_{4}^{2}\ddot{q}_{4}\right) \\ 
\frac{1}{4}\left( 2\ddot{\ddot{q}}_{4}+3\ddot{q}_{4}^{2}+2\dot{q}_{4}^{4}+4%
\dddot{q}_{4}\dot{q}_{4}\right) \\ 
\frac{1}{2}\left( -2\ddot{\ddot{q}}_{4}-3\ddot{q}_{4}^{2}-5\dot{q}_{4}^{4}-4%
\dddot{q}_{4}\dot{q}_{4}-9\dot{q}_{4}^{2}\ddot{q}_{4}\right) \\ 
\ddot{\ddot{q}}_{4}%
\end{array}%
\right) .
\end{equation*}%
These expressions can be used to construct a $4$th-order approximation of
the solution of the closure constraints, denoted $\mathbf{q}^{\left[ 4\right]
}\left( t\right) $. For the particular case of $q_{4}\left( t\right) =\sin t$%
, this is%
\begin{equation*}
\mathbf{q}^{\left[ 4\right] }\left( t\right) =\left( 
\begin{array}{c}
\frac{1}{24}\left( t^{4}+5t^{3}+3t^{2}-12t\right) ,-\frac{1}{48}\left(
t^{4}+4t^{3}-6t^{2}-24t\right) ,-\frac{1}{48}\left(
t^{4}-2t^{3}+12t^{2}+48t\right) ,\sin t%
\end{array}%
\right) ^{T}.
\end{equation*}

\paragraph{b)}

\begin{figure}[h]
a)~\includegraphics[width=0.47\textwidth]{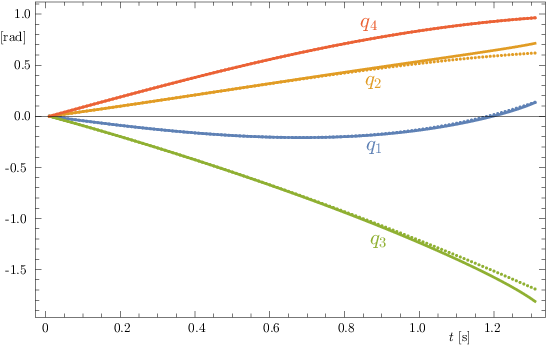}%
~~~~~~~~ b)~\includegraphics[width=0.47%
\textwidth]{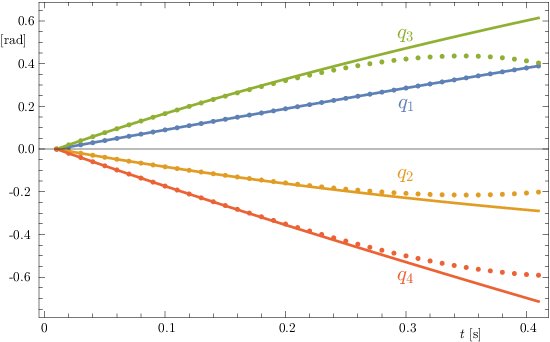}
\caption{Exact (solid line) and approximate solution (dashed line) of $%
\mathbf{q}(t)$ for the planar 4-bar linkage when a) $q_{4}(t)=\sin t$ and b) 
$q_{1}(t)=\sin t$ is used input.}
\label{fig4bar_q1q4}
\end{figure}

Now $q_{1}$ is used as independent coordinate. The index sets are $I_{%
\mathrm{d}}=\{2,3,4\}$ and $I_{\mathrm{u}}=\{1\}$, and the time derivatives
are determined as%
\begin{equation*}
\dot{\mathbf{q}}=\left( 
\begin{array}{c}
\dot{q}_{1} \\ 
-\dot{q}_{1} \\ 
2\dot{q}_{1} \\ 
-2q_{1}%
\end{array}%
\right) ,\ddot{\mathbf{q}}=\left( 
\begin{array}{c}
\ddot{q}_{1} \\ 
2\dot{q}_{1}^{2}-\ddot{q}_{1} \\ 
2\ddot{q}_{1}-4\dot{q}_{1}^{2} \\ 
2\left( \dot{q}_{1}^{2}-\ddot{q}_{1}\right)%
\end{array}%
\right) ,\dddot{\mathbf{q}}=\left( 
\begin{array}{c}
\dddot{q}_{1} \\ 
-\dddot{q}_{1}-12\dot{q}_{1}^{3}+6\dot{q}_{1}\ddot{q}_{1} \\ 
2\left( \dddot{q}_{1}+15\dot{q}_{1}^{3}-6\dot{q}_{1}\ddot{q}_{1}\right) \\ 
-2\left( \dddot{q}_{1}+9\dot{q}_{1}^{3}-3\dot{q}_{1}\ddot{q}_{1}\right)%
\end{array}%
\right) ,\ddot{\ddot{\mathbf{q}}}=\left( 
\begin{array}{c}
\ddot{\ddot{q}}_{1} \\ 
-\ddot{\ddot{q}}_{1}+6\ddot{q}_{1}^{2}+154\dot{q}_{1}^{4}+8\dddot{q}_{1}\dot{%
q}_{1}-72\dot{q}_{1}^{2}\ddot{q}_{1} \\ 
-2\left( -\ddot{\ddot{q}}_{1}+6\ddot{q}_{1}^{2}+184\dot{q}_{1}^{4}+8\dddot{q}%
_{1}\dot{q}_{1}-90\dot{q}_{1}^{2}\ddot{q}_{1}\right) \\ 
-2\ddot{\ddot{q}}_{1}+6\ddot{q}_{1}^{2}+214\dot{q}_{1}^{4}+8\dddot{q}_{1}%
\dot{q}_{1}-108\dot{q}_{1}^{2}\ddot{q}_{1}%
\end{array}%
\right) .
\end{equation*}%
The 4th-order approximation in terms of the independent coordinate $q_{1}$
with $q_{1}\left( t\right) =\sin t$ is%
\begin{equation*}
\mathbf{q}^{\left[ 4\right] }\left( t\right) =\left( 
\begin{array}{c}
\sin t,\frac{73}{12}t^{4}-\frac{11}{6}t^{3}+t^{2}-t,-\frac{2}{3}\left(
22t^{4}-7t^{3}+3t^{2}-3t\right) ,\frac{103}{12}t^{4}-\frac{8}{3}%
t^{3}+t^{2}-2t%
\end{array}%
\right) ^{T}.
\end{equation*}%
For illustration purpose, the approximations are compared with the exact
solutions in Fig. \ref{fig4bar_q1q4}. 
\newpage%

\section{Other Representations of Twists%
\label{secRepresentations}%
}

The advantage of the spatial representation is that the twists of all bodies
are measured and resolved in the same frame, namely the IFR $\mathcal{F}_{0}$%
. For this reason it is used in theoretical kinematics but recently also in
MBS dynamics. For motion control of robotic manipulators and in the
classical formulations of the motion equations for MBS, the body-fixed and
the hybrid representation of twists is used. The kinematics and dynamics
formulations of MBS are traditionally expressed in terms of body-fixed
twists \cite%
{AndersonCritchley2003,Angeles2007,Bae2001,Fijany1995,Hollerbach1980,ParkBobrowPloen1995}%
. Using the spatial representation leads to computationally more efficient
algorithms, however, which is already apparent from the recursive relations (%
\ref{Vsrec}-\ref{Vs3dotrec}) as it does not involve a frame transformation
of twists. Recent formulations therefore employ the spatial representations
of twists (and dually of wrenches) that stem (at least conceptually) from
the spatial operator algebra introduced in \cite%
{Rodriguez1987,Rodriguez1991,Rodriguez1992} and advanced in \cite{Jain1991}.
The Articulated-Body and the Composite-Rigid-Body algorithm \cite%
{Featherstone1983,Featherstone2008} are the most prominent.

Apparently, in various situations it is necessary to relate the different
representations. The relevant relations are presented in the following.

\subsection{Body-Fixed Representation%
\label{secBodyFixedRep}%
}

\subsubsection{Instantaneous joint screws and the body-fixed Jacobian}

The twist of body $i$ in \emph{body-fixed} representation is expressed by
the coordinate vector%
\begin{equation}
\mathbf{V}_{i}^{\text{b}}=\left( 
\begin{array}{c}
\bm{\omega}%
_{i}^{\text{b}} \\ 
\mathbf{v}_{i}^{\text{b}}%
\end{array}%
\right)  \label{DefVb}
\end{equation}%
where $%
\bm{\omega}%
^{\text{b}}$ is the angular velocity of body-fixed frame $\mathcal{F}_{i}$
relative to the world frame $\mathcal{F}_{0}$ resolved in $\mathcal{F}_{i}$,
and $\mathbf{v}^{\text{b}}=\mathbf{R}^{T}\dot{\mathbf{r}}$ is the
translational velocity of the origin of $\mathcal{F}_{i}$ relative to $%
\mathcal{F}_{0}$ resolved in $\mathcal{F}_{i}$.

Denote with ${^{i}}\mathbf{b}_{i,j}\left( \mathbf{q}\right) $ the
instantaneous position vector of a point on the axis of joint $j$ measured
in the body-fixed frame $\mathcal{F}_{i}$, and with ${^{i}}\mathbf{e}%
_{j}\left( \mathbf{q}\right) $ a unit vector along the axis resolved in $%
\mathcal{F}_{i}$. The body-fixed twist coordinate vector is readily
constructed geometrically as \cite{Angeles2007,MUBOScrews1,Murray}%
\begin{eqnarray}
\mathbf{V}_{i}^{\text{b}} &=&\dot{q}_{1}\left( 
\begin{array}{c}
{^{i}}\mathbf{e}_{1} \\ 
{^{i}}\mathbf{b}_{i,1}\times {^{i}}\mathbf{e}_{1}+{^{i}}\mathbf{e}_{1}h_{1}%
\end{array}%
\right) +\dot{q}_{2}\left( 
\begin{array}{c}
{^{i}}\mathbf{e}_{2} \\ 
{^{i}}\mathbf{b}_{i,2}\times {^{i}}\mathbf{e}_{2}+{^{i}}\mathbf{e}_{2}h_{2}%
\end{array}%
\right) +\ldots +\dot{q}_{i}\left( 
\begin{array}{c}
{^{i}}\mathbf{e}_{i} \\ 
{^{i}}\mathbf{b}_{i,i}\times {^{i}}\mathbf{e}_{i}+{^{i}}\mathbf{e}_{i}h_{i}%
\end{array}%
\right)  \label{Vb} \\
&=&\dot{q}_{1}\mathbf{B}_{i,1}+\dot{q}_{2}\mathbf{B}_{i,2}+\ldots +\dot{q}%
_{i}\mathbf{B}_{i,i}  \notag \\
&=&\mathbf{J}_{i}^{\mathrm{b}}\left( \mathbf{q}\right) \dot{\mathbf{q}}
\label{Vbi}
\end{eqnarray}%
where 
\begin{equation}
\mathbf{B}_{i,j}=\left( 
\begin{array}{c}
{^{i}}\mathbf{e}_{j} \\ 
{^{i}}\mathbf{b}_{i,j}\times {^{i}}\mathbf{e}_{j}+{^{i}}\mathbf{e}_{j}h_{j}%
\end{array}%
\right)
\end{equation}%
is the instantaneous screw coordinate vector of joint $j$ represented in
frame $\mathcal{F}_{i}$ fixed at body $i$. The \emph{body-fixed Jacobian} of
body $i$ is thus%
\begin{equation}
\mathbf{J}_{i}^{\mathrm{b}}\left( \mathbf{q}\right) :=%
\Big%
(\mathbf{B}_{i,1}%
\hspace{-0.5ex}%
\left( \mathbf{q}\right) 
\Big%
|\cdots 
\Big%
|\mathbf{B}_{i,i}%
\hspace{-0.5ex}%
\left( \mathbf{q}\right) 
\Big%
|\mathbf{0}%
\Big%
|\cdots 
\Big%
|\mathbf{0}%
\Big%
).  \label{Jib}
\end{equation}%
The body-fixed joint screws are determined analytically as%
\begin{equation}
\mathbf{B}_{i,j}%
\hspace{-0.5ex}%
\left( \mathbf{q}\right) =\mathbf{Ad}_{\mathbf{C}_{i,j}\mathbf{A}_{j}^{-1}}%
\mathbf{Y}_{j},\ j\leq i  \label{Bij}
\end{equation}%
where $\mathbf{C}_{i,j}:=\mathbf{C}_{i}^{-1}\mathbf{C}_{j}=\mathbf{A}%
_{i}^{-1}\exp \left( -\mathbf{Y}_{i}q_{i}\right) \cdot \ldots \cdot \exp
\left( -\mathbf{Y}_{j+1}q_{j+1}\right) \mathbf{A}_{j}$ is the configuration
of body $j$ relative to body $i$. Consequently, the screw of joint $j$
represented in frame $\mathcal{F}_{i}$ on body $i$ depends on the joint
variables $q_{j+1},\ldots ,q_{i}$.

For $i=j$, relation (\ref{Bij}) shows that the screw coordinate vector of
joint $i$ represented in body-fixed frame $\mathcal{F}_{i}$ is constant, and
is denoted with\footnote{%
To be precise, it should be indicated by ${^{i}}\mathbf{X}_{i}$ that the
screw coordinates of joint $i$ are represented in frame $\mathcal{F}_{i}$,
as in \cite{MUBOScrews1}. This is omitted for simplicity.} $\mathbf{X}_{i}:=%
\mathbf{Ad}_{\mathbf{A}_{i}^{-1}}\mathbf{Y}_{i}$. The latter are thus
related to the joint screw coordinates (\ref{Yi}) in spatial representation,
determined in the reference configuration, by the frame transformation $%
\mathbf{Y}_{i}=\mathbf{Ad}_{\mathbf{A}_{i}}\mathbf{X}_{i}$.

The main difference to the spatial representation is that the body-fixed
screw coordinates (\ref{Bij}) of joint $j$ depend on the body $i$. Whereas
in spatial representation the non-zero joint screws (\ref{Si}) in the
Jacobian (\ref{Ji}) are identical, they must be determined for the specific
body $i$ to construct the Jacobian (\ref{Jib}). Yet the body-fixed twist
admits the recursive relation, similarly to (\ref{Vsirec}),

\begin{equation}
\mathbf{V}_{i}^{\text{b}}=\mathbf{Ad}_{\mathbf{C}_{i,i-1}}\mathbf{V}_{i-1}^{%
\text{b}}+\mathbf{X}_{i}\dot{q}_{i}  \label{VbRec}
\end{equation}%
and accordingly 
\begin{equation}
\mathbf{B}_{i,j}=\left\{ 
\begin{array}{l}
\mathbf{Ad}_{\mathbf{C}_{i,i-1}}\mathbf{B}_{i-1,j},j<i \\ 
\mathbf{X}_{i},j=i.%
\end{array}%
\right.  \label{Bijrec}
\end{equation}

\begin{remark}
The body-fixed is traditionally used for dynamics modeling and MBS dynamics
in particular \cite{Angeles2007,WittenburgBook} basically because the
inertia tensor of body $i$ is constant when represented in $\mathcal{F}_{i}$.
\end{remark}

\subsubsection{Partial derivatives of joint screw coordinates in body-fixed
representation}

The instantaneous screw coordinates (\ref{Bij}) of joint $j$ expressed in
the frame affixed at body $i$ depends on the joint variables $q_{j+1},\ldots
,q_{i}$. The nonvanishing partial derivatives are \cite{MMTHighDer}%
\begin{eqnarray}
\dfrac{\partial \mathbf{B}{_{i,j}}}{\partial q_{k}} &=&[\mathbf{B}{_{i,j},%
\mathbf{B}_{i,k}}]  \notag \\
&=&\mathbf{ad}_{\mathbf{B}{_{i,j}}}{\mathbf{B}_{i,k}},~j<k\leq i
\label{derBi}
\end{eqnarray}%
and the nonvanishing repeated $\nu $th-order partial derivatives are 
\begin{eqnarray}
\frac{\partial ^{\nu }\mathbf{B}{_{i,j}}}{\partial q_{\alpha _{1}}\partial
q_{\alpha _{2}}\cdots \partial q_{\alpha _{\nu }}} &=&[\ldots \lbrack
\lbrack \lbrack \mathbf{B}{_{i,j},}\mathbf{B}{_{i,\beta _{1}}}],\mathbf{B}{%
_{i,\beta _{2}}]},\mathbf{B}{_{i,\beta _{3}}]}\ldots ,\mathbf{B}{_{i,\beta
_{\nu }}}],  \notag \\
&=&\left( -1\right) ^{\nu }\mathbf{ad}_{\mathbf{B}{_{i,\beta _{\nu }}}}%
\mathbf{ad}_{\mathbf{B}{_{i,\beta _{\nu }-1}}}\mathbf{ad}_{\mathbf{B}{%
_{i,\beta _{\nu }-2}}}\cdots \mathbf{ad}_{\mathbf{B}{_{i,\beta _{1}}}}%
\mathbf{B}{_{i,j}},\ \ \ \text{if }j<\alpha _{1},\ldots ,a_{\nu }\leq i
\label{dnB}
\end{eqnarray}%
where $j<\beta _{1}\leq \beta _{2}\leq \ldots \leq \beta _{\nu -1}\leq \beta
_{\nu }\leq i$ is the ordered set of indexes $\{\alpha _{1},\ldots ,\alpha
_{\nu }\}$. The sign $\left( -1\right) ^{\nu }$ is caused by swapping the
arguments in the Lie brackets, which is used in order to rearrange the
screws in decreasing order so to avoid long expressing with nested
subscripts of the \textbf{ad} matrices.

Like (\ref{dnS2}) the partial derivatives can be expressed in terms of the
multi-index as%
\begin{equation}
\partial ^{\mathbf{a}}\mathbf{B}{_{i,j}}=\left( -1\right) ^{\nu }\mathbf{ad}%
_{\mathbf{B}{_{i,\alpha _{i}}}}^{a_{i}}\mathbf{ad}_{\mathbf{B}{_{i,\alpha
_{i-1}}}}^{a_{i-1}}\ldots \mathbf{ad}_{\mathbf{B}{_{i,\alpha _{j+1}}}%
}^{a_{j+1}}\mathbf{B}{_{i,j}},\ \text{for }a_{k}\neq 0,j<k\leq i,\nu =|%
\mathbf{a}|.  \label{d2nB}
\end{equation}

\subsubsection{Time derivatives of low degree using closed form relations
for partial derivatives}

The time derivative of the body-fixed twist of body $i$ (\ref{Vbi}) is%
\begin{equation}
\dot{\mathbf{V}}_{i}^{\text{b}}=\sum_{j\leq i}\mathbf{B}{_{i,j}}\ddot{q}%
_{j}+\sum_{j<k\leq i}[\mathbf{B}{_{i,j},\mathbf{B}{_{i,k}}]}\dot{q}_{j}\dot{q%
}_{k}.  \label{Vbdot}
\end{equation}%
The second time derivative can be simplified to 
\begin{equation}
\ddot{\mathbf{V}}_{i}^{\text{b}}=\sum_{j\leq i}\mathbf{B}{_{i,j}}\dddot{q}%
_{j}+2%
\hspace{-1ex}%
\sum_{j<k\leq i}[\mathbf{B}{_{i,j},\mathbf{B}{_{i,k}}]}\ddot{q}_{j}\dot{q}%
_{k}+\sum_{j<k\leq i}[\mathbf{B}{_{i,j},\mathbf{B}{_{i,k}}]}\dot{q}_{j}\ddot{%
q}_{k}+2%
\hspace{-2ex}%
\sum_{j<k\leq r\leq i}[[\mathbf{B}{_{i,j},\mathbf{B}{_{i,k}}],}\mathbf{B}{%
_{i,r}}]\dot{q}_{j}\dot{q}_{k}\dot{q}_{r}.  \label{Vbddot}
\end{equation}%
It is straightforward to derive expression for higher time derivatives, but
these become rather involved.

\subsubsection{Time derivatives of arbitrary degree using recursive
relations for time derivatives of instantaneous joint screws}

Similar to (\ref{DSi}) the time derivatives of $\mathbf{B}_{i,j}$ obey a
recursive relation. From (\ref{derBi}) follows that 
\begin{equation}
\dot{\mathbf{B}}_{i,j}=\sum_{j<r\leq i}\left[ \mathbf{B}_{i,j},\mathbf{B}%
_{i,r}\right] \dot{q}_{r}=\left[ \mathbf{B}_{i,j},\mathsf{B}_{i,j}\right]
\label{Bdot}
\end{equation}%
with%
\begin{equation}
\mathsf{B}_{i,j}(\mathbf{q},\dot{\mathbf{q}}):=\sum_{j<r\leq i}\mathbf{B}%
_{i,r}\left( \mathbf{q}\right) \dot{q}_{j}.  \label{BBij}
\end{equation}%
Higher time derivatives of $\mathbf{B}_{i,j}$ follow immediately from (\ref%
{Bdot}) as%
\begin{equation}
\mathrm{D}^{(k)}\mathbf{B}_{i,j}=\sum_{l=0}^{k-1}\tbinom{k-1}{l}[\mathrm{D}%
^{(l)}\mathbf{B}_{i,j},\mathrm{D}^{(k-l-1)}\mathsf{B}_{i,j}]  \label{DkB}
\end{equation}%
where the time derivatives of $\mathsf{B}_{i,j}$ are found from (\ref{BBij})
as%
\begin{equation}
\mathrm{D}^{(k)}\mathsf{B}_{i,j}=\sum_{j<k\leq i}\sum_{l=0}^{k}\tbinom{k}{l}%
\mathrm{D}^{(l)}\mathbf{B}_{i,r}q_{j}^{(k-l+1)}.  \label{DBk}
\end{equation}%
The definition (\ref{BBij}) allows writing the body-fixed twist of body $i$
as $\mathbf{V}_{i}^{\text{b}}=\mathsf{B}_{i,0}(\mathbf{q},\dot{\mathbf{q}})$%
. The $k$th time derivative of the body-fixed twist is thus%
\begin{equation}
\mathrm{D}^{(k)}\mathbf{V}_{i}^{\text{b}}=\mathrm{D}^{(k)}\mathsf{B}_{i,0}(%
\mathbf{q},\dot{\mathbf{q}}).
\end{equation}

\subsection{Hybrid Form of Twists}

\subsubsection{Instantaneous joint screws and the body-fixed Jacobian}

The \emph{hybrid} form of the twist of body $i$ is given by \cite%
{MUBOScrews1,Murray}%
\begin{equation}
\mathbf{V}_{i}^{\text{h}}=\left( 
\begin{array}{c}
\bm{\omega}%
_{i}^{\text{s}} \\ 
\dot{\mathbf{r}}_{i}%
\end{array}%
\right)  \label{DefVh}
\end{equation}%
where $\dot{\mathbf{r}}_{i}$ is the translational velocity of the origin of $%
\mathcal{F}_{i}$ relative to the spatial frame $\mathcal{F}_{0}$ resolved in 
$\mathcal{F}_{0}$. This form is frequently used in robotics and for path
planning \cite{Waldron1982,Whitney1972}, and in computational MBS dynamics.

The hybrid twist is determined analogously to the body-fixed twist (\ref{Vb}%
), except that all vectors are resolved in the inertial frame. Denote with $%
\mathbf{b}_{i,j}$ the position vector of a point on the axis of joint $j$
measured from the origin of $\mathcal{F}_{i}$, and with $\mathbf{e}_{j}$ a
unit vector along the axis,where both are expressed in $\mathcal{F}_{0}$.
The hybrid twist of body $i$ is given by%
\begin{eqnarray}
\mathbf{V}_{i}^{\text{h}} &=&\dot{q}_{1}\left( 
\begin{array}{c}
\mathbf{e}_{1} \\ 
\mathbf{b}_{i,1}\times \mathbf{e}_{1}+h_{1}\mathbf{e}_{1}%
\end{array}%
\right) +\dot{q}_{2}\left( 
\begin{array}{c}
\mathbf{e}_{2} \\ 
\mathbf{b}_{i,2}\times \mathbf{e}_{2}+h_{2}\mathbf{e}_{2}%
\end{array}%
\right) +\ldots +\dot{q}_{i}\left( 
\begin{array}{c}
\mathbf{e}_{i} \\ 
\mathbf{b}_{i,i}\times \mathbf{e}_{i}+h_{i}\mathbf{e}_{i}%
\end{array}%
\right)  \label{Vh} \\
&=&\dot{q}_{1}\mathbf{H}_{i,1}+\dot{q}_{2}\mathbf{H}_{i,2}+\ldots +\dot{q}%
_{i}\mathbf{H}_{i,i}  \notag \\
&=&\mathbf{J}_{i}^{\mathrm{h}}\left( \mathbf{q}\right) \dot{\mathbf{q}} 
\notag
\end{eqnarray}%
where $\mathbf{H}_{i,j}$ is the hybrid form of the screw coordinates of
joint $j$, and the \emph{hybrid Jacobian} of body $i$ is%
\begin{equation}
\mathbf{J}_{i}^{\mathrm{h}}\left( \mathbf{q}\right) :=%
\Big%
(\mathbf{H}_{i,1}%
\hspace{-0.5ex}%
\left( \mathbf{q}\right) 
\Big%
|\cdots 
\Big%
|\mathbf{H}_{i,i}%
\hspace{-0.5ex}%
\left( \mathbf{q}\right) 
\Big%
|\mathbf{0}%
\Big%
|\cdots 
\Big%
|\mathbf{0}%
\Big%
).
\end{equation}

The reference point where the velocity is measured is the same for hybrid
and body-fixed twist. The only difference is the reference frame in which
the vectors are resolved. Therefore the hybrid twist is given as $\mathbf{V}%
_{i}^{\text{h}}=\mathbf{Ad}_{\mathbf{R}_{i}}\mathbf{V}_{i}^{\text{b}}$
(using (\ref{AdR})) and the instantaneous joint screw coordinates are%
\begin{equation}
\mathbf{H}_{i,j}=\mathbf{Ad}_{\mathbf{R}_{i}}\mathbf{B}_{i,j}  \label{Hij}
\end{equation}%
where $\mathbf{R}_{i}$ is the rotation matrix of body $i$ in (\ref{Cbar}).
From (\ref{Bijrec}) follows the recursive relation 
\begin{equation}
\mathbf{H}_{i,j}=\mathbf{Ad}_{\mathbf{r}_{i,i-1}}\mathbf{H}_{i-1,j},j<i
\end{equation}%
with the relative displacement $\mathbf{r}_{i,i-1}=\mathbf{r}_{i}-\mathbf{r}%
_{i-1}$ of body $i$ (i.e. the origin of $\mathcal{F}_{i}$) and body $i-1$
(i.e. the origin of $\mathcal{F}_{i-1}$).

\subsubsection{Explicit recursive relations for low-order}

The first and second time derivative of (\ref{Hij}) is \cite{MUBOScrews2}%
\begin{eqnarray}
\dot{\mathbf{H}}{_{i,j}} &=&(\mathbf{ad}_{\dot{\mathbf{r}}_{i,j}}+\mathbf{Ad}%
_{\mathbf{r}_{i,j-1}}\mathbf{ad}_{\mathbf{\omega }_{j}^{\text{s}}})\mathbf{H}%
_{j,j}  \label{Hdot} \\
\ddot{\mathbf{H}}{_{i,j}} &=&\left( \mathbf{ad}_{\ddot{\mathbf{r}}_{i,j}}+2%
\mathbf{ad}_{\dot{\mathbf{r}}_{i,j}}\mathbf{ad}_{\mathbf{\omega }_{j}^{\text{%
s}}}+\mathbf{Ad}_{\mathbf{r}_{i,j}}(\mathbf{ad}_{\dot{\mathbf{\omega }}_{j}^{%
\text{s}}}+\mathbf{ad}_{\mathbf{\omega }_{j}^{\text{s}}}\mathbf{ad}_{\mathbf{%
\omega }_{j}^{\text{s}}})\mathbf{H}_{j,j}\right) .
\end{eqnarray}%
This yields the explicit expressions for the acceleration and jerk in hybrid
representation%
\begin{eqnarray}
\dot{\mathbf{V}}_{i}^{\text{h}} &=&\sum_{j\leq i}(\mathbf{J}{_{i,j}^{\text{h}%
}}\ddot{q}_{j}+(\mathbf{ad}_{\dot{\mathbf{r}}_{i,j}}+\mathbf{Ad}_{\mathbf{r}%
_{i,j}}\mathbf{ad}_{\mathbf{\omega }_{j}^{\text{s}}})\mathbf{H}_{j,j}\dot{q}%
_{j}) \\
\ddot{\mathbf{V}}_{i}^{\text{h}} &=&\sum_{j\leq i}\left( \mathbf{J}{_{i,j}^{%
\text{h}}}\dddot{q}_{j}+2\mathbf{ad}_{\dot{\mathbf{r}}_{i,j}}\ddot{q}_{j}+%
\big%
(\mathbf{ad}_{\ddot{\mathbf{r}}_{i,j}}+2\mathbf{ad}_{\dot{\mathbf{r}}_{i,j}}%
\mathbf{ad}_{\mathbf{\omega }_{j}^{\text{s}}}%
\big%
)\dot{q}_{j}\right. +\left. \mathbf{Ad}_{\mathbf{r}_{i,j}}%
\big%
(2\mathbf{ad}_{\mathbf{\omega }_{j}^{\text{s}}}\ddot{q}_{j}+\mathbf{ad}_{%
\dot{\mathbf{\omega }}_{j}^{\text{s}}}+\mathbf{ad}_{\mathbf{\omega }_{j}^{%
\text{s}}}\mathbf{ad}_{\mathbf{\omega }_{j}^{\text{s}}}%
\big%
)\dot{q}_{j})\mathbf{H}_{j,j}\right) .
\end{eqnarray}%
These are the core relation in the so-called 'spatial vector' formulation
(i.e. using the hybrid representation of twists) \cite%
{Fijany1995,Jain1991,LillyOrin1991,Rodriguez1987,Rodriguez1992}. In this
context the Lie bracket, respectively screw product, (\ref{ScrewProd}) has
been termed the 'spatial cross product' \cite%
{Featherstone1983,Featherstone2008}.

\subsection{Relation of the different representations}

While the spatial representation is dominantly used in kinematics and
mechanism theory, in various robotic applications body-fixed (\ref{DefVb})
and hybrid representations (\ref{DefVh}) are used to described EE twist.
Also in MBS dynamics these representations are traditionally used. Yet the
spatial representation is deemed to be computationally more efficient, which
is documented by the recently proposed MBS dynamics algorithms \cite%
{Featherstone2008}. The recursive relations for higher time derivatives (\ref%
{DSik}) and (\ref{DBk}) for the twist of body $i$ have the same complexity.
However, when computing this for another body, the spatial version can reuse
the derivatives (\ref{DSik}) since they are body independent. It seems
therefore, desirable to employ the results for the spatial twist even when
using the body-fixed or hybrid version. To this end, in the following, the
twists and their derivatives are related to the spatial twist derivatives.

The three representations of twists are related as follows%
\begin{equation}
\begin{array}{lll}
\mathbf{V}_{i}^{\text{s}}=\mathbf{Ad}_{\mathbf{C}_{i}}\mathbf{V}_{i}^{\text{b%
}},\ \  & \mathbf{V}_{i}^{\text{s}}=\mathbf{Ad}_{\mathbf{r}_{i}}\mathbf{V}%
_{i}^{\text{h}},\ \  & \mathbf{V}_{i}^{\text{h}}=\mathbf{Ad}_{\mathbf{R}_{i}}%
\mathbf{V}_{i}^{\text{b}} \\ 
\mathbf{V}_{i}^{\text{b}}=\mathbf{Ad}_{\mathbf{C}_{i}}^{-1}\mathbf{V}_{i}^{%
\text{s}}, & \mathbf{V}_{i}^{\text{h}}=\mathbf{Ad}_{-\mathbf{r}_{i}}\mathbf{V%
}_{i}^{\text{s}}, & \mathbf{V}_{i}^{\text{b}}=\mathbf{Ad}_{\mathbf{R}%
_{i}^{T}}\mathbf{V}_{i}^{\text{h}}.%
\end{array}
\label{VbVh}
\end{equation}%
using the notations (\ref{Adr}) and (\ref{AdR}), respectively. The
instantaneous screw coordinates and Jacobians transform accordingly.

With $\frac{d}{dt}\mathbf{Ad}_{\mathbf{C}}^{-1}=-\mathbf{Ad}_{\mathbf{C}%
}^{-1}\mathbf{ad}_{\mathbf{V}^{\text{s}}}$, the relation of the time
derivative of spatial and body-fixed twist is obtained from (\ref{VbVh}) as $%
\ \dot{\mathbf{V}}^{\text{s}}=\mathbf{Ad}_{\mathbf{C}}^{-1}\dot{\mathbf{V}}^{%
\text{s}}$. For higher time derivatives this yields%
\begin{equation}
\mathrm{D}^{\left( k\right) }\mathbf{V}^{\text{b}}=\mathbf{Ad}_{\mathbf{C}%
}^{-1}%
\big%
(\sum_{i=0}^{k-1}\binom{k-1}{i}\left( -1\right) ^{i}\mathbf{ad}_{\mathbf{V}^{%
\text{s}}}^{i}\mathrm{D}^{\left( k-i\right) }\mathbf{V}^{\text{s}}%
\big%
),k\geq 1.  \label{VbVs}
\end{equation}%
With (\ref{Adr}) and (\ref{adeta}) it is $\frac{d}{dt}\mathbf{Ad}_{-\mathbf{r%
}}=-\mathbf{ad}_{\dot{\mathbf{r}}}$, and the time derivatives of spatial and
hybrid twists are related by%
\begin{equation}
\mathrm{D}^{\left( k\right) }\mathbf{V}^{\text{h}}=\mathbf{Ad}_{-\mathbf{r}}%
\mathrm{D}^{\left( k\right) }\mathbf{V}^{\text{s}}-\sum_{i=1}^{k}\binom{k}{i}%
\mathbf{ad}_{\mathbf{r}^{\left( i\right) }}\mathrm{D}^{\left( k-i\right) }%
\mathbf{V}^{\text{s}},k\geq 1.  \label{VhVs}
\end{equation}%
The explicit expressions of derivatives of $\mathbf{V}^{\text{b}}$ in terms
of those of $\mathbf{V}^{\text{s}}$, up to fourth-order, are then%
\begin{eqnarray}
\mathbf{V}^{\text{b}} &=&\mathbf{Ad}_{\mathbf{C}}^{-1}\mathbf{V}^{\text{s}%
},\ \ \dot{\mathbf{V}}^{\text{b}}=\mathbf{Ad}_{\mathbf{C}}^{-1}\dot{\mathbf{V%
}}^{\text{s}},\ \ \ddot{\mathbf{V}}^{\text{b}}=\mathbf{Ad}_{\mathbf{C}}^{-1}%
\big%
(\ddot{\mathbf{V}}^{\text{s}}-\mathbf{ad}_{\mathbf{V}^{\text{s}}}\dot{%
\mathbf{V}}^{\text{s}}%
\big%
)  \notag \\
\dddot{\mathbf{V}}^{\text{b}} &=&\mathbf{Ad}_{\mathbf{C}}^{-1}%
\big%
(\dddot{\mathbf{V}}^{\text{s}}-2\mathbf{ad}_{\mathbf{V}^{\mathrm{s}}}\ddot{%
\mathbf{V}}^{\mathrm{s}}+\mathbf{ad}_{\mathbf{V}^{\text{s}}}^{2}\dot{\mathbf{%
V}}^{\text{s}}%
\big%
)  \label{VbVs2} \\
\ddot{\ddot{\mathbf{V}}}^{\text{b}} &=&\mathbf{Ad}_{\mathbf{C}}^{-1}%
\big%
(\ddot{\ddot{\mathbf{V}}}^{\text{s}}-3\mathbf{ad}_{\mathbf{V}^{\text{s}}}%
\dddot{\mathbf{V}}^{\mathrm{s}}+\left( 3\mathbf{ad}_{\mathbf{V}^{\mathrm{s}%
}}^{2}-2\mathbf{ad}_{\dot{\mathbf{V}}^{\mathrm{s}}}\right) \ddot{\mathbf{V}}%
^{\mathrm{s}}+\left( \mathbf{ad}_{\dot{\mathbf{V}}^{\mathrm{s}}}\mathbf{ad}_{%
\mathbf{V}^{\text{s}}}-\mathbf{ad}_{\mathbf{V}^{\mathrm{s}}}^{3}\right) \dot{%
\mathbf{V}}^{\mathrm{s}}%
\big%
)  \notag
\end{eqnarray}%
and of the derivatives of $\mathbf{V}^{\text{h}}$ are%
\begin{eqnarray}
\mathbf{V}^{\mathrm{h}} &=&\mathbf{Ad}_{-\mathbf{r}}\mathbf{V}^{\mathrm{s}%
},\ \ \dot{\mathbf{V}}^{\mathrm{h}}=\mathbf{Ad}_{-\mathbf{r}}\dot{\mathbf{V}}%
^{\mathrm{s}}-\mathbf{ad}_{\dot{\mathbf{r}}}\mathbf{V}^{\mathrm{s}}\text{,}\
\ \ \ddot{\mathbf{V}}^{\text{h}}=\mathbf{Ad}_{-\mathbf{r}}\ddot{\mathbf{V}}^{%
\mathrm{s}}-2\mathbf{ad}_{\dot{\mathbf{r}}}\dot{\mathbf{V}}^{\mathrm{s}}-%
\mathbf{ad}_{\ddot{\mathbf{r}}}\mathbf{V}^{\mathrm{s}}\text{,}  \notag \\
\dddot{\mathbf{V}}^{\mathrm{h}} &=&\mathbf{Ad}_{-\mathbf{r}}\ \dddot{\mathbf{%
V}}^{\mathrm{s}}-3\mathbf{ad}_{\dot{\mathbf{r}}}\ddot{\mathbf{V}}^{\mathrm{s}%
}-3\mathbf{ad}_{\ddot{\mathbf{r}}}\dot{\mathbf{V}}^{\mathrm{s}}-\mathbf{ad}_{%
\dddot{\mathbf{r}}}\mathbf{V}^{\mathrm{s}}  \label{VhVs2} \\
\ddot{\ddot{\mathbf{V}}}^{\mathrm{h}} &=&\mathbf{Ad}_{-\mathbf{r}}\ \ddot{%
\ddot{\mathbf{V}}}^{\mathrm{s}}-4\mathbf{ad}_{\dot{\mathbf{r}}}\dddot{%
\mathbf{V}}^{\mathrm{s}}-6\mathbf{ad}_{\ddot{\mathbf{r}}}\ddot{\mathbf{V}}^{%
\mathrm{s}}-4\mathbf{ad}_{\dddot{\mathbf{r}}}\dot{\mathbf{V}}^{\mathrm{s}}-%
\mathbf{ad}_{\ddot{\ddot{\mathbf{r}}}}\mathbf{V}^{\mathrm{s}}.  \notag
\end{eqnarray}%
The relations (\ref{VbVs}) and (\ref{VhVs}), or their explicit versions (\ref%
{VbVs2}) and (\ref{VhVs2}), allow to determine the time derivatives in
body-fixed and hybrid representations from given time derivatives of spatial
twists, e.g. for the higher-order forward kinematics in sec. \ref{secFWKin1}
using (\ref{DSik}) or the relations (\ref{Vsrec})-(\ref{Vs3dotrec}). Then
the spatial twists merely serve as algorithmic variables.

The relations $\mathbf{V}^{\text{s}}=\mathbf{Ad}_{\mathbf{C}}\mathbf{V}^{%
\text{b}}$, along with $\frac{d}{dt}\mathbf{Ad}_{\mathbf{C}}=\mathbf{ad}_{%
\mathbf{V}^{\text{s}}}\mathbf{Ad}_{\mathbf{C}}$ yields $\dot{\mathbf{V}}^{%
\text{s}}=\mathbf{Ad}_{\mathbf{C}}\dot{\mathbf{V}}^{\text{b}}$, and thus%
\begin{equation}
\mathrm{D}^{\left( k\right) }\mathbf{V}^{\text{s}}=\mathbf{Ad}_{\mathbf{C}}%
\big%
(\sum_{i=0}^{k-1}\binom{k-1}{i}\mathbf{ad}_{\mathbf{V}^{\text{b}}}^{i}%
\mathrm{D}^{\left( k-i\right) }\mathbf{V}^{\text{b}}%
\big%
).  \label{DVsVb}
\end{equation}%
The relation $\mathbf{V}^{\text{s}}=\mathbf{Ad}_{\mathbf{r}}\mathbf{V}^{%
\text{h}}$, along with $\frac{d}{dt}\mathbf{Ad}_{\mathbf{r}}=\mathbf{ad}_{%
\dot{\mathbf{r}}}$, yields%
\begin{equation}
\mathrm{D}^{\left( k\right) }\mathbf{V}^{\text{s}}=\mathbf{Ad}_{\mathbf{r}}%
\big%
(\mathrm{D}^{\left( k\right) }\mathbf{V}^{\text{h}}+\sum_{i=1}^{k}\binom{k}{i%
}\mathbf{ad}_{\mathbf{r}^{\left( i\right) }}\mathrm{D}^{\left( k-i\right) }%
\mathbf{V}^{\text{h}}%
\big%
).
\end{equation}

The relations (\ref{DVsVb}) for the derivatives of $\mathbf{V}^{\text{s}}$
in terms of those $\mathbf{V}^{\text{b}}$ are explicitly, up to degree $k=4$,%
\begin{eqnarray}
\mathbf{V}^{\text{s}} &=&\mathbf{Ad}_{\mathbf{C}}\mathbf{V}^{\text{b}},\ \ 
\dot{\mathbf{V}}^{\text{s}}=\mathbf{Ad}_{\mathbf{C}}\dot{\mathbf{V}}^{\text{b%
}},\ \ \ddot{\mathbf{V}}^{\text{s}}=\mathbf{Ad}_{\mathbf{C}}%
\big%
(\ddot{\mathbf{V}}^{\text{b}}+\mathbf{ad}_{\mathbf{V}^{\text{b}}}\dot{%
\mathbf{V}}^{\text{b}}%
\big%
)  \notag \\
\dddot{\mathbf{V}}^{\text{s}} &=&\mathbf{Ad}_{\mathbf{C}}%
\big%
(\dddot{\mathbf{V}}^{\text{b}}+2\mathbf{ad}_{\mathbf{V}^{\text{b}}}\ddot{%
\mathbf{V}}^{\text{b}}+\mathbf{ad}_{\mathbf{V}^{\text{b}}}^{2}\dot{\mathbf{V}%
}^{\text{b}}%
\big%
)  \label{VsVb} \\
\ddot{\ddot{\mathbf{V}}}^{\text{s}} &=&\mathbf{Ad}_{\mathbf{C}}%
\big%
(\ddot{\ddot{\mathbf{V}}}^{\text{b}}+3\mathbf{ad}_{\mathbf{V}^{\text{b}}}%
\dddot{\mathbf{V}}^{\text{b}}+3\mathbf{ad}_{\mathbf{V}^{\text{b}}}^{2}\ddot{%
\mathbf{V}}^{\text{b}}+\mathbf{ad}_{\mathbf{V}^{\text{b}}}^{3}\dot{\mathbf{V}%
}^{\text{b}}-\mathbf{ad}_{\dot{\mathbf{V}}^{\text{b}}}^{2}\mathbf{V}^{\text{b%
}}%
\big%
)  \notag
\end{eqnarray}%
and in terms of the derivatives of $\mathbf{V}^{\text{h}}$ these are%
\begin{eqnarray}
\mathbf{V}^{\text{s}} &=&\mathbf{Ad}_{\mathbf{r}}\mathbf{V}^{\text{h}},\ \ 
\dot{\mathbf{V}}^{\text{s}}=\mathbf{Ad}_{\mathbf{r}}\dot{\mathbf{V}}^{\text{h%
}}+\mathbf{ad}_{\dot{\mathbf{r}}}\mathbf{V}^{\text{h}},\ \ \ \ddot{\mathbf{V}%
}^{\text{s}}=\mathbf{Ad}_{\mathbf{r}}\ \ddot{\mathbf{V}}^{\text{h}}+2\mathbf{%
ad}_{\dot{\mathbf{r}}}\dot{\mathbf{V}}^{\text{h}}+\mathbf{ad}_{\ddot{\mathbf{%
r}}}\mathbf{V}^{\text{h}}  \notag \\
\dddot{\mathbf{V}}^{\text{s}} &=&\mathbf{Ad}_{\mathbf{r}}\ \dddot{\mathbf{V}}%
^{\text{h}}+3\mathbf{ad}_{\dot{\mathbf{r}}}\ddot{\mathbf{V}}^{\text{h}}+3%
\mathbf{ad}_{\ddot{\mathbf{r}}}\dot{\mathbf{V}}^{\text{h}}+\mathbf{ad}_{%
\dddot{\mathbf{r}}}\mathbf{V}^{\text{h}}  \label{VsVh} \\
\ddot{\ddot{\mathbf{V}}}^{\text{s}} &=&\mathbf{Ad}_{\mathbf{r}}\ \ddot{\ddot{%
\mathbf{V}}}^{\text{h}}+4\mathbf{ad}_{\dot{\mathbf{r}}}\dddot{\mathbf{V}}^{%
\text{h}}+6\mathbf{ad}_{\ddot{\mathbf{r}}}\ddot{\mathbf{V}}^{\text{h}}+4%
\mathbf{ad}_{\dddot{\mathbf{r}}}\dot{\mathbf{V}}^{\text{h}}+\mathbf{ad}_{%
\ddot{\ddot{\mathbf{r}}}}\mathbf{V}^{\text{h}}  \notag
\end{eqnarray}%
where $\dot{\mathbf{r}},\ddot{\mathbf{r}},\dddot{\mathbf{r}},\ddot{\ddot{%
\mathbf{r}}}$ are known as part of $\mathbf{V}^{\text{h}},\dot{\mathbf{V}}^{%
\text{h}},\ddot{\mathbf{V}}^{\text{b}},\dddot{\mathbf{V}}^{\text{b}}$.

These expressions admit to use the inverse kinematics algorithm in sec. \ref%
{secInvKin}. Then (\ref{VsVb}), respectively (\ref{VsVh}), are used to
determine the EE twist $\mathbf{V}_{n}^{\text{s}}$ and its derivatives. The
necessary derivatives of $\mathbf{V}_{i}^{\text{s}},i\leq n$ of degree less
than $k$ are determined in step 2 of the algorithm evaluating (\ref{DSik}).
Likewise the explicit recursions (\ref{qdots1}-\ref{qdots4}) together with (%
\ref{Vsdotrec})-(\ref{Vs3dotrec}) can be used up to degree 4. Notice that
the spatial twists again only serve as algorithmic variables.

\section{Conclusion and Outlook}

Formulating the kinematics of serial and closed loop linkages using the POE
provides a flexible modeling approach, in terms of readily available
geometric data, and gives rise to very compact and efficient expressions for
all derivatives necessary for various kinematic and dynamic tasks. The basic
(including some novel) relations were summarized in this paper and it was
shown how they can be applied to some of the basic tasks in robotics and
mechanism analysis. The formulations were derived for the spatial
representation of twists. The latter leads to efficient recursive
formulations, which is why it is used in mechanisms theory and recently also
in MBS dynamics. Yet, in several robotics applications the motion is
prescribed using the body-fixed or hybrid representation of twists. Their
relations have been briefly discussed in section \ref{secRepresentations}.
The derivations of the reported formulae using these representations is an
open research topic. Future research will in particular investigate and
compare the numerical efficiency of the relations using different
representations.

In order to make the formulations easier accessible to the reader, several
of the presented relations have been implemented in a Mathematica$^{%
\copyright }$ package that is available as supplementary data in \cite%
{MendeleyDataset}.

\appendix%

\section{Geometric background%
\label{secGeomBackground}%
}

\subsection{Exponential mapping%
\label{secExp}%
}

In the following, the relevant facts about the geometry of rigid body
motions and screws is presented as far as necessary to follow the paper. A
thorough account of the Lie group structure and its relevance for kinematics
can be found in the textbooks \cite{LynchPark2017,Murray,Selig}.

Consider two frames $\bar{\mathcal{F}}_{1}$ and $\bar{\mathcal{F}}_{2}$ that
are moving relative to one another, while initially both frames coincide.
The transformation of homogenous point coordinates from $\bar{\mathcal{F}}%
_{2}$ to $\bar{\mathcal{F}}_{1}$ is described by a matrix 
\begin{equation}
\bar{\mathbf{C}}_{12}=\left( 
\begin{array}{cc}
\bar{\mathbf{R}}_{12} & {^{1}}\bar{\mathbf{r}}_{12} \\ 
\mathbf{0} & 1%
\end{array}%
\right) \in SE\left( 3\right) ,  \label{C12}
\end{equation}%
where ${^{1}}\bar{\mathbf{r}}_{12}$ is the position vector from the origin
of $\bar{\mathcal{F}}_{1}$ to that of $\bar{\mathcal{F}}_{2}$ resolved in $%
\bar{\mathcal{F}}_{1}$ (superscript indicates the frame in which a vector is
resolved), and $\bar{\mathbf{R}}_{12}\in SO\left( 3\right) $ is the rotation
matrix transforming coordinate vectors resolved in $\bar{\mathcal{F}}_{2}$
to those resolved in $\bar{\mathcal{F}}_{1}$. For instance, consider a point 
$P$ fixed to $\bar{\mathcal{F}}_{2}$. Denoted with ${^{2}}\mathbf{x}_{2%
\mathrm{P}}$ its position vector expressed in $\bar{\mathcal{F}}_{2}$, then $%
{^{1}}\mathbf{x}_{1\mathrm{P}}=\bar{\mathbf{R}}_{12}{^{2}}\mathbf{x}_{2%
\mathrm{P}}+{^{1}}\bar{\mathbf{r}}_{12}$ is the position vector of $P$
expressed in $\bar{\mathcal{F}}_{1}$. Since this is true for any point fixed
in $\bar{\mathcal{F}}_{2}$, the $\bar{\mathbf{C}}_{12}$ itself describes the
relative configuration of $\bar{\mathcal{F}}_{2}$ w.r.t. $\bar{\mathcal{F}}%
_{1}$.

The motion of a frame, i.e. of a rigid body, is a screw motion that can be
expressed in terms of an instantaneous screw axis. Let ${^{1}}\mathbf{e}$ be
a unit vector along the screw axis, and ${^{1}}\mathbf{p}\in {\mathbb{R}}%
^{3} $ be a vector to any point on that axis. Then the screw coordinates
associated to the motion of $\bar{\mathcal{F}}_{2}$ relative to $\mathcal{%
\bar{F}}_{1}$, represented in $\bar{\mathcal{F}}_{1}$, are ${^{1}\mathbf{X}}%
=\left( {^{1}}\mathbf{e},{^{1}}\mathbf{p}\times {^{1}}\mathbf{e}+{^{1}}%
\mathbf{e}h\right) ^{T}$, where $h$ is the pitch of the screw. It should be
noticed, that screws are traditionally denoted with the symbol $\$$ \cite%
{Hunt1978}.

There is unique correspondence of a screw coordinate vector and a $4\times 4$
matrix as follows (omitting superscripts)%
\begin{equation}
\mathbf{X}=\left( 
\begin{array}{c}
\mathbf{e} \\ 
\mathbf{p}\times \mathbf{e}+\mathbf{e}h%
\end{array}%
\right) \ \in {\mathbb{R}}^{6}\ \leftrightarrow \ \ \widehat{\mathbf{X}}%
=\left( 
\begin{array}{cc}
\widetilde{\mathbf{e}} & \ \ \ \mathbf{p}\times \mathbf{e}+\mathbf{e}h \\ 
\mathbf{0} & 0%
\end{array}%
\right) \in se\left( 3\right) ,  \label{Xhat}
\end{equation}%
and generally, for an arbitrarily given screw coordinate vector,%
\begin{equation}
\mathbf{X}=\left( 
\begin{array}{c}
\bm{\xi}
\\ 
\bm{\eta}%
\end{array}%
\right) \ \in {\mathbb{R}}^{6}\ \leftrightarrow \ \ \widehat{\mathbf{X}}%
=\left( 
\begin{array}{cc}
\widetilde{%
\bm{\xi}%
} & \ \ 
\bm{\eta}
\\ 
\mathbf{0} & 0%
\end{array}%
\right) \in se\left( 3\right) .  \label{hat}
\end{equation}%
The screw motion of $\bar{\mathcal{F}}_{2}$ relative to $\bar{\mathcal{F}}%
_{1}$ is given as $\bar{\mathbf{C}}_{12}=\exp ({^{1}}\widehat{\mathbf{X}}%
\varphi )$. The exponential mapping attains the closed form%
\begin{equation}
\exp (\varphi \widehat{\mathbf{X}})=\left( 
\begin{array}{cc}
\exp (\varphi \widetilde{\mathbf{e}}) & \ \ \ (\mathbf{I}-\exp (\varphi 
\widetilde{\mathbf{e}}))\mathbf{p}+\varphi h\mathbf{e} \\ 
\mathbf{0} & 1%
\end{array}%
\right) ,\ \text{for }h\neq \infty  \label{SE3expP}
\end{equation}%
with the rotation angle $\varphi $ , and for pure translations, i.e.
infinite pitch,%
\begin{equation}
\exp (\varphi \widehat{\mathbf{X}})=\left( 
\begin{array}{cc}
\mathbf{I} & \ \varphi \mathbf{e} \\ 
\mathbf{0} & 1%
\end{array}%
\right) ,\ \text{for }h=\infty  \label{SE3expP2}
\end{equation}%
where now $\varphi $ is the translation variable. The rotation matrix is
given by the exp mapping on $SO\left( 3\right) $, which possesses the
following closed form expressions (known as Euler-Rodrigues formula) \cite%
{CND2016}%
\begin{eqnarray}
\exp \widehat{\mathbf{x}} &=&\mathbf{I}+\tfrac{\sin \left\Vert \mathbf{x}%
\right\Vert \,}{\left\Vert \mathbf{x}\right\Vert }\widehat{\mathbf{x}}+%
\tfrac{1-\cos \left\Vert \mathbf{x}\right\Vert }{\left\Vert \mathbf{x}%
\right\Vert ^{2}}\,\widehat{\mathbf{x}}^{2}  \label{SO3exp} \\
&=&\mathbf{I}+\mathrm{sinc}\left\Vert \mathbf{x}\right\Vert \widehat{\mathbf{%
x}}+\tfrac{1}{2}\mathrm{sinc}^{2}\tfrac{\left\Vert \mathbf{x}\right\Vert }{2}%
\,\widehat{\mathbf{x}}^{2}  \label{SO3exp2} \\
&=&\mathbf{I}+\sin \varphi \widehat{\mathbf{n}}+\left( 1-\cos \varphi
\right) \,\widehat{\mathbf{n}}^{2}.  \label{SO3exp3}
\end{eqnarray}%
The formulae (\ref{SE3expP}) and (\ref{SE3expP2}) are advantageous since
they describe the frame transformation matrix explicitly in terms of the
screw characteristics \thinspace $\mathbf{p}$, $\mathbf{e}$, and $h$.
Another closed form is available \cite{CND2016,Murray} in terms of a general
screw coordinate vector of the form (\ref{hat}), which is not relevant for
this paper. For sake of simplicity, frequently, the screw coordinate vector
is considered as argument of the exp mapping, and the notion $\exp (\varphi 
\mathbf{X})$ is used.

Clearly, for $\varphi =0$ the exp mapping yields the identity matrix, $\bar{%
\mathbf{C}}_{12}=\mathbf{I}$, i.e. both frames coincide in the reference
configuration. The frame $\bar{\mathcal{F}}_{2}$ is obviously rather
special. The motion of a general frame $\mathcal{F}_{2}$, which is rigidly
connected to $\bar{\mathcal{F}}_{2}$, is obtained via a subsequent
transformation $\mathbf{A}_{2}\in SE\left( 3\right) $ from $\mathcal{F}_{2}$
to $\bar{\mathcal{F}}_{2}$ so that $\mathbf{C}_{12}=\bar{\mathbf{C}}_{12}%
\mathbf{A}_{2}$ is the transformation from $\mathcal{F}_{2}$ to $\bar{%
\mathcal{F}}_{1}$. In the zero reference configuration ($\varphi =0$), it is 
$\mathbf{C}_{12}=\mathbf{A}_{2}$. Therefore, $\mathbf{A}_{2}$ is numerically
identical to the the transformation from $\mathcal{F}_{2}$ to $\bar{\mathcal{%
F}}_{1}$ in the reference configuration.

It is important to observe that all vectors in (\ref{SE3expP}) and (\ref%
{SE3expP2}) are resolved in $\bar{\mathcal{F}}_{1}$, and that the matrix $%
\exp ({^{1}}\widehat{\mathbf{X}}\varphi )$ in (\ref{C12}) transforms point
coordinates measured in $\bar{\mathcal{F}}_{2}$ back to those measured in
frame $\bar{\mathcal{F}}_{1}$.

Throughout the paper, the sub- and superscripts referring to the world frame 
$\mathcal{F}_{0}$ are omitted, i.e. $\bar{\mathbf{r}}$ is used instead of ${%
^{0}}\bar{\mathbf{r}}$, $\mathbf{X}$ is used instead of ${^{0}}\mathbf{X}$,
and $\mathbf{C}_{k}$ instead of $\mathbf{C}_{0k}$ for the configuration
relative to the world frame.

For a constant screw coordinate vector the derivative of exp attains the
simple form 
\begin{equation}
\frac{d}{dt}\exp (t\widehat{\mathbf{X}})=\exp (t\widehat{\mathbf{X}})%
\widehat{\mathbf{X}}=\widehat{\mathbf{X}}\exp (t\widehat{\mathbf{X}}).
\label{dexpdt}
\end{equation}%
This identity follows from the fact that the screw axis is invariant under a
screw motion about that axis: $\widehat{\mathbf{X}}=\mathrm{Ad}_{\exp (t%
\widehat{\mathbf{X}})}(\widehat{\mathbf{X}})=\exp (t\widehat{\mathbf{X}})%
\widehat{\mathbf{X}}\exp (-t\widehat{\mathbf{X}})$ and thus $\exp (t\widehat{%
\mathbf{X}})\widehat{\mathbf{X}}=\widehat{\mathbf{X}}\exp (t\widehat{\mathbf{%
X}})$.

\subsection{Frame Transformations of Screw Coordinates}

Let $\mathbf{S}_{1,2}$ be the matrix transforming homogenous point
coordinates expressed in frame $\mathcal{F}_{2}$ to those expressed in frame 
$\mathcal{F}_{1}$. Then the screw coordinates transform according to%
\begin{equation*}
{^{1}\mathbf{X}}=\left( 
\begin{array}{c}
{^{1}}\mathbf{e} \\ 
\ {^{1}}\mathbf{p}_{1}\times {^{1}}\mathbf{e}+{^{1}}\mathbf{e}h%
\end{array}%
\right) =\left( 
\begin{array}{cc}
\mathbf{R}_{1,2} & \mathbf{0} \\ 
{^{1}}\widetilde{\mathbf{d}}_{1,2}\mathbf{R}_{1,2}\ \  & \mathbf{R}_{1,2}%
\end{array}%
\right) \left( 
\begin{array}{c}
{^{2}}\mathbf{e} \\ 
\ {^{2}}\mathbf{p}_{2}\times {^{2}}\mathbf{e}+{^{2}}\mathbf{e}h%
\end{array}%
\right) =\mathbf{Ad}_{\mathbf{S}_{1,2}}{^{2}}\mathbf{X}
\end{equation*}%
where%
\begin{equation}
\mathbf{Ad}_{\mathbf{C}}=\left( 
\begin{array}{cc}
\mathbf{R} & \mathbf{0} \\ 
\widetilde{\mathbf{r}}\mathbf{R}\ \  & \mathbf{R}%
\end{array}%
\right)  \label{Ad}
\end{equation}%
is the \emph{adjoint transformation} matrix to $\mathbf{C}\in SE\left(
3\right) $. $\mathbf{C}$ describes a frame transformation, while $\mathbf{Ad}%
_{\mathbf{C}}$ describes the corresponding transformation of screw
coordinates that belong to $se\left( 3\right) $. Using the $4\times 4$
representation of screw coordinates, the adjoint transformation $\mathbf{Ad}%
_{\mathbf{C}}\mathbf{X}$ is described by 
\begin{equation}
\mathrm{Ad}_{\mathbf{C}}({\widehat{\mathbf{X}}})=\mathbf{C}{\widehat{\mathbf{%
X}}}\mathbf{C}^{-1}.  \label{Ad1}
\end{equation}%
A helpful property is that%
\begin{equation}
\mathbf{Ad}_{\mathbf{C}_{1}\mathbf{C}_{2}}=\mathbf{Ad}_{\mathbf{C}_{1}}%
\mathbf{Ad}_{\mathbf{C}_{2}}.  \label{Ad12}
\end{equation}%
With slight abuse of notation, the adjoint transformation for pure
translations (i.e. $\mathbf{R}=\mathbf{I}$) is denoted with%
\begin{equation}
\mathbf{Ad}_{\mathbf{r}}=\left( 
\begin{array}{cc}
\mathbf{I} & \mathbf{0} \\ 
\widetilde{\mathbf{r}} & \mathbf{I}%
\end{array}%
\right)  \label{Adr}
\end{equation}%
and for pure rotations with 
\begin{equation}
\mathbf{Ad}_{\mathbf{R}}=\left( 
\begin{array}{cc}
\mathbf{R} & \mathbf{0} \\ 
\mathbf{0} & \mathbf{R}%
\end{array}%
\right) .  \label{AdR}
\end{equation}

The tangential aspect of the frame transformation of a constant screw
coordinate vector $\mathbf{Y}$ under the screw motion $\exp (t\widehat{%
\mathbf{X}})$ is determined by%
\begin{eqnarray}
\frac{d}{dt}\mathrm{Ad}_{\exp (t\widehat{\mathbf{X}})}({\widehat{\mathbf{Y}}}%
) &=&\frac{d}{dt}\exp (t\widehat{\mathbf{X}})\widehat{\mathbf{Y}}\exp (-t%
\widehat{\mathbf{X}})+\exp (t\widehat{\mathbf{X}})\widehat{\mathbf{Y}}\frac{d%
}{dt}\exp (-t\widehat{\mathbf{X}})  \notag \\
&=&\widehat{\mathbf{X}}\exp (t\widehat{\mathbf{X}})\widehat{\mathbf{Y}}\exp
(-t\widehat{\mathbf{X}})-\exp (t\widehat{\mathbf{X}})\widehat{\mathbf{Y}}%
\exp (-t\widehat{\mathbf{X}})\widehat{\mathbf{X}}  \notag \\
&=&\widehat{\mathbf{X}}\mathrm{Ad}_{\exp (t\widehat{\mathbf{X}})}({\widehat{%
\mathbf{Y}}})-\mathrm{Ad}_{\exp (t\widehat{\mathbf{X}})}({\widehat{\mathbf{Y}%
}}){\widehat{\mathbf{X}}}  \notag \\
&{=}&{[\widehat{\mathbf{X}},\mathrm{Ad}_{\exp (t\widehat{\mathbf{X}})}({%
\widehat{\mathbf{Y}}})]=:\mathrm{ad}}_{{\widehat{\mathbf{X}}}}({\mathrm{Ad}%
_{\exp (t\widehat{\mathbf{X}})}({\widehat{\mathbf{Y}}}))}  \label{dAd}
\end{eqnarray}%
where ${[\widehat{\mathbf{X}}}_{1}{,{\widehat{\mathbf{X}}}}_{2}{]=\widehat{%
\mathbf{X}}}_{1}{{\widehat{\mathbf{X}}}}_{2}-{{\widehat{\mathbf{X}}}}_{2}{%
\widehat{\mathbf{X}}}_{1}$ is the commutator of the matrices ${\widehat{%
\mathbf{X}}}_{1}${\ }and ${{\widehat{\mathbf{X}}}}_{2}$ that defines the 
\emph{Lie bracket} on $se\left( 3\right) $. Since in particular ${\mathrm{ad}%
}_{{\widehat{\mathbf{X}}}}\widehat{\mathbf{Y}}=[\widehat{\mathbf{X}},%
\widehat{\mathbf{Y}}]=\frac{d}{dt}\mathrm{Ad}_{\exp (t\widehat{\mathbf{X}}%
)}\left. ({\widehat{\mathbf{Y}}})\right\vert _{t=0}$ the Lie bracket is
regarded as the adjoint operator on $se\left( 3\right) $. Throughout the
paper either notation is used depending on which one simplifies the
notation. The Lie bracket reads explicitly%
\begin{equation}
\lbrack \widehat{\mathbf{X}}_{1},\widehat{\mathbf{X}}_{2}]={\mathrm{ad}}_{{%
\widehat{\mathbf{X}}}_{1}}(\widehat{\mathbf{X}}_{2})=\left( 
\begin{array}{cc}
\widetilde{%
\bm{\xi}%
}_{1}\widetilde{%
\bm{\xi}%
}_{2}-\widetilde{%
\bm{\xi}%
}_{2}\widetilde{%
\bm{\xi}%
}_{1}\ \  & \widetilde{%
\bm{\xi}%
}_{1}%
\bm{\eta}%
_{2}-\widetilde{%
\bm{\xi}%
}_{2}%
\bm{\eta}%
_{1} \\ 
\mathbf{0} & 0%
\end{array}%
\right) =\left( 
\begin{array}{cc}
\widetilde{%
\bm{\xi}%
_{1}\times 
\bm{\xi}%
_{2}}\ \  & 
\bm{\eta}%
_{1}\times 
\bm{\xi}%
_{2}+%
\bm{\xi}%
_{1}\times 
\bm{\eta}%
_{2} \\ 
\mathbf{0} & 0%
\end{array}%
\right) .  \label{LieBracket}
\end{equation}%
Applying the correspondence (\ref{hat}), this can be represented in vector
notation of screws as%
\begin{equation}
\lbrack \mathbf{X}_{1},\mathbf{X}_{2}]:=\mathbf{a\mathbf{d}_{\mathbf{X}_{1}}X%
}_{2}=\left( 
\begin{array}{cc}
\widetilde{%
\bm{\xi}%
}_{1}\ \  & \mathbf{0} \\ 
\widetilde{%
\bm{\eta}%
}_{1}\ \  & \widetilde{%
\bm{\xi}%
}_{1}%
\end{array}%
\right) \left( 
\begin{array}{c}
\bm{\xi}%
_{2} \\ 
\bm{\eta}%
_{2}%
\end{array}%
\right) =\left( 
\begin{array}{c}
\bm{\xi}%
_{1}\times 
\bm{\xi}%
_{2} \\ 
\bm{\eta}%
_{1}\times 
\bm{\xi}%
_{2}+%
\bm{\xi}%
_{1}\times 
\bm{\eta}%
_{2}%
\end{array}%
\right)  \label{ScrewProd}
\end{equation}%
where the ${\mathrm{ad}}_{{\widehat{\mathbf{X}}}}$ operator is represented
by the matrix 
\begin{equation}
\mathbf{a\mathbf{d}_{\mathbf{X}}}=\left( 
\begin{array}{cc}
\widetilde{%
\bm{\xi}%
}\ \  & \mathbf{0} \\ 
\widetilde{%
\bm{\eta}%
}\ \  & \widetilde{%
\bm{\xi}%
}%
\end{array}%
\right) .  \label{ad}
\end{equation}%
The relation (\ref{hat}) is thus an isomorphism of $se\left( 3\right) $ and $%
{\mathbb{R}}^{6}$ as Lie algebra, with the corresponding Lie bracket (\ref%
{LieBracket}) and (\ref{ScrewProd}), respectively. The vector form (\ref%
{ScrewProd}) of Lie bracket is referred to as the \emph{screw product} of
the screw coordinate vectors $\mathbf{X}_{1}$ and $\mathbf{X}_{2}$.
Throughout the paper, the vector notation is used, which is also preferable
for actual implementations. In vector notation, the relation (\ref{dAd})
reads%
\begin{equation}
\frac{d}{dt}\mathbf{Ad}_{\exp \left( t\mathbf{X}\right) }\mathbf{Y}=\mathbf{a%
\mathbf{d}_{\mathbf{X}}\mathbf{Ad}}_{\exp \left( t\mathbf{X}\right) }\mathbf{%
Y}=[\mathbf{X},\mathbf{\mathbf{Ad}}_{\exp \left( t\mathbf{X}\right) }\mathbf{%
Y}].  \label{dAd2}
\end{equation}%
In analogy to (\ref{Adr}), the notation%
\begin{equation}
\mathbf{a\mathbf{d}_{%
\bm{\eta}%
}}=\left( 
\begin{array}{cc}
\mathbf{0}\ \  & \mathbf{0} \\ 
\widetilde{%
\bm{\eta}%
}\ \  & \mathbf{0}%
\end{array}%
\right)  \label{adeta}
\end{equation}%
is used when appropriate.

The Lie bracket is bilinear, i.e. $[a\mathbf{X},b\mathbf{Y}]=ab[\mathbf{X},%
\mathbf{Y}]$, and satisfies the Jacobi identity%
\begin{eqnarray}
\lbrack \mathbf{X},[\mathbf{Y},\mathbf{Z}]]+[\mathbf{Z},[\mathbf{X},\mathbf{Y%
}]]+[\mathbf{Y},[\mathbf{Z},\mathbf{X}]] &=&\mathbf{0}  \label{JacobiIdent}
\\
\lbrack \lbrack \mathbf{X},\mathbf{Y}],\mathbf{Z}]+[[\mathbf{Y},\mathbf{Z}],%
\mathbf{X}]+[[\mathbf{Z},\mathbf{X}],\mathbf{Y}] &=&\mathbf{0}  \notag
\end{eqnarray}%
The Ad operator is linear, so that $\mathbf{Ad}_{\mathbf{C}}[\mathbf{X},%
\mathbf{Y}]=[\mathbf{Ad}_{\mathbf{C}}\mathbf{X},\mathbf{Ad}_{\mathbf{C}}%
\mathbf{Y}]$. This shows that the screw product is frame invariant, in the
sense that it can be computed in an arbitrary frame and the result be
transformed in another frame of reference. The reader will recognize this
fact for the ordinary cross product (the first component in (\ref{ScrewProd}%
)).

\section{Notation%
\label{secNotation}%
}

Denote with $n_{\mathrm{R}}$ the number of revolute joints, and with $n_{%
\mathrm{P}}$ the number of prismatic or helical joints of a kinematic chain.
The parameter manifold of this kinematic chain is ${\mathbb{V}}^{n}={\mathbb{%
R}}^{n_{\mathrm{P}}}\times {\mathbb{T}}^{n_{\mathrm{R}}}$, where $n$ is the
total number of 1-DOF screw joints.

Joints are indexed with lattin letters $i,j,k,l\in \{1,\ldots ,n\}$.
Alternatively, instead of lattin letter, the indexes $\alpha _{1},\alpha
_{2},\alpha _{3},\ldots \in \{1,\ldots ,n\}$ are used when dealing with a
large number of indexes.

Also the multi-index notation is used. Properties and applications can be
found e.g. in \cite{Raymond1991}. A multi-index is a $n$-tuple of the form $%
\mathbf{a}=\left( a_{1},a_{2},\ldots ,a_{n}\right) \in \mathbb{N}^{n}$. The
norm of $\mathbf{a}$ is $\left\vert \mathbf{a}\right\vert
:=a_{1}+a_{2}+\ldots +a_{n}$. The multi-factorial of a multi-index is
defined as $\mathbf{a}!=a_{1}!a_{2}!\cdots a_{n}!$.

Let $\mathbf{x}=\left( x_{1},\ldots ,x_{n}\right) \in {\mathbb{R}}^{n}$ be
an $n$-tuple. The power of $\mathbf{x}$ by $\mathbf{a}$ is defined as $%
\mathbf{x}^{\mathbf{a}}=x^{a_{1}}x^{a_{2}}\cdots x^{a_{n}}$. This is a
monomial of degree $\left\vert \mathbf{a}\right\vert $. Analogously, the
multi-index defines an ordered (right) product of matrices $\mathbf{M}%
_{1},\ldots ,\mathbf{M}_{n}$ as%
\begin{equation}
\prod\limits_{1\leq j\leq n}\mathbf{M}_{j}^{a_{j}}=\mathbf{M}_{1}^{a_{1}}%
\mathbf{M}_{2}^{a_{2}}\cdots \mathbf{M}_{n}^{a_{n}}.  \label{multiProd}
\end{equation}%
Being ordered refers to the fact that the matrices are arranged with
increasing index from left to right.

The multiple partial derivative operator is defined as $\partial ^{\mathbf{a}%
}=\left( \frac{\partial }{\partial q_{1}}\right) ^{a_{1}}\ldots \left( \frac{%
\partial }{\partial q_{n}}\right) ^{a_{n}}$ so that $\partial ^{\mathbf{a}}f=%
\frac{\partial ^{k}f}{\partial q_{1}^{a_{1}}\partial q_{2}^{a_{2}}\ldots
\partial q_{n}^{a_{n}}}$ when applied to a $C^{k}$ mapping $f$. This is a
compact form for expressing the $k$th partial derivative, where $a_{i}$
indicates the number of derivations w.r.t. $q_{i}$.

A truncated multi-index is defined as $\mathbf{a}_{i}=\left(
a_{1},a_{2},\ldots ,a_{i}\right) \in \mathbb{N}^{i}$ by removing the last $%
n-i$ indexes from $\mathbf{a}$.

Throughout the paper, the compact summation convention $\sum_{l\leq
i}=\sum_{l=1}^{i}$ is used whenever it is clear that the index $l$ start
with 1. Also the compact form $\sum_{j\leq l\leq
i}=\sum_{j=1}^{l}\sum_{l=1}^{i}$ of the double summation, and for multiple
summations are used.

The abbreviated notions $\mathrm{D}^{(k)}:=\frac{d^{k}}{dt^{k}}$ and $%
q_{j}^{(k)}%
\hspace{-0.5ex}%
:=%
\hspace{-0.2ex}%
\frac{d^{k}}{dt^{k}}q_{j}$ are used for multiple derivatives.

The $k\times k$ identity matrix is denoted with $\mathbf{I}_{k}$.

Finally, $_{i}\mathbf{r}\in {\mathbb{R}}^{3}$ denotes the coordinate vector
when vector $\mathbf{r}$ is resolved in frame $\mathcal{F}_{i}$.

\section*{Acknowledgment}

This work has been partly supported by the LCM -- K2 Center within the
framework of the Austrian COMET-K2 program

\end{document}

%% file: abstract.tex
\textit{Abstract--} The motions of mechanisms can be described in terms of
screw coordinates by means of an exponential mapping. The product of
exponentials (POE) describes the configuration of a chain of bodies
connected by lower pair joints. The kinematics is thus given in terms of
joint screws. The POE serves to express loop constraints for mechanisms as
well as the forward kinematics of serial manipulators. Besides the compact
formulations, the POE gives rise to purely algebraic relations for
derivatives wrt. joint variables. It is known that the partial derivatives
of the instantaneous joint screws (columns of the geometric Jacobian) are
determined by Lie brackets the joint screws. Lesser-known is that derivative
of arbitrary order can be compactly expressed by Lie brackets. This has
significance for higher-order forward/inverse kinematics and dynamics of
robots and multibody systems. Various relations were reported but are
scattered in the literature and insufficiently recognized. This paper aims
to provide a comprehensive overview of the relevant relations. Its original
contributions are closed form and recursive relations for higher-order
derivatives and Taylor expansions of various kinematic relations. Their
application to kinematic control and dynamics of robotic manipulators and
multibody systems is discussed.

\textit{Keywords--} Kinematics, dynamics, screws, product of exponentials, {%
kinematic mapping, higher-order derivatives, inverse kinematics, inverse
dynamics, series elastic actuators, model-based control, }Taylor series

%% file: DifferentialsKinMapping_final.bbl
\begin{thebibliography}{999}
\bibitem{Abdel-Malek} K. Abdel-Malek, W. Yu, J. Yang: Placement of Robot
Manipulators to Maximize Dexterity, International Journal of Robotics and
Automation, Vol. 19, No. 1, 2004

\bibitem{AndersonCritchley2003} K.S. Anderson, J.H. Critchley: Improved
`Order-N' Performance Algorithm for the Simulation of Constrained
Multi-Rigid-Body Dynamic Systems, Multibody System Dynamics, Vol. 9, No. 2,
2003, pp 185-212

\bibitem{Angeles1992} J. Angeles: The Design of Isotropic Manipulator
Architectures in the Presence of Redundancies, The International Journal of
Robotics Research, Vol, 11, 1992, pp. 196-201

\bibitem{AngelesMMT2006} J. Angeles: Is there a characteristic length of a
rigid-body displacement?, Mechanism and Machine Theory, Vol. 41, 2006, pp.
884-96

\bibitem{Angeles2007} J. Angeles: Fundamentals of robotic mechanical systems
- 3rd edition, Springer, 2007

\bibitem{Asada} H. Asada: A Geometrical Representation of Manipulator
Dynamics and its Application to Arm Design, ASME J. Dynamic Syst., Meas.,
Contr., Vol. 105, pp. 131-142

\bibitem{Bae2001} D. S. Bae et al.: A generalized recursive formulation for
constrained flexible multibody dynamics, Int. J. Numer. Meth. Engng, Vol.
50, 2001, pp. 1841-1859

\bibitem{Bayle} B. Bayle, J.Y. Fourquet, M. Renaud: Manipulability analysis
for mobile manipulators, IEEE International Conference on Robotics and
Automation (ICRA), 2001, pp. 1251 - 1256

\bibitem{BeltaKumar2002} C. Belta, V. Kumar: On the Computation of Rigid
Body Motion, Proc. Computational Kinematics Conference, Seoul, South Korea,
2001. Published in Electronic Journal of Computational Kinematics (EJCK),
Vol. 1, No. 1, 2002

\bibitem{Blaschke1911} W. Blaschke: Euklidische Kinematik und
nichteuklidische Geometrie, I, II. Zeit. f. Math. u. Phys. Vol. 60, 1911,
pp. 61-91 and 203-204

\bibitem{Blaschke1942} W. Blaschke: Nicht-Euklidische Geometrie und
Mechanik, Vol. I, II, III, Teubner, Leipzig, 1942

\bibitem{BuondonnaDeLuca2015} G. Buondonno, A. De Luca: A recursive
Newton-Euler algorithm for robots with elastic joints and its application to
control, 2015 IEEE/RSJ IROS, 5526-5532

\bibitem{BuondonnaDeLuca2016} G. Buondonno, A. De Luca: Efficient
Computation of Inverse Dynamics and Feedback Linearization for VSA-Based
Robots, IEEE Rob. Aut. Letters, 1(2), 2016, 908-915

\bibitem{Bustos2012} I.F. de Bustos, et al.: Second order mobility analysis
of mechanisms using closure equations, Meccanica, 2012, Volume 47, Issue 7,
pp 1695-1704

\bibitem{BottemaRoth1990} O. Bottema, B. Roth: Theoretical Kinematics, North
Holland Press, New York, 1979 (reprint Dover Publications 1990)

\bibitem{Brand1947} L. Brand: Vector and tensor analysis, Wiley, New York,
1947

\bibitem{Brockett1984} R. W. Brockett: Robotic manipulators and the product
of exponentials formula, Mathematical Theory of Networks and Systems,
Lecture Notes in Control and Information Sciences Vol. 58, 1984, pp 120-129

\bibitem{Cauchy1827} A.L. Cauchy: Exercises de math\'{e}matiques, Vol. 2,
1827, page 87

\bibitem{Cervantes2004} J.J. Cervantes-S\'{a}nchez, M.A. Moreno-Baez, J.M.
Rico-Martinez, E.J. Gonzalez-Galvan: A novel geometrical derivation of the
Lie product, Mech. Mach. Theory, Vol. 39, 2004, pp. 1067-1079

\bibitem{Cervantes2009} J.J. Cervantes-S\'{a}nchez et al.: The differential
calculus of screws: Theory, geometrical interpretation, and applications,
Proc. Inst. Mech. E., Part C: Journal of Mechanical Engineering Science,
Vol. 223, 2009, pp. 1449-1468

\bibitem{Chen2011} C. Chen: The order of local mobility of mechanisms, Mech.
Mach. Theory, Vol. 46, 2011, pp. 1251-1264

\bibitem{Chevallier1984} D.P. Chevallier: La formation des \'{e}quations de
la dynamique. Examen des diverses m\'{e}thodes, M\'{e}chanique, 1984

\bibitem{Chevallier1991} D.P. Chevallier: Lie Algebras, Modules, Dual
Quaternions and algebraic Methods in Kinematics, Mech. Mach. Theory, Vol.
26, No. 6, 1991, pp. 613-627

\bibitem{Chevallier1994} D.P. Chevallier: Lie Groups and Multibody Dynamics
Formalism, Proc. EUROMECH Colloquium 320, Prague, 1994, pp. 1-20

\bibitem{Chiacchio1990} P. Chiacchio: Exploiting Redundancy in Minimum-Time
Path Following Robot Control, American Control Conference, 23-25 May 1990,
pp. 2313-2318

\bibitem{ChirikjianAMR2018} G. S. Chirikjian: Discussion of
\textquotedblleft Geometric Algorithms for Robot Dynamics: A Tutorial
Review" by Frank C. Park, Beobkyoon Kim, Cheongjae, Jang, and Jisoo Hong,
FMANU-AMR-17-102

\bibitem{Chiu1988} S.L. Chiu: Task Compatibility of Manipulator Postures,
Vol 7, No. 5, Int. J. Robotics Research, 1988, pp. 13-21

\bibitem{ConnellyServatius1994} R. Connelly, H. Servatius: Higher-Order
Rigidity--What Is the Proper Definition?, Discrete Comput Geom. Vol. 11,
1994, pp. 193-200

\bibitem{ConstantinescuCroft2000} D. Constantinescu, E.A. Croft: Smooth and
Time-Optimal Trajectory Planning for Industrial Manipulators along Specified
Paths, Journal of Robotic Systems. Vol. 17, No. 5, 2000, pp. 233-249

\bibitem{CoxLittleOShea2007} D. Cox, J. Little, and D. O'Shea, Ideals,
Varieties and Algorithms, 3rd ed., Springer, Berlin, 2007

\bibitem{deLuca1998} A. De Luca: Decoupling and feedback linearization of
robots with mixed rigid/elastic joints, Int. J. Rob. Nonlin. Cont., 8, 1998,
965-977

\bibitem{DenavitHartenberg1955} J. Denavit, R. Hartenberg: A kinematic
notation for lower-pair mechanisms based on matrices, Journal of Applied
Mechanics, Vol. 22, 1955, pp. 215-221

\bibitem{DoelPai1996} K. van den Doel, D.K. Pai: Performance Measures for
Robot Manipulators: A Unified Approach, The Int. J. Rob. Research, Vol. 15,
No. 1, 1996, pp. 92-111

\bibitem{Donelan2007} P. Donelan: Singularity-Theoretic Methods in Robot
Kinematics, Robotica, 25 (2007) pp. 641-659

\bibitem{Elkady} A.Y. Elkady, M. Mohammed, T. Sobh: A New Algorithm for
Measuring and Optimizing the Manipulability Index, J Intell Robot Syst, Vol.
59, No. 1, 2010, pp. 75-86

\bibitem{Featherstone1983} R. Featherstone: The Calculation of Robot
Dynamics using Articulated-Body Inertias, Int. J. Robotics Research, Vol. 2,
No. 1, 1983, pp. 13-30

\bibitem{Featherstone2008} R. Featherstone: Rigid Body Dynamics Algorithms,
Springer, 2008

\bibitem{Fijany1995} A. Fijany: Parallel O(1og N) Algorithms for Computation
of Manipulator Forward Dynamics, IEEE Trans. Rob. and Automat., Vol. 11, No.
3, 1995, pp. 389-400

\bibitem{GallardoRico2001} J. Gallardo-Alvarado, J.M. Rico-Martinez: Jerk
Influence Coefficients, via Screw Theory, of Closed Chains, Meccanica, Vol.
36, No. 2, 2001, pp. 213-228

\bibitem{Gallardo2008} J. Gallardo-Alvarado, H. Orozco-Mendoza, R. Rodr\'{\i}%
guez-Castro: Finding the jerk properties of multi-body systems using
helicoidal vector fields, Proc. IMechE, Vol. 222 Part C: J. Mech. Eng. Sci.,
2008, pp. 2217-2229

\bibitem{GallettiFanghella2001} C. Galletti and P. Fanghella, Single-loop
kinematotropic mechanisms, Mech. Mach. Theory, Vol. 36, pp. 743-761, 2001

\bibitem{Giusti2018} A. Giusti, J. Malzahn, N. G. Tsagarakis, M. Althoff: On
the Combined Inverse-Dynamics/Passivity-Based Control of Elastic-Joint
Robots, IEEE Trans. Rob., 2018

\bibitem{Gogu2005} G. Gogu: Mobility of mechanism: a critical review, Mech.
Mach. Theory, vol. 40, 2005, pp. 1068-1097

\bibitem{GosselinAngeles1991} C. Gosselin, J. Angeles: A Global Performance
Index for the Kinematic Optimization of Robotic Manipulators, J. Mech. Des.
Vol. 113, No. 3, 1991, pp. 220-226

\bibitem{SingularBook} G.M. Greuel, G. Pfister: A Singular Introduction to
Commutative Algebra, Springer, 2012

\bibitem{Guarino2006} C. Guarino Lo Bianco, E. Fantini: A recursive
Newton-Euler approach for the evaluation of generalized forces derivatives,
12th IEEE Int. Conf. Methods Models Autom. Robot., 2006, 739-744

\bibitem{Guarino2009} C. Guarino Lo Bianco: Evaluation of Generalized Force
Derivatives by Means of a Recursive Newton--Euler Approach, IEEE Trans.
Rob., 25(4), 2009, 954-959

\bibitem{Gupta1986} K.C. Gupta: Kinematic Analysis of Manipulators Using the
Zero Reference Position Description, Int. J. Robotics Research, 1986, Vol.
5, No. 2, 1986

\bibitem{Hao1998} K. Hao: Dual number method, rank of a screw system and
generation of Lie sub-algebras, Mech. Mach. Theory, Vol. 33, no. 7, 1998,
pp. 1063-1084

\bibitem{Herve1978} J.M. Herv\'{e}: Analyse Structurelle des M\'{e}canismes
par Groupe des D\'{e}placements, Mech. Mach. Theory, Vol. 13, 1978, pp.
437-450

\bibitem{Herve1982} J.M. Herv\'{e}: Intrinsic formulation of problems of
geometry and kinematics of mechanisms, Mech. Mach. Theory, vol. 17, No. 3,
1982, pp. 179-184

\bibitem{Hollerbach1980} J.M. Hollerbach: A Recursive Lagrangian Formulation
of Maniputator Dynamics and a Comparative Study of Dynamics Formulation
Complexity, IEEE Tran. Systems, Man, Cyb., Vol. SMC-10, No. 11, 1980, pp.
730-736

\bibitem{Hunt1978} K.H. Hunt, Kinematic Geometry of Mechanism, 2 and 3,
Oxford University Press, Clarendon, 1978

\bibitem{HustySchroecker2009} M.L. Husty, H.P. Schr\"{o}cker: Algebraic
Geometry and Kinematics. In: Emiris I., Sottile F., Theobald T. (eds):
Nonlinear Computational Geometry. The IMA Volumes in Mathematics and its
Applications, Vol. 151, 2009, Springer, NY

\bibitem{JalonMUBO2013} J. Garc\'{\i}a de Jal\'{o}n, M. D. Guti\'{e}rrez-L%
\'{o}pez: Multibody dynamics with redundant constraints and singular mass
matrix: existence, uniqueness, and determination of solutions for
accelerations and constraint forces, Multibody Systems Dynamics, Springer,
Vol. 30, No. 3, Oct. 2013, pp. 311-341

\bibitem{Jain1991} A. Jain: Unified Formulation of Dynamics for Serial Rigid
Multibody Systems, Journal of Guidance, Control and Dynamics, Vol. 14, No.
3, 1991, pp. 531-542

\bibitem{deJong2018} J.J. de Jong, A. M\"{u}ller, J.L. Herder: Higher-order
Taylor approximation of finite motions in mechanisms, Robotica, Vol. 37, No.
7, 2019, pp. 1190-1201

\bibitem{KargerNovak1985} A. Karger, J. Nov\'{a}k: Space Kinematics and Lie
Groups, Gordon and Breach Science Publishers, New York, 1985

\bibitem{Karger1989} A. Karger: Curvature Properties Of 6-Parametric Robot
Manipulators, Manuscripta Mathematica, Vol. 65, 1989. pp. 311-328

\bibitem{Karger1996} A. Karger: Singularity Analysis of Serial
Robot-Manipulators, ASME J. Mech. Des. Vol. 118, No. 4, 1996, pp. 520-525

\bibitem{Karsai2002} G. Karsai: Method for the calculation of the combined
motion time derivatives of optional order and solution for the inverse
kinematic problems, Mech. and Mach. Theory, Vol. 36, No. 2, 2002, pp. 261-272

\bibitem{KleinBlaho1987} C.A. Klein, B.E. Blaho: Dexterity Measures for the
Design and Control of Kinematically Redundant Manipulators, The Int. J.of
Robotics Research, Vol. 6, No. 2, 1987, pp. 72-83

\bibitem{KhalilKleinfinger1986} W. Khalil, J.F. Kleinfinger: A New Geometric
Notation for Open and Closed-Loop Robots. In: Proceedings of the IEEE
International Conference on Robotics and Automation, 1986, S. 1174-1179

\bibitem{Lee} J. Lee: A Study on the Manipulability Measures for Robot
Manipulators, Proc. International Conference on Intelligent Robots and
Systems IEEE/RSJ (IROS), 1997, pp. 1458-1465

\bibitem{Lerbet1999} J. Lerbet: Analytic Geometry and Singularities of
Mechanisms, ZAMM, Z. angew. Math. Mech., Vol. 78, No. 10b, 1999, pp. 687-694

\bibitem{LerbetBook} J. Lerbet: Multi-Body Kinematics and Dynamics with Lie
Groups, ISTE Press - Elsevier, 2017

\bibitem{Lipkin2005} H. Lipkin: Time derivatives of screws with applications
to dynamics and stiffness, Mechanism and Machine Theory, Vol. 40, 2005, pp.
259-273

\bibitem{LillyOrin1991} W. Lilly, D.E. Orin: Alternate Formulations for the
Manipulator Inertia Matrix, Int. J. Rob. Res., Vol. 10, No. 1, 1991, pp.
64-74

\bibitem{LopezCustodio2017} P.C. L\'{o}pez-Custodio, et al.: Verification of
the higher order kinematic analyses equations, European Journal of Mechanics
A/Solids, Vol. 61, 2017, pp. 198-215

\bibitem{PabloMMT2018} P.C. L\'{o}pez-Custodio, A. M\"{u}ller, J.M. Rico,
J.S. Dai: A synthesis method for 1-DOF mechanisms with a cusp in the
configuration space, Mech. Mach. Theory, Vol. 132, 2019, pp. 154-175

\bibitem{LynchPark2017} K.M. Lynch, F.C. Park: Modern Robotics, Cambridge,
2017

\bibitem{Ma} R.R. Ma, A.M. Dollar: On Dexterity and Dexterous Manipulation,
The 15th International Conference on Advanced Robotics, Tallinn, Estonia,
June 20-23, 2011

\bibitem{McCarthy1990} J.M. McCarthy: Introduction to Theoretical
Kinematics, MIT Press Cambridge, 1990

\bibitem{Merlet} J.P. Merlet: Jacobian, Manipulability, Condition Number,
and Accuracy of Parallel Robots, ASME J.Mech.Des.,Vol.128,2006,pp.199-206

\bibitem{Milenkovic2012} P. Milenkovic: Series Solution for Finite
Displacement of Single-Loop Spatial Linkages, ASME J. Mech. Rob., Vol. 4,
2012, 8 pages

\bibitem{MED2003} A. M\"{u}ller: A dexterity maximizing continuation method
for the inverse kinematics of redundant manipulators, 11th IEEE
Mediterranean Conf. Contr. Aut. 2003 (MED'03), June 18-20, 2003, Rhodes,
Greece

\bibitem{RedConstraints2014} A. M\"{u}ller: Implementation of a Geometric
Constraint Regularization for Multibody System Models, Archive of Mechanical
Engineering, Vol. 61, No. 2, 2014, pp. 376-383

\bibitem{MMTHighDer} A. M\"{u}ller: Higher Derivatives of the Kinematic
Mapping and some Applications, Mechanism and Machine Theory, Vol. 76, 2014,
Pages 70-85

\bibitem{MMTConstraints} A. M\"{u}ller: Recursive Higher-Order Constraints
for Linkages with Lower Linematic Pairs, Mechanism and Machine Theory, Vol.
100, 2016, pp. 33-43

\bibitem{JMR2016} A. M\"{u}ller: Local Kinematic Analysis of Closed-Loop
Linkages -Mobility, Singularities, and Shakiness, ASME J. Mech. Rob., Vol.
8, August 2016, 041013-1

\bibitem{CND2016} A. M\"{u}ller: Coordinate Mappings for Rigid Body Motions,
ASME Journal of Computational and Nonlinear Dynamics, 12(2) 2016,
doi:10.1115/1.4034730

\bibitem{MUBOScrews1} A. M\"{u}ller: Screw and Lie group theory in multibody
dynamics -- Motion representation and recursive kinematics of tree-topology
systems, Multibody System Dynamics, Vol. 43, Vol. 1, pp. 1-34

\bibitem{MUBOScrews2} A. M\"{u}ller: Screw and Lie group theory in multibody
dynamics -- Recursive algorithms and equations of motion of tree-topology
systems, Multibody System Dynamics, 42(2), 219-248

\bibitem{Topology} A. M\"{u}ller: Representation of the Kinematic Topology
of Mechanisms for Kinematic Analysis, Mechanical Sciences, Vol. 6, pp. 1-10,
2015

\bibitem{Robotica2017} A. M\"{u}ller: Kinematic Topology and Constraints of
Multi-Loop Linkages, Robotica, Vol., 36, No. 11, 2018, pp. 1641 - 1663

\bibitem{ICRA2017} A. M\"{u}ller: Recursive Second-Order Inverse Dynamics
for Serial Manipulators, IEEE Int. Conf. Robotics Automations (ICRA), May
29-June 3, 2017, Singapore

\bibitem{JMR2018} A. M\"{u}ller: Higher-Order Analysis of Kinematic
Singularities of Lower Pair Linkages and Serial Manipulators, ASME J. Mech.
Rob., Vol. 10, No 1., 2018

\bibitem{JMR2018LocApprox} A. M\"{u}ller: A Screw Approach to the
Approximation of the Local Geometry of the Configuration Space and of the
set of Configurations of Certain Rank of Lower Pair Linkages, ASME J. Mech.
Rob., Vol. 11, No 2., 2019, 020910 (9 pages)

\bibitem{Murray} R.M. Murray, Z. Li, and S.S. Sastry, A Mathematical
Introduction to Robotic Manipulation, CRC Press Boca Raton, 1994

\bibitem{OrinSchrader1984} D. Orin, W. Schrader: Efficient computation of
the Jacobian for robot manipulators, Int. J. Robotics Research, Vol. 3, No.
4, 1984

\bibitem{PalliMelchiorriDeLuca2008} G. Palli, C. Melchiorri, A. De Luca: On
the Feedback Linearization of Robots with Variable Joint Stiffness, IEEE
Int. Conf. Rob. Aut. (IROS), Pasadena, CA, USA, May 19-23, 2008

\bibitem{ParkBobrowPloen1995} F. C. Park, J. E. Bobrow, S. R. Ploen: A Lie
group formulation of robot dynamics, Int. J. Rob. Research, Vol. 14, No. 6,
1995, pp. 609-618

\bibitem{ParkKim1998} F.C. Park, J.W. Kim: Manipulability of closed
kinematic chains, ASME J. Mech. Des. Vol. 120, No. 4, 1998, pp. 542-548

\bibitem{ParkKimJangHong2018} F.C. Park, B. Kim, C. Jang and J. Hong:
Geometric Algorithms for Robot Dynamics: A Tutorial Review, ASME. Appl.
Mech. Rev., Vol. 79, No. 1, 2018 (8 pages)

\bibitem{PfurnerKong2017} M. Pfurner, X. Kong: Algebraic Analysis of a New
Variable-DOF 7R Mechanism. In: Wenger P., Flores P. (eds) New Trends in
Mechanism and Machine Science. Mechanisms and Machine Science, Vol. 43,
2017, Springer

\bibitem{Raymond1991} X. Saint Raymond: Elementary Introduction to the
Theory of Pseudodifferential Operators, CRC Press, 1991.

\bibitem{ReiterTII2018} A. Reiter, A. M\"{u}ller, H. Gattringer: On
Higher-Order Inverse Kinematics Methods in Time-Optimal Trajectory Planning
for Kinematically Redundant Manipulators, IEEE Trans. Industrial
Informatics, Vol. 14, No. 4, 2018, pp. 1681 - 1690

\bibitem{RicoDuffy1996} J. M. Rico, J. Duffy: An application of screw
algebra to the acceleration analysis of serial chains, Mech. and Mach.
Theory, Vol. 31, No. 4, 1996, pp. 445-457

\bibitem{Rico1999} J.M. Rico, J. Gallardo, J. Duffy: Screw theory and higher
order kinematic analysis of open serial and closed chains, Mech. Mach.
Theory, Vol. 34, No. 4, 1999, pp. 559-586

\bibitem{RicoGallargoRavani2003} J.M. Rico, J. Gallargo, B. Ravani: Lie
Algebra and the Mobility of Kinematic Chains, Journal of Robotics System,
Vol. 20. No. 8, pp. 477-499, 2003

\bibitem{RicoRavani2003} J.M. Rico Martinez, B. Ravani: On mobility analysis
of linkages using group theory, ASME J. Mech. Des. vol. 135, 2003, pp. 70-80

\bibitem{RicoRavani2006} J.M. Rico, B. Ravani: On Calculating the Degrees of
Freedom or Mobility of Overconstrained Linkages: Single-Loop Exceptional
Linkages, J. Mech. Des., Vol. 129, No. 3, 2006, 301-311

\bibitem{Rico2007} J.M. Rico, J.J. Cervantes, J. Gallardo, L.D. Aguilera,
G.I. P\'{e}rez, A. Tadeo: Mobility of single loop linkages: A final word?,
31st ASME Mechanisms and Robotics Conference (MR), September 4-7, 2007, Las
Vegas, Nevada, USA

\bibitem{Rodriguez1987} G. Rodriguez: Kalman Filtering, Smoothing, and
ecursive Robot Arm Forward and Inverse Dynamics, IEEE J. Rob. and
Automation, Vol. RA-3, No. 6, 1987, pp. 624-639

\bibitem{Rodriguez1991} G. Rodriguez, A. Jain, K. Kreutz-Delgado: A Spatial
Operator Algebra for Manipulator Modelling and Control, Int. J. Robotics
Research, Vol. 10, No. 4, 1991, pp. 371-381

\bibitem{Rodriguez1992} G. Rodriguez, A. Jain, K. Kreutz-Delgado: Spatial
Operator Algebra for Multibody System Dynamics, J. Astron Sciences, Vol. 40,
1992, pp. 27-50

\bibitem{Selig} J. Selig: Geometric Fundamentals of Robotics, Springer, New
York, 2005

\bibitem{ShethUicker1971} P.N. Sheth, J.J. Uicker: A Generalized Symbolic
Notation for Mechanisms, J. Eng. Ind, Vol. 93, No. 1, 1971, pp. 102-112

\bibitem{Steeb1991} W.H. Steeb: Kronecker producs of matrices and
applications, BI-Wiss.-Verlag, Mannheim, Wien, Z\"{u}rich, 1991

\bibitem{Study1891} E. Study: Von den Bewegungen und Umlegungen, Math. Ann.,
Vol. 39, 1891, pp. 44- 566

\bibitem{Study1903} E. Study: Geometrie der Dynamen, B. G. Teubner, Leipzig,
1903

\bibitem{Sugimoto2001} K. Sugimoto: Kinematic Analysis and Derivation of
Equations of Motion for Mechanisms with Loops of Different Motion Spaces,
International Journal of the JSME, series C, Vol. 44, No. 3 , 2001, pp.
610-617

\bibitem{Tsai} Y.C. Tsai, A.H Soni: Accessible Region and Synthesis of Robot
Arms, ASME J. Mech. Design, Vol. 103, 1981, pp. 803-811

\bibitem{Uicker2013} J.J. Uicker, B. Ravani, P.N. Sheth: Matrix Methods in
the Design Analysis of Mechanisms and Multibody Systems, Cambridge
University Press, 2013

\bibitem{Veter1973} W.J. Vetter: Matrix calculus operations and taylor
expansions, SIAM Review Vol. 15, No. 2, April 1973

\bibitem{Waldron1982} K.J. Waldron: Geometrically based manipulator rate
control algorithms. Mechanism and Machine Theory, Vol. 17, No. 6, 1982, pp.
379-385

\bibitem{WalterHusty2010} Walter, Dominic R., Husty, Manfred L.: On
Implicitization of Kinematic Constraint Equations, Machine Design \&
Research (CCMMS 2010), Vol. 26, pp. 218-226, Shanghai, 2010 synthesis.
Robotica, Vol. 25, No. 6, 2007, pp. 661-675

\bibitem{Whitney1972} D. E. Whitney: The Mathematics of Coordinated Control
of Prosthetic Arms and Manipulators, Trans. ASME J. Dyn. Sys., Meas.,
Control, Vol. 94, No. 4, 1972, pp. 303-309

\bibitem{WittenburgBook} J. Wittenburg: Dynamics of Multibody Systems,
Springer, 2008

\bibitem{Wojtyra2005} M. Wojtyra: Joint Reaction Forces in Multibody Systems
with Redundant Constraints, Multibody System Dynamics, Vol. 14, No.1, 2005,
14: 23-46

\bibitem{Wohlhart1996} K. Wohlhart: Kinematotropic Linkages, in: J. Lenar%
\v{c}i\v{c}, V. Parent-Castelli (eds.): Recent Advances in Robot Kinematics,
Kluwer, 1996, pp. 359-368

\bibitem{WohlhartShakiness} K. Wohlhart: Degrees of Shakiness, Mech. Mach.
Theory, Vol. 34, 1999, pp. 1103-1126

\bibitem{Wohlhart2010} K. Wohlhart: From higher degrees of shakiness to
mobility, Mech. Mach. Theory, Vol. 45, 2010, pp. 467-476

\bibitem{WuIDETC2018} L. Wu, A. M\"{u}ller, J. Dai: Matrix analysis of
second-order kinematic constraints of single-loop linkages in screw
coordinates, 42nd ASME Mechanisms and Robotics Conference (MR), August
26-29, 2018, Quebec City, Canada

\bibitem{Yoshikawa} T. Yoshikawa, Manipulability of Robotic Mechanisms, The
Int. J. Robotics Research, 1985, Vol. 4, pp. 3-9

\bibitem{ZlatanoNenchev2005} D. Zlatanov, D.N. Nenchev: On the Use of
Metric-Dependent Methods in Robotics, Proc. ASME 2005 International Design
Engineering Technical Conferences, September 24-28, 2005, Long Beach,
California USA

\bibitem{MendeleyDataset} A. M\"{u}ller: Data for: An Overview of Formulae
for the Higher-Order Kinematics of Lower-Pair Chains wit Applications in
Robotics and Mechanism Theory, Mendeley Data, v1
\end{thebibliography}
